%% file: 2019_RSS_arxiv.tex
\documentclass[conference]{IEEEtran}
\usepackage[firstpage=true]{background}
\newcommand\copyrighttext{%
\parbox{\textwidth}{
\footnotesize
In Proceedings of Robotics: Science and Systems (RSS), Freiburg, Germany, June 2019.
\\DOI: 10.15607/RSS.2019.XV.014
}
}

\SetBgContents{\copyrighttext}
\SetBgScale{1}
\SetBgColor{black}
\SetBgAngle{0}
\SetBgOpacity{1}
\SetBgPosition{current page.north}
\SetBgVshift{-0.8cm}

\usepackage[bottom]{footmisc}

\usepackage{times}

\usepackage[numbers]{natbib}
\usepackage{multicol}
\usepackage[bookmarks=true]{hyperref}

\usepackage{makecell}
\usepackage{textcomp}
\usepackage{soul}
\usepackage{todonotes}

\usepackage[capitalize]{cleveref}
\crefname{section}{Sec.}{Sec.}
\crefname{table}{Tab.}{Tab.}

\usepackage{tikz}
\usepackage{pgfplots}
\usepackage{hyperref}

\usepackage{pgfkeys}
    \newenvironment{customlegend}[1][]{%
        \begingroup
        \csname pgfplots@init@cleared@structures\endcsname
        \pgfplotsset{#1}%
    }{%
        \csname pgfplots@createlegend\endcsname
        \endgroup
    }%
    \def\addlegendimage{\csname pgfplots@addlegendimage\endcsname}

\usepgfplotslibrary{fillbetween}
\usetikzlibrary{arrows}
\usetikzlibrary{arrows.meta}
\usetikzlibrary{positioning,calc}
\usetikzlibrary{decorations.pathreplacing}
\usetikzlibrary{decorations.markings}
\usetikzlibrary{fit}
\usetikzlibrary{shapes.callouts}
\usetikzlibrary{shapes.geometric}
\usetikzlibrary{matrix}

\newcommand{\etal}{et\ al.\ }

\newcommand{\eg}{e.g.,\ }

\begin{document}

\title{Value Iteration Networks on Multiple Levels of Abstraction}
\author{\authorblockN{Daniel Schleich, Tobias Klamt, and Sven Behnke}
\authorblockA{Rheinische Friedrich-Wilhelms-Universit\"at Bonn, 
		Autonomous Intelligent Systems, Bonn, Germany\\ Email: \{schleich@ais.uni-bonn.de, klamt@ais.uni-bonn.de, behnke@cs.uni-bonn.de\}}}




%

\maketitle

\begin{abstract}

Learning-based methods are promising to plan robot motion without performing extensive search, which is needed by many non-learning approaches.
Recently, Value Iteration Networks (VINs) received much interest since---in contrast to standard CNN-based architectures---they learn goal-directed behaviors which generalize well to unseen domains.
However, VINs are restricted to small and low-dimensional domains, limiting their applicability to real-world planning problems.

To address this issue, we propose to extend VINs to representations with multiple levels of abstraction.
While the vicinity of the robot is represented in sufficient detail, the representation gets spatially coarser with increasing distance from the robot.
The information loss caused by the decreasing resolution is compensated by increasing the number of features representing a cell.
We show that our approach is capable of solving significantly larger 2D grid world planning tasks than the original VIN implementation.
In contrast to a multiresolution coarse-to-fine  VIN implementation which does not employ additional descriptive features, our approach is capable of solving challenging environments, which demonstrates that the proposed method learns to encode useful information in the additional features. 
As an application for solving real-world planning tasks, we successfully employ our method to plan omnidirectional driving for a search-and-rescue robot in cluttered terrain.

\end{abstract}

\IEEEpeerreviewmaketitle

\section{Introduction}

While search-based and sampling-based methods are well investigated for motion planning~\cite{hart1968formal, lavalle1998rapidly, kavraki1994probabilistic}, they tend to perform extensive, iterative searches for complex high-dimensional tasks. We hypothesize that this issue might be addressed by a higher level of scene understanding.

In other domains, especially in perceptual contexts, hierarchical convolutional neural networks (CNNs) are highly successful, because they learn increasingly abstract representations of their input by decreasing resolution and increasing the number of feature maps~\cite{krizhevsky2012imagenet, schwarz2018rgb}.   
A number of works applied CNNs to robot motion planning in recent years.
This is promising since CNNs, which can be parallelized efficiently on \eg GPUs, enable planning without extensive search.
Standard CNN architectures have been used to map system state observations directly to actions~\cite{levine2016end, bojarski2016end}.
However, those approaches have difficulties to understand the goal-directed behavior of planning and to generalize to unseen domains.

This issue is addressed by, \eg Value Iteration Networks (VINs)~\cite{tamar2016value} or Universal Planning Networks (UPNs)~\cite{srinivas2018universal}. 
Instead of following a strict feed-forward approach, values iterate multiple times in an inner loop to be propagated through the representation.
Those methods show promising results in terms of goal-directed behavior and generalization to unseen domains.
However, they have only been applied to small, low-dimensional problems.
Planning in larger state spaces requires more complex network designs and significantly more training data which becomes at some point infeasible on currently available hardware.
Thus, it is challenging to apply them to most real-world planning problems.

A well-established idea to handle large state spaces is abstraction~\cite{kulkarni2016hierarchical, klamt2018planning}.
An abstract representation describes neighboring states in a spatially/temporally coarser resolution while enriching the representation with additional features.

\begin{figure}
	\input{fig/teaser/teaser.pgf}
	\caption{The general idea of VINs on multiple levels of abstraction (AVINs).}
	\vspace{-0.5cm}
	\label{fig:teaser}
\end{figure}
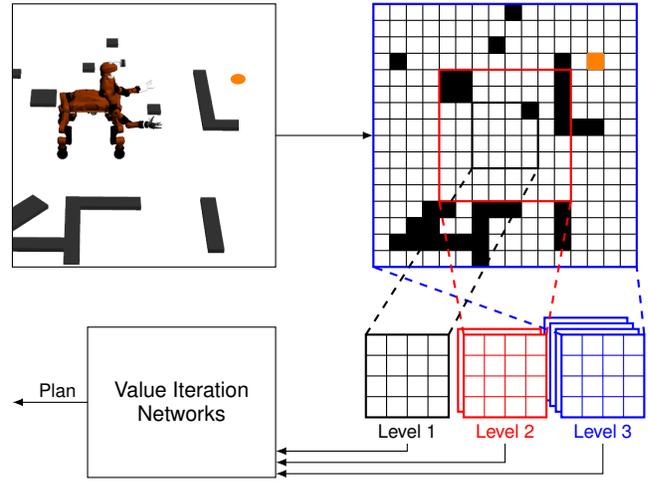

We propose a method to combine multiple environment representations with increasing level of abstraction with VINs to obtain a learning-based planner which is capable of handling more complex tasks (AVINs) (\cref{fig:teaser}).
With increasing distance from the robot, the level of abstraction increases.
While the spatial resolution decreases with an increasing level of abstraction, the number of cells is constant for all levels which results in larger covered areas for more abstract maps.
In addition, an increasing level of abstraction comes along with an increasing number of descriptive features for each cell. 

Experiments show that our proposed AVINs outperform VINs in 2D grid worlds in their original implementation. 
While the result quality of AVINs is comparable or even better, they are capable of planning for environments which are up to 16 times larger while the memory requirement significantly decreases.
In comparison to Hierarchical VINs (HVINs)~\cite{tamar2016value}, which employ multiresolution representations in coarse-to-fine planning without the introduction of additional features, we obtain a better result quality with lower memory requirements.
We especially can show that AVINs learn to encode useful information in their abstract representations since the performance on challenging environments is considerably better in comparison to HVINs.
As a demonstration of the applicability to challenging real-world problems, we apply AVINs to plan omnidirectional driving locomotion for a search-and-rescue robot while considering its individual configuration of ground contact areas (which we refer to as the robot footprint).

\section{Related Work}

Most planning problems can be described as a Markov Decision Process (MDP) which consists of state and action spaces, state transition probabilities for the actions, and reward expectations for each transition~\cite{bellman2013dynamic}.
The goal is to find a policy which results in high long-term rewards.
One common algorithm to find such an optimal policy is Value Iteration (VI).
By applying the Bellman equation~\cite{bellman1957mdp} multiple times, it calculates the expected long-term reward (value) for each state.
The optimal policy is obtained by greedily choosing the action based on the value of possible successor states.

While CNNs are well investigated for tasks such as image classification~\cite{krizhevsky2012imagenet} and robot perception~\cite{schwarz2018rgb}, their application to motion planning arose in recent years.
Traditional CNN architectures have been used to learn policies and directly derive actions from state observations.
Levine et al.~\cite{levine2016end} and Bojarski et al.~\cite{bojarski2016end}, for example, trained CNNs to map raw images to robot motor torques for real-world manipulation tasks and autonomous car steering, respectively.
Although the results of these applications are impressive, such approaches have poor capabilities to plan long-term goal directed behavior and generalization to unseen domains is also an issue.

In 2016, Tamar \etal\cite{tamar2016value} proposed VINs. 
An explicit planning module approximates the VI algorithm by rewriting the application of Bellman equations (which we refer to as Bellman update) as a CNN. 
Since this planning module is fully differentiable, standard backpropagation can be used to learn the parameters of the model, like a suitable reward function or state transition probabilities. 
The embedded planning operation enables VINs to generalize well to unseen environments and understand the desired goal-directed behavior. 
However, VINs do not scale well to larger map sizes and higher-dimensional state spaces since the number of required Bellman updates depends on the path length and larger state spaces require considerably more training data, longer training times, and have large memory requirements.
Hence, evaluation was limited to small 2D grid worlds.
In the appendix of \cite{tamar2016value}, HVINs were proposed to reduce the number of necessary Bellman updates.
Value iteration is first performed on a down-sampled copy of the input map to generate rough state-value estimates, which are up-sampled and used as initialization for another value iteration module working on the full resolution. 
This model can be extended to multiple hierarchical levels.
However, the information loss through down-sampling is not compensated.
Furthermore, all levels operate on the whole environment size resulting in only slightly decreasing memory requirements.

VINs have been applied in other domains.
Niu \etal\cite{gvin}, proposed Generalized Value Iteration Networks which work on arbitrary irregular graph structures and can be applied to real world data like street maps.
\mbox{Karkus \etal\cite{karkus2017qmdp}} proposed QMDP-nets which handle partially observable environments and express VI through a CNN.
Gupta \etal\cite{gupta2017cognitive} propose a Cognitive Mapper and Planner to plan actions from first person views in unknown environments.
They combine a neural network processing first person images to generate a latent representation map of the environment with a hierarchical planning module based on VINs.

UPNs by Srinivas \etal\cite{srinivas2018universal} learn useful latent state representations from images of the current scene and the desired goal scene.
They infer motion trajectories by performing gradient descent planning and iterating over action sequences in the learned internal representations.
Considered environments may have more than two dimensions but are rather small.
The gradient descent planner is very time consuming and hinders scaling to larger environments.
Impressively, UPNs are able to generalize to modified robot morphologies.

In other domains, abstraction is an established method to handle large state spaces.
Abstract states unify multiple detailed states.
This can be realized through coarser resolutions or lower-dimensional representations while the loss of information is compensated by additional features which increase the semantic expressiveness of the representation.
In~\cite{klamt2018planning} the search-based approach for the high-dimensional problem of hybrid driving-stepping locomotion planning~\cite{klamt2017anytime} is extended to plan on multiple levels of abstraction which results in significantly shorter planning times while the result quality stays comparable.
In~\cite{kulkarni2016hierarchical} temporal abstraction is applied to reinforcement learning which generates an efficient space to explore complex environments.

We propose a method to combine VINs with the idea of planning on multiple levels of abstraction to obtain a learning-based planner which is capable of solving planning tasks on challenging, larger state spaces.
The information loss in coarser representations is compensated by increasing the number of features.
In addition, detailed representations are only generated for parts of the environment which decreases memory requirements. 
This increases the applicability of learning-based planning approaches to real-world problems.

\section{Method}
\begin{figure*}
 \centering
  \input{fig/net_architecture_new.pgf}
  \caption{Network architecture. Elements which only account to 2D grid world planning are shown in green. Elements for 3D locomotion planning are purple. The depicted map sizes correspond to $32\times32$ input maps.}
  \label{fig:architecture}
  \vspace{-0.3cm}
\end{figure*}
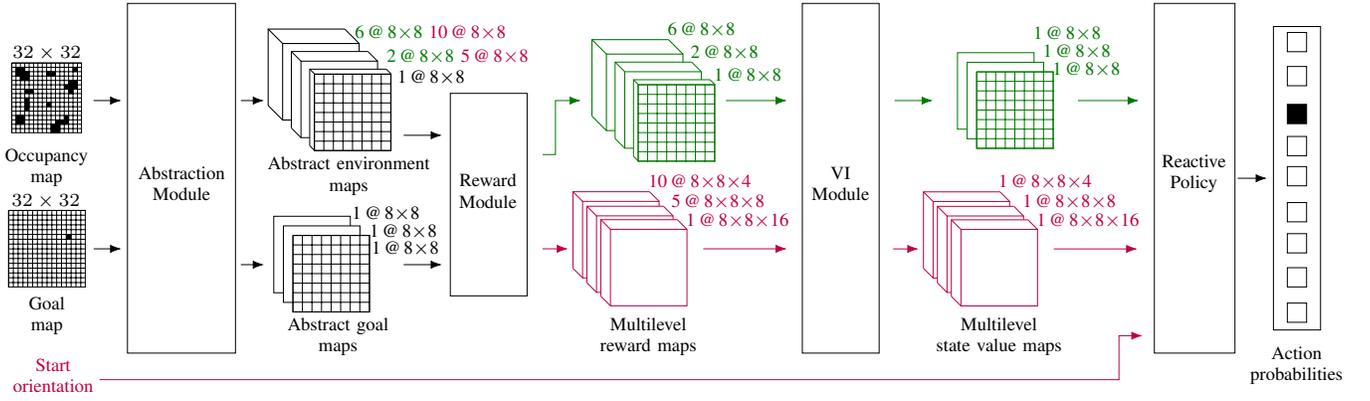

VINs internally represent each state as one cell of a multi-dimensional grid and compute a reward and state-value for each of these grid cells. 
To enable information flow from the goal to the start state, Bellman updates are performed repeatedly within the VI module.
The number of required Bellman updates depends on the maximum possible path length. 
For large and high-dimensional grids, this leads to large computation graphs for the gradients during backpropagation, resulting in long training times and high memory consumption. 
Since the number of states within the VI module is limited, we change what each state represents.
In the vicinity of the robot, which is defined to be always in the center of each map, spatial precision is needed to plan the next robot action.
Regions which are further away from the robot can be described in a coarser, more abstract representation.

As an example, we define three levels of abstraction with a constant number of cells but decreasing resolution.
\mbox{\emph{Level-1}} has the original input resolution but only covers the vicinity of the robot. 
For \mbox{\emph{Level-2}}, the resolution is halved resulting in a four times larger covered area.
This step is repeated to obtain \mbox{\emph{Level-3}}. 
Hence, \mbox{\emph{Level-3}} covers an area which is 16 times larger than the \mbox{\emph{Level-1}} area.
The spatial arrangement of the three representations is depicted in~\cref{fig:teaser}. 

To compensate the information loss in coarser representations, additional features are introduced for each abstract cell and are learned during training.
Experiments showed that one, two, and six features for \mbox{\emph{Level-1}}, \mbox{\emph{Level-2}}, and \mbox{\emph{Level-3}}, respectively, achieved best results.

\subsection{Network Architecture}

Input to the network (\cref{fig:architecture}) is an occupancy map of the environment and an equally sized goal map which only contains zeros except for the goal cell (one-hot-map). 
In contrast to original VINs, we do not provide the system explicit information about the start state, but define that input maps are always robot centered, as also shown in~\cite{Behnke:RoboCup03}.

\begin{figure}
  \centering
  \input{fig/abstraction_module.pgf}
  \caption{Abstraction Module. Both convolutions use kernels of size $3\times 3$ followed by a $2\times2$ max pooling operation. The goal map is processed using only $2\times2$ max pooling operations without prior convolutions. The depicted map sizes correspond to $32\times32$ input maps.}
  \label{fig:learn_abstraction_layers}
  \vspace{-0.55cm}
\end{figure}
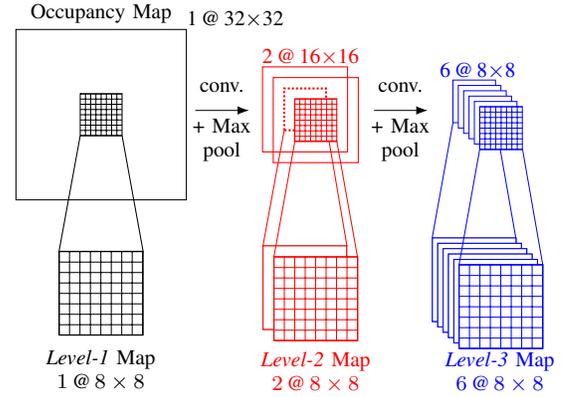

In a first step, the \textbf{Abstraction Module} (\cref{fig:learn_abstraction_layers}) processes the input environment map to three, equally sized abstract environment maps. 
The \mbox{\emph{Level-1}} map is extracted as a patch around the center of the occupancy map. 
A convolution and subsequent max pooling operation generate the \mbox{\emph{Level-2}} representation with halved resolution from the input map.
While the \mbox{\emph{Level-2}} map is again extracted from the map center, the whole \mbox{\emph{Level-2}} representation is processed similarly to obtain the \mbox{\emph{Level-3}} map.
The goal map is processed similarly using max pooling operations without convolutions.

\begin{figure*}
   \centering
   \input{fig/reward_module_final.pgf}
   \caption{Reward Module. \mbox{\emph{Level-1}} maps are shown in black, red parts belong to \mbox{\emph{Level-2}} and blue parts to \mbox{\emph{Level-3}}. 
   Depicted map sizes correspond to $32\times32$ input maps.}
 \label{fig:reward_computation}
 \vspace{-0.5cm}
\end{figure*}
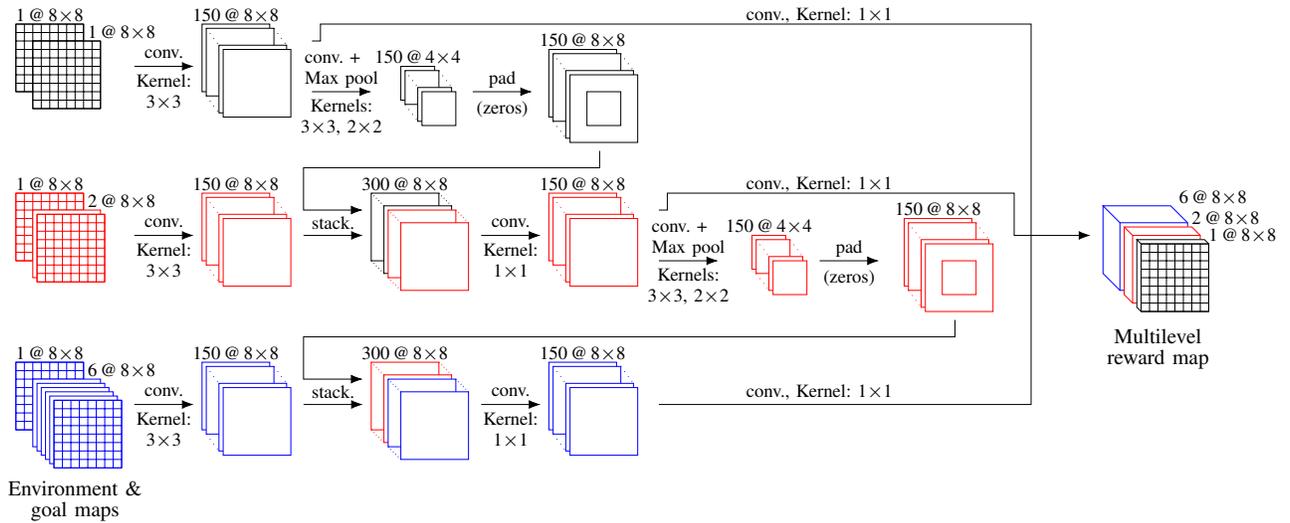

Subsequently, the abstract environment maps and the goal maps are fed into the \textbf{Reward Module} (\cref{fig:reward_computation}) generating rewards for each state.
Since the abstract environment maps have multiple features per cell, this needs to be considered in the reward computation.
It, \eg might be possible that an abstract map cell can be entered from one direction but not from another, which might be encoded in the features.
Hence, multiple reward features are necessary for each cell to represent such information.
It is important to understand that information encoded at the same cell position of different abstraction level maps refer to different locations in the environment. 
We support the network in understanding this relation with the following method:
The \mbox{\emph{Level-1}} reward map is obtained by stacking the environment and goal maps and processing them with two convolutions whose parametrization is inspired by original VINs. 
These convolutions use a padding to keep the map size constant. 
Thus, the relation between cell position and environment location stays fixed.

To enable information flow between levels, the intermediate \mbox{\emph{Level-1}} features extracted by the first convolution are also used within the \mbox{\emph{Level-2}} reward map generation.
Similar to the Abstraction Module, a convolution and subsequent max pooling operation abstract the \emph{Level-1} feature map and match the resolution with \mbox{\emph{Level-2}}.
The result is padded with zeros to match the size of the \mbox{\emph{Level-2}} map.
This procedure ensures that information at the same cell position in both maps describe the same environment location.
The \mbox{\emph{Level-2}} environment and goal maps are processed similarly to \mbox{\emph{Level-1}} to obtain the \mbox{\emph{Level-2}} reward map.
However, after the first convolution, the intermediate \mbox{\emph{Level-2}} feature map is stacked with the abstracted \mbox{\emph{Level-1}} feature map and processed by a $1\times1$ convolution.
Similarly, the resulting combined feature map is used for the \mbox{\emph{Level-3}} reward map generation.


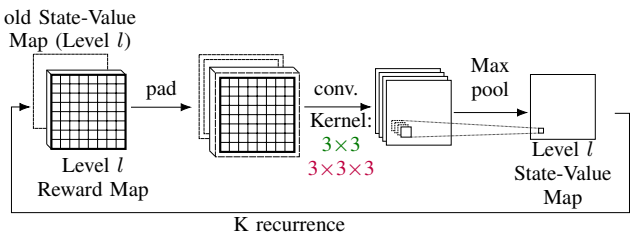
\begin{figure}[!b]
  \centering
  \vspace{-0.3cm}
 \input{fig/vi_module_new.pgf}
  \caption{Value Iteration Module. The depicted operation is performed for each level individually. The padding operation (\cref{fig:padding}) enables information flow between levels. Elements which only belong to 2D grid world/3D robot locomotion planning are shown in green/purple.}
 \label{fig:vi_module}
 \vspace{-0.0cm}
\end{figure}

\begin{figure}
\vspace{-3.6cm}
  \centering
  \input{fig/padding.pgf}
  \caption{Left: Padding the map of abstraction level $l$ (bottom) to allow information flow from 
the map of level $l+1$ (top) to level $l$. The numbers indicate which values are copied where. 
Right: Orientation padding during 3D VIs to emphasize that the orientations $\theta=15$ and $\theta=0$ are neighbors.}
  \label{fig:padding}
  \vspace{-0.5cm}
\end{figure}
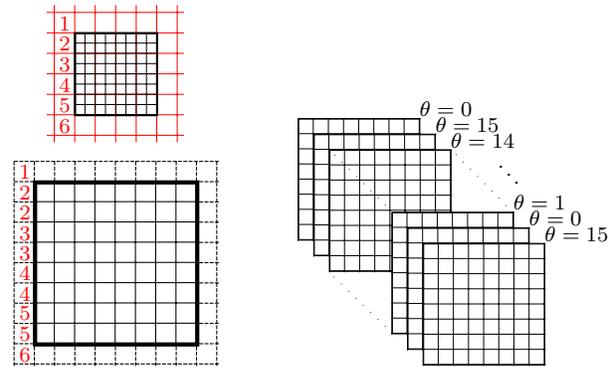

Reward maps are input to the \textbf{VI Module} (\cref{fig:vi_module}) where they are processed to state-value maps. 
Each iteration of the Bellman update is realized by a convolution and subsequent max pooling operation.
The kernel is chosen such that it covers the set of possible actions and thus can propagate state values through the map, respectively. 
Unlike the reward maps, state-value maps consist of only one channel as they describe the expected long-term reward  for a state.
To enable information flow between levels, we apply a padding to the input maps at the beginning of each iteration, as shown in~\cref{fig:padding}.
The padded area contains values of the neighboring cells of the next higher abstraction level.
Since the reward maps vary in their number of features, a mapping from higher-level to lower-level features is required.
We found that the best result quality was achieved by using the average over all features of one higher level-cell as the padding value for all corresponding lower level-cells.
Learning a mapping from higher-level to lower-level features with fully connected layers performed worse.

Finally, for all neighbors of the start state, their state-values are mapped to probabilities over actions through a \textbf{Reactive Policy}, which simply is a fully connected layer.

We apply the proposed architecture to two planning problems: 2D grid worlds and 3D robot locomotion.
The former is used to compare against original VINs and HVINs while the latter demonstrates the capabilities of our approach to handle problems of higher complexity.
Necessary specifications and modifications for each planning domain are described in the following.
The network is implemented using Python 2.7 and PyTorch 0.4.1. Respective source code is available online\footnote{\scriptsize{\url{https://github.com/AIS-Bonn/abstract_vin}}}.

\subsubsection{2D Grid Worlds}
\begin{figure}
  \centering
a)\hspace*{-1ex}\mbox{\input{fig/move_2d.pgf}}
  \hspace{1em}
b)\hspace*{-1ex}\mbox{\input{fig/move_3d.pgf}}
  \hspace{1em}
c)\hspace*{-2ex}\mbox{\input{fig/rotate.pgf}\hspace*{-2ex}}
  \caption{Possible actions for planning domains. a) Moving to an adjacent neighbor cell in 2D grid 
worlds, b) drive to an adjacent neighbor state with fixed orientation in 3D robot locomotion 
planning, and c) turn to the next discrete orientation with fixed position in 3D robot locomotion 
planning.}
\label{fig:actions}
\vspace{-0.4cm}
\end{figure}
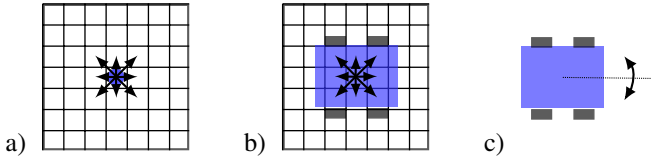

The planner is given queries for a point-like agent in 2D grid worlds.
The goal map is input as a one-hot map.
As actions, the agent can move to one of the eight adjacent neighbor cells (\cref{fig:actions}\,a). 
 
\subsubsection{3D Robot Locomotion}
Given is a robot that can perform omnidirectional driving and has a fixed footprint.
Possible actions for the agent are (\cref{fig:actions}\,b,\,c):
\begin{itemize}
	\item Move to one of the eight adjacent neighbor states with fixed orientation and
	\item turn to the next discrete orientation with fixed position.
\end{itemize}
 
When generating training data and evaluating the network, collision checking is done by checking if any cell which is occupied by the robot footprint is also occupied by an obstacle.
Hence, for robots with modular footprints, it is possible to, \eg take obstacles between their legs.
   
To enable the network to handle 3D agent states, reward and value maps are extended by one additional dimension for the orientation (\cref{fig:architecture}).
We represent the robot orientation for \mbox{\emph{Level-1}} in 16, for \mbox{\emph{Level-2}} in eight, and for \mbox{\emph{Level-3}} in four discrete orientations of equal angular distance.
Due to the increased complexity of the agent states, we increase the number of features for \mbox{\emph{Level-2}} to five and for \mbox{\emph{Level-3}} to ten.
Furthermore, we increase the number of convolutions within the Reward Module by two additional convolutions for processing the \mbox{\emph{Level-1}} map and one additional convolution for the \mbox{\emph{Level-2}} map.
To consider detailed collision checking for the robot footprint, we transform the reward map at the end of the Reward Module:
For each possible robot base pose, we sum over the four cells corresponding to the wheel positions and assign the result to the cell of the robot base pose.

In the VI Module, the convolution kernel needs to cover all possible actions which results in a 3D kernel.
Since the neighborhood relation for the orientation is cyclic, we pad the reward maps and state-value maps on the orientation channel on each end with the values of the opposite end (\cref{fig:padding}).

Other than 2D planning, the architecture for 3D planning needs information about the start and goal orientation. 
The start orientation is fed into our system as an additional parameter. 
It is only used within the Reactive Policy to select those state-values which belong to neighbor states of the start state.  
The goal orientation is encoded in the goal map in which all cell entries are $0$, except for the goal cell which carries the index of the discrete orientation ($1 - 16$).

\subsection{Training}

Training data is generated by placing obstacles of random number, size and position into a 2D grid world.
In addition, multiple goal states are placed randomly. 
Since similar data sets are used by Tamar \etal\cite{tamar2016value}, they offer a high comparability to the original VIN implementation.
The same method is used to generate training data for the 3D Locomotion application.
In addition, more challenging maps are obtained by generating mazes in those 2D grid worlds.
For all maps, the start state is defined to be in the map center.
Subsequently, we use an $A^*$ planner as an expert to generate optimal paths.

Overall, we generated 5,000 environments of the 2D random obstacle grid worlds and 5,000 environments of the 2D maze grid worlds. Seven planning tasks were defined for each environment, resulting in 35,000 different training scenes for each domain. 
The validation and test sets both consist of 715 additionally generated environments with seven planning tasks each, resulting in 5,005 different scenes for each set. 
This applies to the 2D random obstacle grid world, the 2D maze grid world, and the random obstacle grid worlds which are used for 3D planning.
To increase data efficiency during training, we do not only use the whole expert paths but also sub-paths, which are generated by randomly placing the start and goal states on the expert path.
Hence, the amount of training data increases significantly.
We discovered that in the training data set, some actions were chosen more often than other actions.
To support the training, we weight the losses for the different actions by the inverse action frequencies.

\begin{figure}
\input{fig/plot/new_training_performance.pgf}
\vspace{-0.1cm}
\caption{Training performance of VINs, HVINs, and AVINs on the validation set. Left: Fixed learning rate.  Right: Cyclic learning rate scheduling. While the fixed learning rate training was evaluated every 30 epochs, the training with cyclic learning rate scheduling was only evaluated at the end of each learning rate cycle. Especially VINs showed partially unstable training behavior which has also been reported in \eg\cite{lee2018gated}.}
\label{fig:training_performance_success}
\vspace{-0.4cm}
\end{figure}
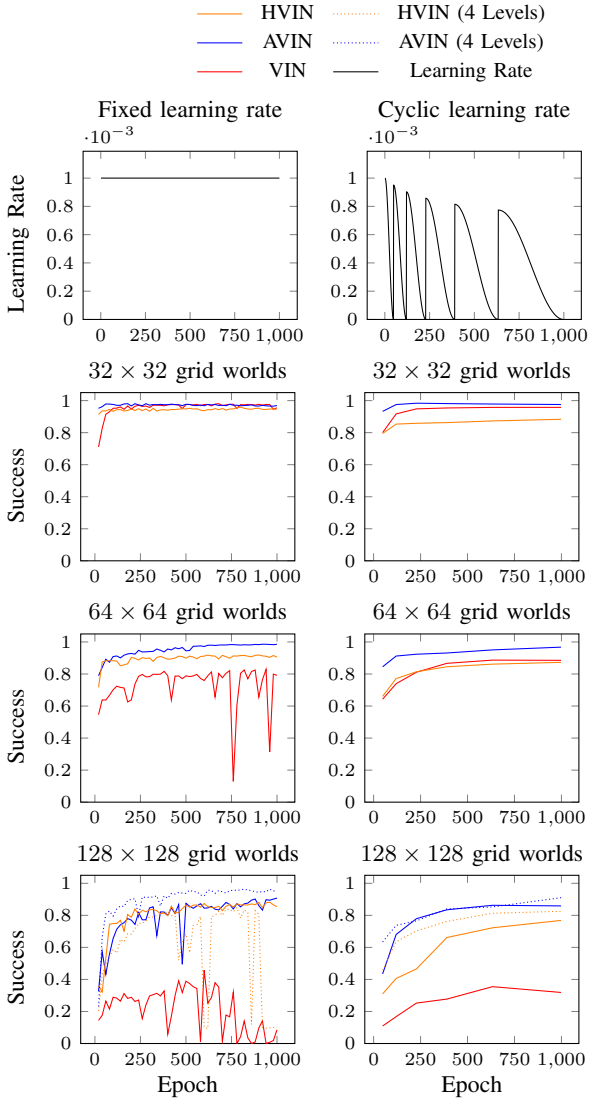

When evaluating our approach against VINs and HVINs, all networks are trained using the RMSprop optimizer as proposed in~\cite{RMSprop}, which was also used in the original VIN publication. 
However, when using the RMSprop without any further learning rate scheduler, the network converges to sub-optimal local minima.
Employing the cyclic learning rate scheduler proposed in~\cite{cyclic_lr} results in a stabilized training performance and better results on the validation set:
During training, the learning rate decreases following a cosine annealing scheme. 
After several training epochs, the learning rate is reset to a higher value. 
We call the time between learning rate resets a learning rate cycle. 
Initially, the length of a learning rate cycle is set to $48$ epochs and the learning rate is $0.001$. 
After each cycle, the cycle length increases to $150$\% while the initial learning rate decreases to $95$\% of the previous one.
We compare the results of the cyclic learning rate scheduling to a fixed learning rate of $0.001$ in the experiment section. 
\Cref{fig:training_performance_success} visualizes the learning rate behavior, depicts the training performance on the 2D random obstacle grid domain, and compares it to original VINs and HVINs.

The network is designed to output the next action for a given input. 
To obtain a path for solving a planning problem, we iteratively let the network predict the next action and update the input maps according to the new robot position.
A path is considered successful if it reaches the goal without obstacle collision and within no more than twice the optimal number of actions, as determined by the expert $A^*$ planner. 
The \emph{success} measures if the network was able to plan a path to the goal.
While HVINs and AVINs were normally implemented using three representation levels, we implemented an additional version with four representation levels for the $128\times128$ maps. 
For AVINs, \emph{Level-4} describes each map cell with ten features. 

The training performance in \cref{fig:training_performance_success} indicates that our method obtains better \emph{success} rates than VINs and HVINs on the validation set.
VINs become unstable for large maps.
For the $128\times128$ maps, it can be seen that both HVINs and AVINs benefit from a forth representation level.


\section{Experiments}
\label{sec:experiments}
All experiments were done on a system equipped with an Intel Core i7-8700K@3.70\,GHz, 64\,GB RAM, and an NVidia GeForce GTX 1080Ti with 11\,GB memory.
A video with additional footage of the experiments is available online\footnote{\scriptsize{\url{https://www.ais.uni-bonn.de/videos/RSS_2019_Schleich/}}}.

To evaluate the network performance, we consider two measures.
The above-mentioned \emph{success} rate evaluates the network performance on the desired task: path planning.
In addition, to compare against VINs, we evaluated the \emph{accuracy} describing how often the network chooses the same next action as the expert $A^*$ planner.
Please note that in many cases there is more than one optimal next action.
However, as in original VINs, the network is trained to output one next action which is compared to the planner.
Hence, there occur cases in which the output of the network is different from the output of the $A^*$ planner but the network still unrolls an optimal path although the \emph{accuracy} measures a mistake.
Stated planning times and memory consumption of our approach include input maps shifting after each network inference to concatenate the next-action network outputs to paths.

\subsection{Path Planning in 2D Random Obstacle Grid Worlds}
\label{sec:exp_2d_random_obstacle}

In a first experiment, we compared our AVINs against VINs and HVINs on the test sets of the random obstacle grid world domain.
Since a similar test set was used in the original VIN publication, this experiment provides good comparability of the methods' capabilities.
We used an implementation with three representation levels for HVINs and AVINs, each level halving the resolution of the previous one.
For HVINs, the lowest resolution level used the same number of Bellman updates $K$ as proposed for original VINs.
This coarse state-value initialization was then refined by two Bellman updates on the medium resolution map and two consecutive Bellman updates on the fine resolution map.

\begin{table}[b]
\vspace{-0.2cm}
  \caption{Results for 2D random obstacle grid worlds with fixed learning rate training.
  		   All stated numbers are averaged over five network instances with different random seed initializations. 
  		   For the $128\times128$ map size, results of HVINs and AVINs with three and four representation levels are given.}
  \label{tab:2Drandom_results}
  \vspace{-0.2cm}
\newcommand{\mc}[3]{\multicolumn{#1}{#2}{#3}}
\begin{center}
\begin{tabular}{l|ccc}
\mc{1}{c|}{$32\times32$} & VIN & HVIN & AVIN\\
\hline
Accuracy & 80.38\% & 80.08\% & \textbf{84.52\%}\\
Success & 91.72\% & 93.84\% & \textbf{97.18\%}\\
Path difference & 2.48\% & 2.23\% & \textbf{1.63\%}\\
Graphics memory [MB] & 761 & 739  & \textbf{685}
\end{tabular}
\end{center}

\begin{center}
\begin{tabular}{l|ccc}
\mc{1}{c|}{$64\times64$} & VIN & HVIN & AVIN\\
\hline
Accuracy & 70.08\% & 77.42\% & \textbf{81.60\%}\\
Success & 71.98\% & 86.82\% & \textbf{94.02\%}\\
Path difference & 2.94\% & 2.55\% & \textbf{1.34\%}\\
Graphics memory [MB] & 1815 & 1399 & \textbf{969}
\end{tabular}
\end{center}

\begin{center}
\begin{tabular}{l|ccccc}
\mc{1}{c|}{$128\times128$} & VIN & HVIN-3 & HVIN-4 & AVIN-3 & AVIN-4\\
\hline
Accuracy & 55.96\% & 76.50\% & 78.04\% & 77.70\% & \textbf{83.54\%} \\
Success & 31.56\% & 83.24\% & 84.00\% & 78.52\% & \textbf{88.72\%} \\
Path diff. & 8.46\% & 2.09\% & 2.80\% & 3.60\% & \textbf{1.84\%} \\
GRAM [MB] & 8247 & 4085 & 4049 & 2189 & \textbf{1167}
\end{tabular}

\end{center}

\vspace{-0.0cm}
\end{table}

\begin{figure}
\centering
\includegraphics[width=0.48\linewidth, clip, trim=0 0 0 0]{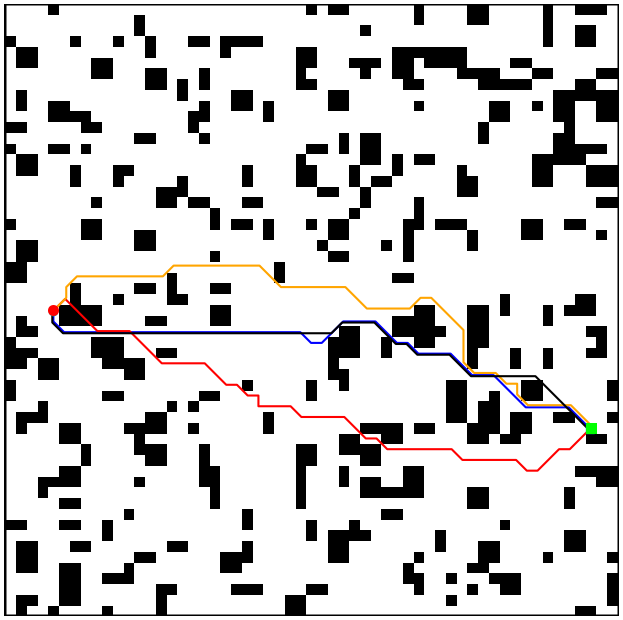}
\includegraphics[width=0.48\linewidth, clip, trim=0 0 0 0]{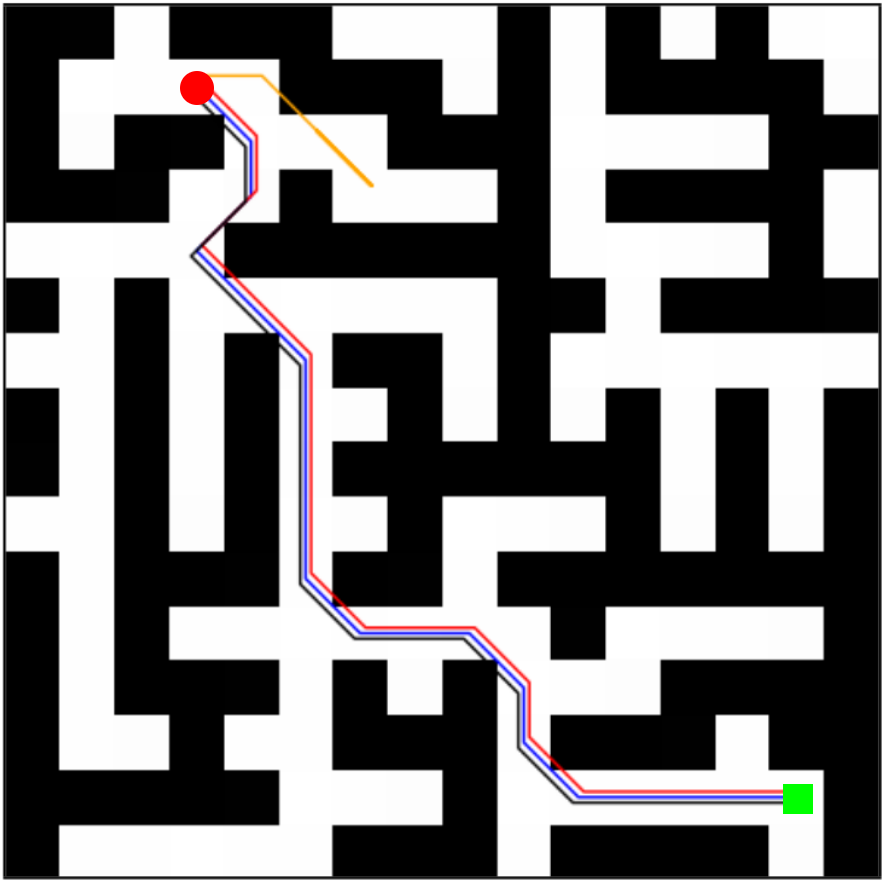}
\caption{Result paths. The figure only depicts the corresponding map sections. The start is marked with a red circle and the goal with a green sqaure. Left: $128\times 128$ random obstacle grid world. VINs (red) fail to find a collision free path, HVINs (orange) react to obstacles when approaching them while AVINs (blue) shows better long-term understanding. A optimal path obtained by the $A^*$ planner is depicted in black. Right: $32\times 32$ maze. HVINs (orange) are not able to find a path. VINs (red), our AVINs (blue), and the $A^*$ planner (black) provide the same optimal solution.}
\label{fig:result_2d} 
\vspace{-0.3cm}
\end{figure}

In~\cite{tamar2016value}, grid world sizes from $8\times 8$ to $28\times 28$ are considered.
We perform tests on slightly larger maps with $32\times 32$, and significantly larger maps with $64 \times 64$ and $128 \times 128$ cells.
For the largest map size, we tested the additional implementation of HVINs and AVINs using four representation levels.
Table~\ref{tab:2Drandom_results} states the results for training with the fixed learning rate while results of a training with cyclic learning rate scheduling are given in \cref{tab:2Drandom_results_lr_cycle}.
\Cref{fig:result_2d} depicts paths on a $128\times128$ map.

\begin{table}
  \caption{Results for 2D grid worlds with cyclic learning rate scheduling. Graphics memory usage is similar to \Cref{tab:2Drandom_results}.}
  \vspace{-0.1cm}
  \label{tab:2Drandom_results_lr_cycle}
  \vspace{-0.2cm}
\newcommand{\mc}[3]{\multicolumn{#1}{#2}{#3}}
\begin{center}
\begin{tabular}{l|ccc}
\mc{1}{c|}{$32\times32$} & VIN & HVIN & AVIN\\
\hline
Accuracy & 84.92\% & 81.36\% & \textbf{85.00\%}\\
Success & 95.26\% & 88.27\% & \textbf{97.56\%}\\
Path difference & \textbf{1.01\%} & \textbf{1.01\%} & 1.56\% 
\end{tabular}
\end{center}

\begin{center}
\begin{tabular}{l|ccc}
\mc{1}{c|}{$64\times64$} & VIN & HVIN & AVIN\\
\hline
Accuracy & 78.88\% & 80.46\% & \textbf{83.75\%}\\
Success & 89.17\% & 88.33\% & \textbf{94.99\%}\\
Path difference & \textbf{1.02\%} & \textbf{1.02\%} & 1.28\% 
\end{tabular}
\end{center}

\begin{center}
\begin{tabular}{l|ccccc}
\mc{1}{c|}{$128\times128$} & VIN & HVIN-3 & HVIN-4 & AVIN-3 & AVIN-4\\
\hline
Accuracy & 66.41\% & 77.46\% & 79.34\% & 84.28\% & \textbf{85.01\%} \\
Success & 34.89\% & 77.40\% & 83.56\% & 86.85\% & \textbf{91.59\%} \\
Path diff. & 1.02\% & 1.02\% & 1.02\% & \textbf{0.80\%} & 1.31\% 
\end{tabular}
\end{center}

\vspace{-0.2cm}
\end{table}

The results indicate that our AVINs outperform VINs and HVINs on all map sizes with both learning rate behaviors in terms of \emph{accuracy}, \emph{success}, and memory consumption.
While VINs and AVINs show a consistently better performance with the cyclic learning rate scheduling, HVINs perform better with the constant learning rate.
It can be furthermore seen that on the $128\times128$ maps, both HVINs and AVINs benefit from a forth representation level in all measures.

\subsection{Path Planning in 2D Maze Grid Worlds}

In a second experiment, we aimed at investigating the limitations of our proposed method and the quality of its abstraction.
We compared it to VINs and HVINs on 2D maze grid worlds.
Mazes possess a larger information density in comparison to the random obstacle grid worlds since the occupancy of nearly every single grid cell is important.
Hence, when generating coarser representations, the effect of information loss is large.
This puts the focus on the quality of the abstraction which shall learn to encode all required information in the additional features.
Table~\ref{tab:results_maze} states the performance of VINs, HVINs, and AVINs on map size of $16\times16$, $32\times32$, and $64\times64$.
Training was performed with fixed learning rate.
An example maze and generated paths are depicted in \cref{fig:result_2d}.

\begin{table}
  \caption{Results for 2D maze grid worlds.}
  \label{tab:results_maze}
  \vspace{-0.2cm}
\newcommand{\mc}[3]{\multicolumn{#1}{#2}{#3}}
\begin{center}
\begin{tabular}{l|ccc}
\mc{1}{c|}{$16\times16$} & VIN & HVIN & AVIN\\
\hline
Accuracy & \textbf{94.42\%} & 87.20\% & 85.59\%\\
Success & \textbf{94.48\%} & 87.42\% & 86.88\%\\
Path difference & \textbf{0.51\%} & 2.02\% & 1.96\%\\
Graphics memory & \textbf{569\,MB}  & 575\,MB  & 635\,MB
\end{tabular}
\end{center}

\begin{center}
\begin{tabular}{l|ccc}
\mc{1}{c|}{$32\times32$} & VIN & HVIN & AVIN\\
\hline
Accuracy & \textbf{85.60\%} & 69.94\% & 82.17\%\\
Success & \textbf{82.10\%} & 48.54\% & 71.50\%\\
Path difference & \textbf{0.88\%} & 2.02\% & 1.09\%\\
Graphics memory & 761\,MB & 739\,MB & \textbf{685\,MB}
\end{tabular}
\end{center}

\begin{center}
\begin{tabular}{l|ccc}
\mc{1}{c|}{$64\times64$} & VIN & HVIN & AVIN\\
\hline
Accuracy & \textbf{84.58\%} & 58.82\% & 81.57\%\\
Success & \textbf{78.02\%} & 14.22\% & 59.39\%\\
Path difference & 1.49\% & 1.99\% & \textbf{0.68\%}\\
Graphics memory & 1815\,MB & 1399\,MB & \textbf{969\,MB}
\end{tabular}
\end{center}

\vspace{-0.2cm}
\end{table}

Since original VINs perform no abstraction procedure, it was to expect that they obtain the best \emph{accuracy} and \emph{success} rates in this challenging domain. 
However, while the performance difference between HVINs and VINs is rather small in the random obstacle domain, this difference increases considerably for large maze worlds. 
For the $16\times 16$ maps, our approach performs worse than HVINs.
An explanation for this might be that, for this input map size,
\mbox{\emph{Level-1}} only has a size of $4\times4$ which might be
insufficient to detailly plan next actions in the vicinity of the robot.
Nevertheless, our method significantly outperforms HVINs on larger maps indicating the advantage of our abstraction method---which introduces additional features to compensate information loss---compared to HVINs, which employ no additional features.

\subsection{Planning 3D Locomotion with Footprint Consideration}

\begin{figure}
  \centering
	\includegraphics[width=0.5\linewidth]{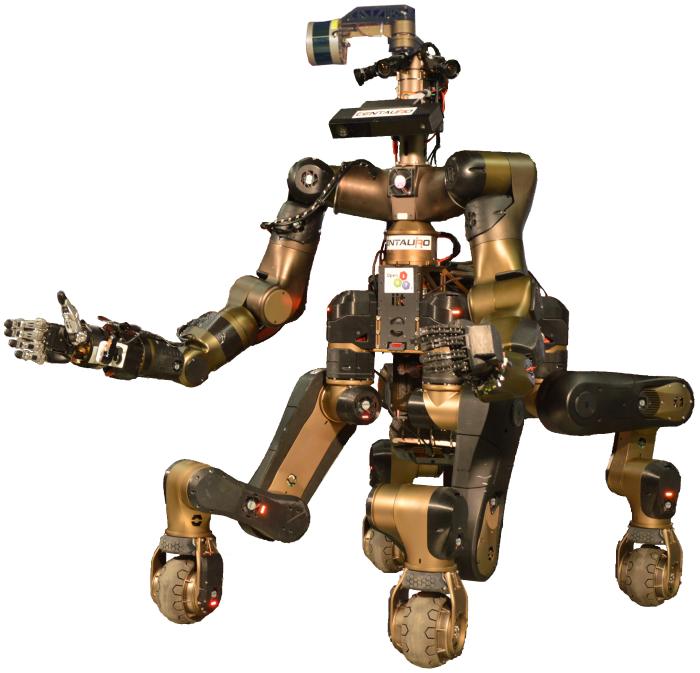}
	\caption{The Centauro robot.}
	\label{fig:Centauro} 
	\vspace{-0.4cm}
\end{figure}

In a third experiment, we investigated the capabilities of the proposed method to solve significantly more complex planning tasks.
We employed the 3D version of AVINs to plan omnidirectional driving locomotion while considering the robot footprint. 
An example platform is the quadrupedal disaster response robot Centauro \cite{klamt2018supervised} whose legs end in 360\textdegree~steerable, active wheels (\cref{fig:Centauro}).
We chose a fixed leg configuration with $0.8$\,m longitudinal and lateral distance between wheels.
Environment maps had a resolution of $0.2$\,m.

At first, we employed our method to plan paths for the described footprint on the $32\times32$ maps of the random obstacle grid domain.
Averaged over five training runs, we achieved an average success rate of 74.20\% for the 5,005 tasks in the test set while our paths were on average 1.86\% longer than the optimal solution. 
The required graphics memory was 865\,MB.
We further compared the planning times for both the $A^*$ planner and AVINs for this 3D locomotion task and the 2D planning task (see \Cref{sec:exp_2d_random_obstacle}) for the $32\times32$ map size. 

As can be seen in \Cref{tab:planning_time_comp}, the $A^*$ planner is in average about 23 times faster than AVINs on the 2D planning task.
However, for the more complex 3D planning tasks with footprint consideration, both planners have similar planning times.
This observation supports our assumption that learning-based planners are promising for complex planning tasks since they do not perform extensive, iterative searches, as traditional planners do.

\begin{table}[b]
\vspace{-0.1cm}
  \caption{Planning times comparison between the $A^*$ planner and AVINs for 2D and 3D planning tasks.}
  \label{tab:planning_time_comp}
  \vspace{-0.1cm}
\newcommand{\mc}[3]{\multicolumn{#1}{#2}{#3}}
\begin{center}

\begin{tabular}{l|cc}
\mc{1}{c|}{$32\times32$} & $A^*$ planner & AVIN\\
\hline
2D planning task & 0.004\,sec & 0.093\,sec \\
3D planning task with footprint & 0.263\,sec & 0.283\,sec
\end{tabular}

\end{center}
\vspace{-0.0cm}
\end{table}

\begin{table}[b]
 \caption{Results of our approach and the $A^*$-planner for the tasks depicted in~\cref{fig:experiment_3d}.}
  \label{tab:results_3d}
  \vspace{-0.2cm}
\newcommand{\mc}[3]{\multicolumn{#1}{#2}{#3}}
\begin{center}
\begin{tabular}{cclcl}
 & \mc{2}{c}{AVIN} & \mc{2}{c}{$A^*$-planner}\\
 & Path length & \mc{1}{c}{Planning Time} & Path length & \mc{1}{c}{Planning Time}\\
I) & $24.59$ & \mc{1}{c}{$0.431$\,sec} & $23.41$ & \mc{1}{c}{$0.169$\,sec}\\
II) & \mc{2}{c}{Not found} & $24.14$ & \mc{1}{c}{$0.980$\,sec}\\
III) & $18.49$ & \mc{1}{c}{$0.342$\,sec} & $17.90$ & \mc{1}{c}{$0.102$\,sec}\\
IV) & $18.80$ & \mc{1}{c}{$0.363$\,sec} & $18.80$ & \mc{1}{c}{$0.341$\,sec}\\
V) & \mc{2}{c}{Not found} & $27.76$ & \mc{1}{c}{$2.117$\,sec}\\
VI) & $18.65$ & \mc{1}{c}{$0.321$\,sec} & $17.01$ & \mc{1}{c}{$0.172$\,sec}\\
VII) & $15.55$ & \mc{1}{c}{$0.321$\,sec} & $15.55$ & \mc{1}{c}{$0.051$\,sec}\\
VIII) & $24.67$ & \mc{1}{c}{$0.449$\,sec} & $22.92$ & \mc{1}{c}{$0.223$\,sec}\\
IX) & $21.13$ & \mc{1}{c}{$0.405$\,sec} & $21.13$ & \mc{1}{c}{$0.705$\,sec}
\end{tabular}
\end{center}
\vspace{-0.2cm}
\end{table}

\begin{figure}
\centering
\includegraphics[width=0.5\linewidth]{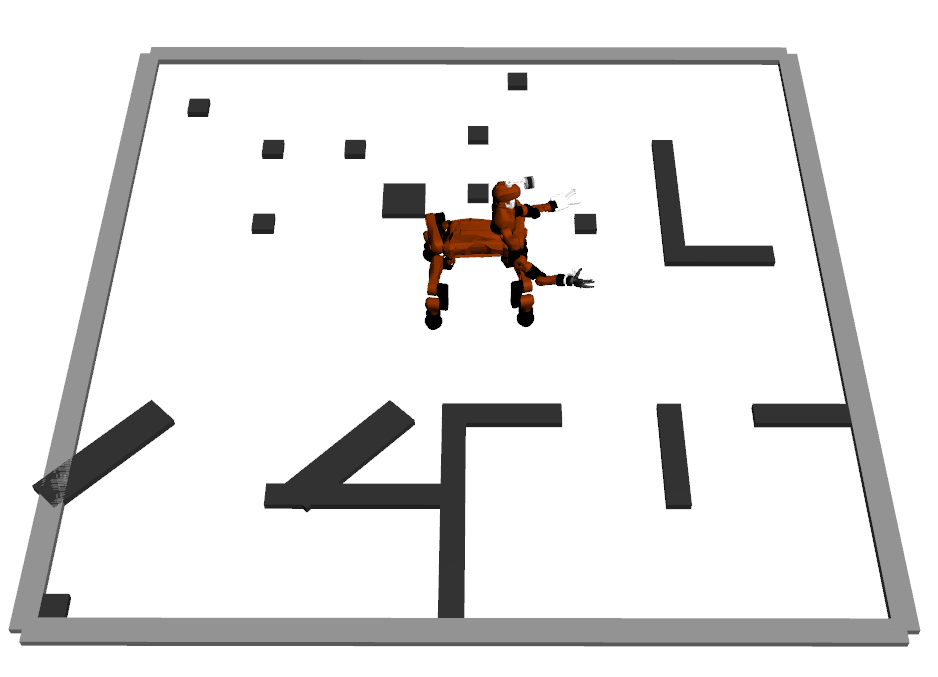}
\hspace{1em}
\input{fig/3D/occ_map_final.pgf}
\caption{3D Locomotion planning experiment. Left: Gazebo arena with Centauro. Obstacle heights were chosen to be rather small to prevent the laser scanner from coping with occlusions. Right: The corresponding occupancy map with the nine chosen goals and one example result path.}
\label{fig:experiment_3d} 
\vspace{-0.3cm}
\end{figure}

Finally, we integrated AVINs into the locomotion planning pipeline of Centauro.
A 3D rotating laser scanner with spherical field-of-view perceived the environment.
Measurements were processed using the SLAM method of Droeschel \etal\cite{mapping}.
Occupancy maps with a resolution of 0.2\,m were generated from these point clouds and were input to AVINs.
Robot perception and control was implemented in C++. 
Communication with AVINs was realized using ROS.
We designed a test arena in the Gazebo simulation environment which contained challenging obstacles of different shape and size, as shown in \cref{fig:experiment_3d}.
We placed nine different goal states in the map and compared our approach to the expert $A^*$ planner.

The results in~\cref{tab:results_3d} indicate that our AVIN planner obtained optimal or close to optimal paths in most cases. 
Even challenging tasks which require the robot to take obstacles between its leg (\eg III and VIII) could be solved.
However, the AVIN planner did not plan successful paths for problems II and V, which required turning actions in narrow sections.
Here, the obtained paths did not reach the goal but ended oscillating between two adjacent poses.
Moreover, AVIN planning times had a considerably smaller distribution than the $A^*$ planner.
Although AVIN planning times do not show a considerable advantage over $A^*$, this experiment demonstrates the application of AVINs to significantly more challenging tasks compared to original VIN applications.

However, the above-described comparison between the 2D and 3D application indicates the large potential of learning-based planners to outperform traditional planners in complex planning tasks.
Nevertheless, increasing hardware requirements and decreasing success rates currently limit this development.
Future work might combine the strengths of AVIN and $A^*$ to create a planner with perfect success rate and low planning times for high-dimensional domains, e.g. by using AVIN to generate an informed heuristic for $A^*$.
Different methods to combine search- and learning-based planners have been proposed in \cite{groshev2018learning}, \cite{klamt2019} and \cite{spies2019bounded}.


\section{Conclusion}

In this paper, we propose an extension to Value Iteration Networks (VINs) to employ multiple levels of abstractions.
While the state resolution gets coarser, additional features compensate the information loss.
Our approach outperforms VINs in terms of result quality on random obstacle grid worlds and is capable of solving considerably larger planning tasks while requiring only a fraction of the graphics memory.
We can further demonstrate that our approach learns to encode important information in its representation which lets it obtain significantly better results in challenging environments compared to Hierarchical VINs.
In addition, we successfully extended our method to plan omnidirectional driving locomotion for the disaster-response robot Centauro while considering its footprint.
Comparing planning times to a search-based planner supports the assumption that learning-based planners are promising to outperform traditional planners in complex tasks.
In summary, we demonstrated how abstraction enables learning-based planners to handle more complex state spaces---increasing their applicability towards real-world motion planning problems.

\section*{Acknowledgments}

This work was supported by the European Union's Horizon 2020 Programme under Grant Agreement 644839 (CENTAURO).


\bibliographystyle{plainnat}
\bibliography{references}

\end{document}

%% file: fig/teaser/teaser.pgf
\begin{tikzpicture}[
 	font=\sffamily\footnotesize,
    every node/.append style={text depth=.2ex},
	l/.style={font=\sffamily\scriptsize},
]


\node[anchor=south west,inner sep=0] (image) at (0,0) {\includegraphics[width=3cm, clip, trim=0 0 0 0]{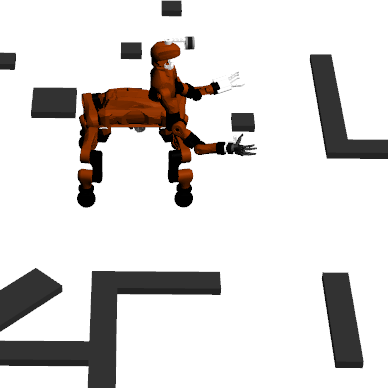}};

\def\squaresize{3.5};
\coordinate (sceneorigin) at (0,0);
\coordinate (gridorigin) at (4.8,0);


\draw (sceneorigin) rectangle ++(\squaresize,\squaresize);
\draw (gridorigin) rectangle ++(\squaresize,\squaresize);
\draw (1,-2.8) rectangle ++(2.5,2);


\draw[-latex]($(sceneorigin) + (\squaresize, 0.5 * \squaresize)$) -- ($(gridorigin) + (0, 0.5 * \squaresize)$);

\def\dist{\squaresize / 16};

\foreach \x in {1,...,15}
	\draw ($(gridorigin) + (\x * \dist,0)$) -- ++(0, \squaresize);

\foreach \y in {1,...,15}
	\draw ($(gridorigin) + (0, \y * \dist,0)$) -- ++( \squaresize,  0);

\fill[black] ($(gridorigin) + (3 * \dist, 3 * \dist)$) rectangle ++(2*\dist, \dist);
\fill[black] ($(gridorigin) + (2 * \dist, 2 * \dist)$) rectangle ++(2*\dist, \dist);
\fill[black] ($(gridorigin) + (1 * \dist, 1 * \dist)$) rectangle ++(2*\dist, \dist);
\fill[black] ($(gridorigin) + (4 * \dist, 10 * \dist)$) rectangle ++(2*\dist, 2*\dist);
\fill[black] ($(gridorigin) + (11 * \dist, 8 * \dist)$) rectangle ++(\dist, 5*\dist);
\fill[black] ($(gridorigin) + (11 * \dist, 8 * \dist)$) rectangle ++(3*\dist, 1*\dist);
\fill[black] ($(gridorigin) + (11 * \dist, 1 * \dist)$) rectangle ++(1*\dist, 3*\dist);
\fill[black] ($(gridorigin) + (6 * \dist, 3 * \dist)$) rectangle ++(3*\dist, 1*\dist);
\fill[black] ($(gridorigin) + (6 * \dist, 0 * \dist)$) rectangle ++(1*\dist, 3*\dist);
\fill[black] ($(gridorigin) + (3 * \dist, 1 * \dist)$) rectangle ++(3*\dist, 1*\dist);
\fill[black] ($(gridorigin) + (1 * \dist, 12 * \dist)$) rectangle ++(\dist, \dist);
\fill[black] ($(gridorigin) + (7 * \dist, 13 * \dist)$) rectangle ++(\dist, \dist);
\fill[black] ($(gridorigin) + (8 * \dist, 15 * \dist)$) rectangle ++(\dist, \dist);
\fill[black] ($(gridorigin) + (9 * \dist, 9 * \dist)$) rectangle ++(\dist, \dist);

\fill[orange] (3,2.5) ellipse (0.1 and 0.07);

\fill[orange] ($(gridorigin) + (13 * \dist, 12 * \dist)$) rectangle ++(\dist, \dist);

\draw[black, thick] ($(gridorigin) + (6 * \dist, 6 * \dist)$) rectangle ++(4 * \dist, 4 * \dist);
\draw[red, thick] ($(gridorigin) + (4 * \dist, 4 * \dist)$) rectangle ++(8 * \dist, 8 * \dist);
\draw[blue, thick] ($(gridorigin) + (0 * \dist, 0 * \dist)$) rectangle ++(16 * \dist, 16 * \dist);

\def\levelsize{1.1}
\def\leveldist{\levelsize / 4};

\definecolor {levelcolor}{RGB}{0, 0, 255};
\coordinate (levelorigin) at (7.075, -1.775);
\draw[levelcolor, thick](levelorigin) rectangle ++(\levelsize, \levelsize);

\coordinate (levelorigin) at (7.15, -1.85);
\draw[levelcolor, thick, fill=white](levelorigin) rectangle ++(\levelsize, \levelsize);

\coordinate (levelorigin) at (7.225, -1.925);
\draw[levelcolor, thick, fill=white](levelorigin) rectangle ++(\levelsize, \levelsize);

\coordinate (levelorigin) at (7.3, -2);
\draw[levelcolor, thick, fill=white](levelorigin) rectangle ++(\levelsize, \levelsize);
\foreach \x in {1,...,3}
	\draw[levelcolor]($(levelorigin) + (\x * \leveldist, 0)$) -- ++(0, \levelsize);
\foreach \y in {1,...,3}
	\draw[levelcolor]($(levelorigin) + (0, \y * \leveldist)$) -- ++(\levelsize, 0);
\draw[levelcolor, dashed, thick]($(levelorigin) + (0, \levelsize)$) -- (gridorigin);
\draw[levelcolor, dashed, thick]($(levelorigin) + (\levelsize, \levelsize)$) -- ($(gridorigin) + (\squaresize, 0)$);
\draw[-latex]($(levelorigin) + (0.5 * \levelsize, -0.35)$) -|++(0, -0.4) -- ++(-4.35,0);
\node[l, align=center, levelcolor] at ($(levelorigin) + (0.5 * \levelsize, -0.2)$) {Level 3};

\definecolor {levelcolor}{RGB}{255, 0, 0};
\coordinate (levelorigin) at (5.925, -1.925);
\draw[levelcolor, thick](levelorigin) rectangle ++(\levelsize, \levelsize);

\coordinate (levelorigin) at (6, -2);
\draw[levelcolor, thick, fill=white](levelorigin) rectangle ++(\levelsize, \levelsize);
\foreach \x in {1,...,3}
	\draw[levelcolor]($(levelorigin) + (\x * \leveldist, 0)$) -- ++(0, \levelsize);
\foreach \y in {1,...,3}
	\draw[levelcolor]($(levelorigin) + (0, \y * \leveldist)$) -- ++(\levelsize, 0);
\draw[levelcolor, dashed, thick]($(levelorigin) + (0, \levelsize)$) -- ($(gridorigin) + (4 * \dist, 4 * \dist)$);
\draw[levelcolor, dashed, thick]($(levelorigin) + (\levelsize, \levelsize)$) -- ($(gridorigin) + (12 * \dist, 4 * \dist)$);
\draw[-latex]($(levelorigin) + (0.5 * \levelsize, -0.35)$) -|++(0, -0.25) -- ++(-3.05,0);
\node[l, align=center, levelcolor] at ($(levelorigin) + (0.5 * \levelsize, -0.2)$) {Level 2};

\definecolor {levelcolor}{RGB}{0, 0, 0};
\coordinate (levelorigin) at (4.7, -2);
\draw[levelcolor, thick, fill=white](levelorigin) rectangle ++(\levelsize, \levelsize);
\foreach \x in {1,...,3}
	\draw[levelcolor]($(levelorigin) + (\x * \leveldist, 0)$) -- ++(0, \levelsize);
\foreach \y in {1,...,3}
	\draw[levelcolor]($(levelorigin) + (0, \y * \leveldist)$) -- ++(\levelsize, 0);
\draw[levelcolor, dashed, thick]($(levelorigin) + (0, \levelsize)$) -- ($(gridorigin) + (6 * \dist, 6 * \dist)$);
\draw[levelcolor, dashed, thick]($(levelorigin) + (\levelsize, \levelsize)$) -- ($(gridorigin) + (10 * \dist, 6 * \dist)$);
\draw[-latex]($(levelorigin) + (0.5 * \levelsize, -0.35)$) -|++(0, -0.1) -- ++(-1.75,0);
\node[l, align=center, levelcolor] at ($(levelorigin) + (0.5 * \levelsize, -0.2)$) {Level 1};

\node[align=center] at (2.25, -1.8) {Value Iteration\\Networks};

\draw[-latex](1, -1.8) -- ++(-1,0);
\node[align=center, l] at (0.6, -1.65) {Plan};

\end{tikzpicture}

%% file: fig/net_architecture_new.pgf
\definecolor{cffffff}{RGB}{255,255,255}
\definecolor{c0000ff}{RGB}{0,0,255}
\definecolor{cff0000}{RGB}{255,0,0}
\definecolor{c00ff00}{RGB}{0,120,0}

\begin{tikzpicture}[y=0.80pt, x=0.80pt, yscale=-1.000000, xscale=1.000000, inner sep=0pt, outer sep=0pt,font=\scriptsize]
\begin{scope}[shift={(0,-62.36221)}]
  \begin{scope}[cm={{0.91065,0.0,0.0,0.91065,(-10.52379,27.87347)}}, shift={(-355,85)}]
    \path[draw=black,fill=cffffff,miter limit=4.00,line width=0.200pt,rounded
      corners=0.0000cm] (536.2592,106.3139) rectangle (576.0107,146.0655);
    \path[draw=black,fill=cffffff,miter limit=4.00,line width=0.200pt,rounded
      corners=0.0000cm] (541.2592,111.3139) rectangle (581.0107,151.0655);
      \path[draw=black,fill=cffffff,miter limit=4.00,line width=0.200pt,rounded
        corners=0.0000cm] (546.2592,116.3139) rectangle (586.0107,156.0655);
      \path[draw=black,line join=miter,line cap=butt,miter limit=4.00,fill
        opacity=0.750,even odd rule,line width=0.100pt] (546.1667,116.2514) --
        (546.1667,156.2204)(551.1318,116.2514) --
        (551.1318,156.2204)(556.0969,116.2514) --
        (556.0969,156.2204)(561.0620,116.2514) --
        (561.0620,156.2204)(566.0271,116.2514) --
        (566.0271,156.2204)(570.9922,116.2514) --
        (570.9922,156.2204)(575.9573,116.2514) --
        (575.9573,156.2204)(580.9224,116.2514) --
        (580.9224,156.2204)(585.8875,116.2514) --
        (585.8875,156.2204)(546.1667,116.2514) --
        (586.1357,116.2514)(546.1667,121.2165) --
        (586.1357,121.2165)(546.1667,126.1816) --
        (586.1357,126.1816)(546.1667,131.1467) --
        (586.1357,131.1467)(546.1667,136.1118) --
        (586.1357,136.1118)(546.1667,141.0769) --
        (586.1357,141.0769)(546.1667,146.0420) --
        (586.1357,146.0420)(546.1667,151.0071) --
        (586.1357,151.0071)(546.1667,155.9721) -- (586.1357,155.9721);
  \path[fill=c00ff00,line join=miter,line cap=butt,line width=0.800pt]
    (577,109) node[above right] (text6765-3-43-8-1-1)
    {1\,@\,8$\times$8};
  \path[fill=c00ff00,line join=miter,line cap=butt,line width=0.800pt]
   (586,118)node[above right] (text6765-3-43-8-1-3)
    {1\,@\,8$\times$8};
  \path[fill=c00ff00,line join=miter,line cap=butt,line width=0.800pt]
    (587,127) node[above right,] (text6765-3-43-8-1-8)
    {1\,@\,8$\times$8};
  \end{scope}

  \path[draw=black,fill=cffffff,miter limit=4.00,line width=0.182pt,rounded
    corners=0.0000cm] (570.9056,101.3827) rectangle (609.0073,266.8450);
  \path[draw=black,fill=cffffff,miter limit=4.00,line width=0.182pt,rounded
    corners=0.0000cm] (238.0143,143.9465) rectangle (274.3303,239.7659);
  \path[draw=black,fill=cffffff,miter limit=4.00,line width=0.182pt,rounded
    corners=0.0000cm] (85.3702,101.3827) rectangle (136.4391,266.8450);
  \path[draw=black,fill=cffffff,miter limit=4.00,line width=0.182pt,rounded
    corners=0.0000cm] (404.4600,101.3827) rectangle (440.7760,266.8450);

\begin{scope}[shift={(-2,-5)}]
  \begin{scope}[cm={{0.91065,0.0,0.0,0.91065,(-6.51085,-388.84916)}},miter limit=4.00,line width=0.200pt]
    \begin{scope}[cm={{0.5,0.0,0.0,0.5,(26.0228,220.84976)}},miter limit=4.00,line width=0.400pt]
      \path[draw=black,fill=cffffff,miter limit=4.00,line width=0.400pt]
        (300.2484,752.1138) -- (300.2484,752.1138) -- (300.2484,672.6107) --
        (379.7516,672.6107) -- (379.7516,672.6107);
      \path[draw=black,fill=cffffff,miter limit=4.00,line width=0.400pt]
        (324.5164,767.1138) -- (315.2484,767.1138) -- (315.2484,687.6107) --
        (394.7515,687.6107) -- (394.7515,697.4637);
      \path[draw=black,line join=miter,line cap=butt,miter limit=4.00,even odd
        rule,line width=0.400pt] (315.2484,767.1138) -- (300.2484,752.1138);
      \path[draw=black,line join=miter,line cap=butt,miter limit=4.00,even odd
        rule,line width=0.400pt] (315.2484,687.6107) -- (300.2484,672.6107);
      \path[draw=black,line join=miter,line cap=butt,miter limit=4.00,even odd
        rule,line width=0.400pt] (394.7516,687.6107) -- (379.7516,672.6107);
    \end{scope}
  \end{scope}
  \begin{scope}[cm={{0.91065,0.0,0.0,0.91065,(-7.43979,-271.92989)}}]
    \path[draw=black,fill=cffffff,miter limit=4.00,line width=0.200pt]
      (189.1997,480.1774) -- (189.1997,480.1774) -- (189.1997,440.4258) --
      (228.9512,440.4258) -- (228.9512,440.4258);
    \path[draw=black,fill=cffffff,miter limit=4.00,line width=0.200pt]
      (194.1997,445.4258) -- (233.9513,445.4258) -- (233.9513,485.1774) --
      (194.1997,485.1774) -- cycle;
    \path[draw=black,line join=miter,line cap=butt,miter limit=4.00,even odd
      rule,line width=0.200pt] (194.1997,485.1774) -- (189.1997,480.1774);
    \path[draw=black,line join=miter,line cap=butt,miter limit=4.00,even odd
      rule,line width=0.200pt] (194.1997,445.4258) -- (189.1997,440.4259);
    \path[draw=black,line join=miter,line cap=butt,miter limit=4.00,even odd
      rule,line width=0.200pt] (233.9513,445.4258) -- (228.9512,440.4259);
  \end{scope}
  \path[draw=black,fill=cffffff,miter limit=4.00,line width=0.182pt]
    (173.5361,173.6873) -- (173.5361,173.6873) -- (173.5361,137.4875) --
    (209.7359,137.4875) -- (209.7359,137.4875);
  \path[draw=black,line join=miter,line cap=butt,miter limit=4.00,even odd
    rule,line width=0.182pt] (175.8127,175.9639) -- (173.5361,173.6873);
  \path[draw=black,line join=miter,line cap=butt,miter limit=4.00,even odd
    rule,line width=0.182pt] (175.8127,139.7641) -- (173.5361,137.4875);
  \path[draw=black,line join=miter,line cap=butt,miter limit=4.00,even odd
    rule,line width=0.182pt] (212.0125,139.7641) -- (209.7359,137.4875);
  \path[draw=black,fill=cffffff,miter limit=4.00,line width=0.182pt]
    (175.8126,139.7923) -- (211.9843,139.7923) -- (211.9843,175.9640) --
    (175.8126,175.9640) -- cycle;
  \path[draw=black,line join=miter,line cap=butt,miter limit=4.00,fill
    opacity=0.750,even odd rule,line width=0.091pt] (175.6952,139.9015) --
    (175.6952,176.2993)(180.2167,139.9015) --
    (180.2167,176.2993)(184.7381,139.9015) --
    (184.7381,176.2993)(189.2596,139.9015) --
    (189.2596,176.2993)(193.7811,139.9015) --
    (193.7811,176.2993)(198.3025,139.9015) --
    (198.3025,176.2993)(202.8240,139.9015) --
    (202.8240,176.2993)(207.3455,139.9015) --
    (207.3455,176.2993)(211.8669,139.9015) --
    (211.8669,176.2993)(175.6952,139.9015) --
    (212.0930,139.9015)(175.6952,144.4230) --
    (212.0930,144.4230)(175.6952,148.9445) --
    (212.0930,148.9445)(175.6952,153.4659) --
    (212.0930,153.4659)(175.6952,157.9874) --
    (212.0930,157.9874)(175.6952,162.5088) --
    (212.0930,162.5088)(175.6952,167.0303) --
    (212.0930,167.0303)(175.6952,171.5518) --
    (212.0930,171.5518)(175.6952,176.0732) -- (212.0930,176.0732);
  \path[fill=black,line join=miter,line cap=butt,line width=0.800pt]
    (195,124) node[above right,c00ff00] (text6765-3) {6\,@\,8$\times$8};
  \path[fill=black,line join=miter,line cap=butt,line width=0.800pt]
    (230,124) node[above right, purple] (text6765-3) {10\,@\,8$\times$8};
  \path[fill=black,line join=miter,line cap=butt,line width=0.800pt]
    (210,135) node[above right,c00ff00] (text6765-3-5) {2\,@\,8$\times$8};
  \path[fill=black,line join=miter,line cap=butt,line width=0.800pt]
    (245,135) node[above right,purple] (text6765-3-5) {5\,@\,8$\times$8};
  \path[fill=black,line join=miter,line cap=butt,line width=0.800pt]
    (215,145) node[above right,black] (text6765-3-4) {1\,@\,8$\times$8};
\end{scope}

  \begin{scope}[cm={{0.91065,0.0,0.0,0.91065,(2.24676,25.54826)}}]
  \begin{scope}[shift={(2,-15)}]
    \begin{scope}[cm={{0.91065,0.0,0.0,0.91065,(2.24676,25.54826)}}]
      \path[draw=black,fill=cffffff,miter limit=4.00,line width=0.200pt,rounded
        corners=0.0000cm] (29.6416,113.9280) rectangle (69.3623,153.6487);
      \path[draw=black,line join=miter,line cap=butt,fill opacity=0.750,even odd
        rule,line width=0.100pt] (29.5183,113.8048) --
        (29.5183,153.7738)(32.0183,113.8048) -- (32.0183,153.7738)(34.5183,113.8048)
        -- (34.5183,153.7738)(37.0183,113.8048) --
        (37.0183,153.7738)(39.5183,113.8048) -- (39.5183,153.7738)(42.0183,113.8048)
        -- (42.0183,153.7738)(44.5183,113.8048) --
        (44.5183,153.7738)(47.0183,113.8048) -- (47.0183,153.7738)(49.5183,113.8048)
        -- (49.5183,153.7738)(52.0183,113.8048) --
        (52.0183,153.7738)(54.5183,113.8048) -- (54.5183,153.7738)(57.0183,113.8048)
        -- (57.0183,153.7738)(59.5183,113.8048) --
        (59.5183,153.7738)(62.0183,113.8048) -- (62.0183,153.7738)(64.5183,113.8048)
        -- (64.5183,153.7738)(67.0183,113.8048) --
        (67.0183,153.7738)(29.5183,113.8048) -- (69.4873,113.8048)(29.5183,116.3048)
        -- (69.4873,116.3048)(29.5183,118.8048) --
        (69.4873,118.8048)(29.5183,121.3048) -- (69.4873,121.3048)(29.5183,123.8048)
        -- (69.4873,123.8048)(29.5183,126.3048) --
        (69.4873,126.3048)(29.5183,128.8048) -- (69.4873,128.8048)(29.5183,131.3048)
        -- (69.4873,131.3048)(29.5183,133.8048) --
        (69.4873,133.8048)(29.5183,136.3048) -- (69.4873,136.3048)(29.5183,138.8048)
        -- (69.4873,138.8048)(29.5183,141.3048) --
        (69.4873,141.3048)(29.5183,143.8048) -- (69.4873,143.8048)(29.5183,146.3048)
        -- (69.4873,146.3048)(29.5183,148.8048) --
        (69.4873,148.8048)(29.5183,151.3048) -- (69.4873,151.3048);
    \end{scope}
    \path[draw=black,fill=black,miter limit=4.00,line width=0.087pt,rounded
      corners=0.0000cm] (35.9575,151.9510) rectangle (38.1257,154.1192);
    \path[draw=black,fill=black,miter limit=4.00,line width=0.087pt,rounded
      corners=0.0000cm] (35.9575,142.8445) rectangle (38.1257,145.0127);
    \path[draw=black,fill=black,miter limit=4.00,line width=0.087pt,rounded
      corners=0.0000cm] (56.5556,140.5679) rectangle (58.7238,142.7361);
    \path[draw=black,fill=black,miter limit=4.00,line width=0.087pt,rounded
      corners=0.0000cm] (61.0004,131.4614) rectangle (63.1687,133.6296);
    \path[draw=black,fill=black,miter limit=4.00,line width=0.087pt,rounded
      corners=0.0000cm] (58.8322,142.9529) rectangle (61.0004,145.1211);
    \path[draw=black,fill=black,miter limit=4.00,line width=0.087pt,rounded
      corners=0.0000cm] (56.4472,158.8893) rectangle (58.6154,161.0575);
    \path[draw=black,fill=black,miter limit=4.00,line width=0.182pt,rounded
      corners=0.0000cm] (31.4043,156.5043) rectangle (35.9575,161.0575);
    \path[draw=black,fill=black,miter limit=4.00,line width=0.182pt,rounded
      corners=0.0000cm] (58.7238,138.2912) rectangle (63.2771,142.8445);
    \path[draw=black,fill=black,miter limit=4.00,line width=0.182pt,rounded
      corners=0.0000cm] (31.4043,131.4614) rectangle (35.9575,136.0146);
    \path[draw=black,fill=black,miter limit=4.00,line width=0.087pt,rounded
      corners=0.0000cm] (54.1706,158.8893) rectangle (56.3388,161.0575);
    \path[draw=black,fill=black,miter limit=4.00,line width=0.087pt,rounded
      corners=0.0000cm] (47.3407,149.6744) rectangle (49.5089,151.8426);
    \path[draw=black,fill=black,miter limit=4.00,line width=0.087pt,rounded
      corners=0.0000cm] (52.0023,158.8893) rectangle (54.1706,161.0575);
    \path[draw=black,fill=black,miter limit=4.00,line width=0.087pt,rounded
      corners=0.0000cm] (56.5556,156.6127) rectangle (58.7238,158.7809);
    \path[draw=black,fill=black,miter limit=4.00,line width=0.087pt,rounded
      corners=0.0000cm] (47.3407,133.7380) rectangle (49.5089,135.9062);
    \path[draw=black,fill=black,miter limit=4.00,line width=0.087pt,rounded
      corners=0.0000cm] (49.6173,133.8464) rectangle (51.7855,136.0146);
    \path[draw=black,fill=black,miter limit=4.00,line width=0.182pt,rounded
      corners=0.0000cm] (33.6809,133.7380) rectangle (38.2341,138.2912);
    \path[draw=black,fill=black,miter limit=4.00,line width=0.182pt,rounded
      corners=0.0000cm] (49.6173,161.0575) rectangle (54.1706,165.6108);
    \path[draw=black,fill=black,miter limit=4.00,line width=0.087pt,rounded
      corners=0.0000cm] (36.0659,149.6744) rectangle (38.2342,151.8426);
    \path[draw=black,fill=black,miter limit=4.00,line width=0.087pt,rounded
      corners=0.0000cm] (38.2342,142.8445) rectangle (40.4024,145.0127);
    \path[draw=black,fill=black,miter limit=4.00,line width=0.087pt,rounded
      corners=0.0000cm] (35.9575,147.3978) rectangle (38.1257,149.5660);
  \end{scope}
 \end{scope}

  \begin{scope}[cm={{0.91065,0.0,0.0,0.91065,(2.24677,8.92456)}}]
    \path[draw=black,fill=cffffff,miter limit=4.00,line width=0.200pt,rounded
      corners=0.0000cm] (29.6416,209.1372) rectangle (69.3623,248.8580);
    \path[draw=black,line join=miter,line cap=butt,fill opacity=0.750,even odd
      rule,line width=0.100pt] (29.5183,209.0747) --
      (29.5183,249.0437)(32.0183,209.0747) -- (32.0183,249.0437)(34.5183,209.0747)
      -- (34.5183,249.0437)(37.0183,209.0747) --
      (37.0183,249.0437)(39.5183,209.0747) -- (39.5183,249.0437)(42.0183,209.0747)
      -- (42.0183,249.0437)(44.5183,209.0747) --
      (44.5183,249.0437)(47.0183,209.0747) -- (47.0183,249.0437)(49.5183,209.0747)
      -- (49.5183,249.0437)(52.0183,209.0747) --
      (52.0183,249.0437)(54.5183,209.0747) -- (54.5183,249.0437)(57.0183,209.0747)
      -- (57.0183,249.0437)(59.5183,209.0747) --
      (59.5183,249.0437)(62.0183,209.0747) -- (62.0183,249.0437)(64.5183,209.0747)
      -- (64.5183,249.0437)(67.0183,209.0747) --
      (67.0183,249.0437)(29.5183,209.0747) -- (69.4873,209.0747)(29.5183,211.5747)
      -- (69.4873,211.5747)(29.5183,214.0747) --
      (69.4873,214.0747)(29.5183,216.5747) -- (69.4873,216.5747)(29.5183,219.0747)
      -- (69.4873,219.0747)(29.5183,221.5747) --
      (69.4873,221.5747)(29.5183,224.0747) -- (69.4873,224.0747)(29.5183,226.5747)
      -- (69.4873,226.5747)(29.5183,229.0747) --
      (69.4873,229.0747)(29.5183,231.5747) -- (69.4873,231.5747)(29.5183,234.0747)
      -- (69.4873,234.0747)(29.5183,236.5747) --
      (69.4873,236.5747)(29.5183,239.0747) -- (69.4873,239.0747)(29.5183,241.5747)
      -- (69.4873,241.5747)(29.5183,244.0747) --
      (69.4873,244.0747)(29.5183,246.5747) -- (69.4873,246.5747);
    \path[draw=black,fill=black,miter limit=4.00,line width=0.095pt,rounded
      corners=0.0000cm] (59.5443,221.5394) rectangle (61.9252,223.9204);
  \end{scope}
  \begin{scope}[cm={{0.91065,0.0,0.0,0.91065,(-13.46747,25.54826)}}]
    \path[draw=black,fill=cffffff,miter limit=4.00,line width=0.200pt,rounded
      corners=0.0000cm] (703.5293,95.2944) rectangle (728.0043,252.9521);
    \path[draw=black,fill=cffffff,miter limit=4.00,line width=0.200pt,rounded
      corners=0.0000cm] (710.7452,238.9750) rectangle (720.7452,248.9750);
    \path[draw=black,fill=cffffff,miter limit=4.00,line width=0.200pt,rounded
      corners=0.0000cm] (710.7452,220.5966) rectangle (720.7452,230.5966);
    \path[draw=black,fill=cffffff,miter limit=4.00,line width=0.200pt,rounded
      corners=0.0000cm] (710.7452,202.9190) rectangle (720.7452,212.9190);
    \path[draw=black,fill=cffffff,miter limit=4.00,line width=0.200pt,rounded
      corners=0.0000cm] (710.7452,186.7565) rectangle (720.7452,196.7565);
    \path[draw=black,fill=cffffff,miter limit=4.00,line width=0.200pt,rounded
      corners=0.0000cm] (710.7452,168.0687) rectangle (720.7452,178.0687);
    \path[draw=black,fill=cffffff,miter limit=4.00,line width=0.200pt,rounded
      corners=0.0000cm] (710.7452,152.4113) rectangle (720.7452,162.4113);
    \path[draw=black,fill=black,miter limit=4.00,line width=0.200pt,rounded
      corners=0.0000cm] (710.7452,135.7438) rectangle (720.7452,145.7438);
    \path[draw=black,fill=cffffff,miter limit=4.00,line width=0.200pt,rounded
      corners=0.0000cm] (710.7452,116.0459) rectangle (720.7452,126.0459);
    \path[draw=black,fill=cffffff,miter limit=4.00,line width=0.200pt,rounded
      corners=0.0000cm] (710.7452,98.3682) rectangle (720.7452,108.3682);
  \end{scope}
  \path[fill=black,line join=miter,line cap=butt,line width=0.800pt]
    (255.8372,190.1337) node[align=center] (text5885-4-7) {Reward \\ Module};
  \path[fill=black,line join=miter,line cap=butt,line width=0.800pt]
    (422.5646,186.6600) node[align=center] (text5885-4-6-4) {VI\\Module};
  \begin{scope}[cm={{0.91065,0.0,0.0,0.91065,(53.59336,-659.42204)}},miter limit=4.00,line 
width=0.200pt]
    \path[draw=purple,fill=cffffff,miter limit=4.00,line width=0.200pt]
      (266.1881,973.0662) -- (266.1881,973.0662) -- (266.1881,933.3147) --
      (305.9396,933.3147) -- (305.9396,933.3147);
    \path[draw=purple,fill=cffffff,miter limit=4.00,line width=0.200pt]
      (273.7191,978.0662) -- (271.1881,978.0662) -- (271.1881,938.3147) --
      (310.9396,938.3147) -- (310.9396,940.7448);
    \path[draw=purple,line join=miter,line cap=butt,miter limit=4.00,even odd
      rule,line width=0.200pt] (271.1881,938.3147) -- (266.1881,933.3147);
    \path[draw=purple,line join=miter,line cap=butt,miter limit=4.00,even odd
      rule,line width=0.200pt] (310.9396,938.3147) -- (305.9396,933.3147);
    \path[draw=purple,line join=miter,line cap=butt,miter limit=4.00,even odd
      rule,line width=0.200pt] (271.1881,978.0662) -- (266.1881,973.0662);
  \end{scope}
  \begin{scope}[cm={{0.91065,0.0,0.0,0.91065,(60.17632,-652.56766)}},miter limit=4.00,line 
width=0.200pt]
    \path[draw=purple,fill=cffffff,miter limit=4.00,line width=0.200pt]
      (266.1881,973.0662) -- (266.1881,973.0662) -- (266.1881,933.3147) --
      (305.9396,933.3147) -- (305.9396,933.3147);
    \path[draw=purple,fill=cffffff,miter limit=4.00,line width=0.200pt]
      (273.2611,978.0662) -- (271.1881,978.0662) -- (271.1881,938.3147) --
      (310.9396,938.3147) -- (310.9396,940.2091);
    \path[draw=purple,line join=miter,line cap=butt,miter limit=4.00,even odd
      rule,line width=0.200pt] (271.1881,938.3147) -- (266.1881,933.3147);
    \path[draw=purple,line join=miter,line cap=butt,miter limit=4.00,even odd
      rule,line width=0.200pt] (310.9396,938.3147) -- (305.9396,933.3147);
    \path[draw=purple,line join=miter,line cap=butt,miter limit=4.00,even odd
      rule,line width=0.200pt] (271.1881,978.0662) -- (266.1881,973.0662);
  \end{scope}
  \path[draw=purple,fill=cffffff,miter limit=4.00,line width=0.182pt]
    (309.1638,239.9229) -- (309.1638,239.9229) -- (309.1638,203.7231) --
    (345.3636,203.7231) -- (345.3636,203.7231);
  \path[draw=purple,fill=cffffff,miter limit=4.00,line width=0.182pt]
    (313.7171,208.2764) -- (349.9169,208.2764) -- (349.9169,244.4762) --
    (313.7171,244.4762) -- cycle;
  \path[draw=purple,line join=miter,line cap=butt,miter limit=4.00,even odd
    rule,line width=0.182pt] (313.7171,208.2764) -- (309.1638,203.7231);
  \path[draw=purple,line join=miter,line cap=butt,miter limit=4.00,even odd
    rule,line width=0.182pt] (349.9169,208.2764) -- (345.3636,203.7231);
  \path[draw=purple,line join=miter,line cap=butt,miter limit=4.00,even odd
    rule,line width=0.182pt] (313.7171,244.4761) -- (309.1638,239.9229);
  \begin{scope}[cm={{0.91065,0.0,0.0,0.91065,(-22.914,27.30719)}}]
    \begin{scope}[shift={(183.92435,-456.98759)},draw=c00ff00,miter limit=4.00,line width=0.200pt]
      \begin{scope}[cm={{0.5,0.0,0.0,0.5,(26.0228,220.84976)}},draw=c00ff00,miter limit=4.00,line 
width=0.400pt]
        \path[draw=c00ff00,fill=cffffff,miter limit=4.00,line width=0.400pt]
          (300.2484,752.1138) -- (300.2484,752.1138) -- (300.2484,672.6107) --
          (379.7516,672.6107) -- (379.7516,672.6107);
        \path[draw=c00ff00,fill=cffffff,miter limit=4.00,line width=0.400pt]
          (324.5164,767.1138) -- (315.2484,767.1138) -- (315.2484,687.6107) --
          (394.7515,687.6107) -- (394.7515,697.4637);
        \path[draw=c00ff00,line join=miter,line cap=butt,miter limit=4.00,even odd
          rule,line width=0.400pt] (315.2484,767.1138) -- (300.2484,752.1138);
        \path[draw=c00ff00,line join=miter,line cap=butt,miter limit=4.00,even odd
          rule,line width=0.400pt] (315.2484,687.6107) -- (300.2484,672.6107);
        \path[draw=c00ff00,line join=miter,line cap=butt,miter limit=4.00,even odd
          rule,line width=0.400pt] (394.7516,687.6107) -- (379.7516,672.6107);
      \end{scope}
    \end{scope}
    \begin{scope}[shift={(182.90427,-328.59677)},draw=c00ff00]
      \path[draw=c00ff00,fill=cffffff,miter limit=4.00,line width=0.200pt]
        (189.1997,480.1774) -- (189.1997,480.1774) -- (189.1997,440.4258) --
        (228.9512,440.4258) -- (228.9512,440.4258);
      \path[draw=c00ff00,fill=cffffff,miter limit=4.00,line width=0.200pt]
        (194.1997,445.4258) -- (233.9513,445.4258) -- (233.9513,485.1774) --
        (194.1997,485.1774) -- cycle;
      \path[draw=c00ff00,line join=miter,line cap=butt,miter limit=4.00,even odd
        rule,line width=0.200pt] (194.1997,485.1774) -- (189.1997,480.1774);
      \path[draw=c00ff00,line join=miter,line cap=butt,miter limit=4.00,even odd
        rule,line width=0.200pt] (194.1997,445.4258) -- (189.1997,440.4259);
      \path[draw=c00ff00,line join=miter,line cap=butt,miter limit=4.00,even odd
        rule,line width=0.200pt] (233.9513,445.4258) -- (228.9512,440.4259);
    \end{scope}
      \path[draw=c00ff00,fill=cffffff,miter limit=4.00,line width=0.200pt]
        (381.6366,160.7422) -- (381.6366,160.7422) -- (381.6366,120.9907) --
        (421.3881,120.9907) -- (421.3881,120.9907);
      \path[draw=c00ff00,line join=miter,line cap=butt,miter limit=4.00,even odd
        rule,line width=0.200pt] (384.1366,163.2422) -- (381.6366,160.7422);
      \path[draw=c00ff00,line join=miter,line cap=butt,miter limit=4.00,even odd
        rule,line width=0.200pt] (384.1366,123.4907) -- (381.6366,120.9907);
      \path[draw=c00ff00,line join=miter,line cap=butt,miter limit=4.00,even odd
        rule,line width=0.200pt] (423.8882,123.4907) -- (421.3881,120.9907);
      \path[draw=c00ff00,fill=cffffff,miter limit=4.00,line width=0.200pt]
        (384.1365,123.5216) -- (423.8572,123.5216) -- (423.8572,163.2423) --
        (384.1365,163.2423) -- cycle;
      \path[draw=c00ff00,line join=miter,line cap=butt,miter limit=4.00,fill
        opacity=0.750,even odd rule,line width=0.100pt] (384.0076,123.6416) --
        (384.0076,163.6105)(388.9727,123.6416) --
        (388.9727,163.6105)(393.9378,123.6416) --
        (393.9378,163.6105)(398.9029,123.6416) --
        (398.9029,163.6105)(403.8679,123.6416) --
        (403.8679,163.6105)(408.8330,123.6416) --
        (408.8330,163.6105)(413.7981,123.6416) --
        (413.7981,163.6105)(418.7632,123.6416) --
        (418.7632,163.6105)(423.7283,123.6416) --
        (423.7283,163.6105)(384.0076,123.6416) --
        (423.9766,123.6416)(384.0076,128.6066) --
        (423.9766,128.6066)(384.0076,133.5717) --
        (423.9766,133.5717)(384.0076,138.5368) --
        (423.9766,138.5368)(384.0076,143.5019) --
        (423.9766,143.5019)(384.0076,148.4670) --
        (423.9766,148.4670)(384.0076,153.4321) --
        (423.9766,153.4321)(384.0076,158.3972) --
        (423.9766,158.3972)(384.0076,163.3623) -- (423.9766,163.3623);
  \end{scope}
  \begin{scope}[cm={{0.91065,0.0,0.0,0.91065,(-10.52379,27.87347)}}]
    \path[draw=c00ff00,fill=cffffff,miter limit=4.00,line width=0.200pt,rounded
      corners=0.0000cm] (536.2592,106.3139) rectangle (576.0107,146.0655);
    \path[draw=c00ff00,fill=cffffff,miter limit=4.00,line width=0.200pt,rounded
      corners=0.0000cm] (541.2592,111.3139) rectangle (581.0107,151.0655);
      \path[draw=c00ff00,fill=cffffff,miter limit=4.00,line width=0.200pt,rounded
        corners=0.0000cm] (546.2592,116.3139) rectangle (586.0107,156.0655);
      \path[draw=c00ff00,line join=miter,line cap=butt,miter limit=4.00,fill
        opacity=0.750,even odd rule,line width=0.100pt] (546.1667,116.2514) --
        (546.1667,156.2204)(551.1318,116.2514) --
        (551.1318,156.2204)(556.0969,116.2514) --
        (556.0969,156.2204)(561.0620,116.2514) --
        (561.0620,156.2204)(566.0271,116.2514) --
        (566.0271,156.2204)(570.9922,116.2514) --
        (570.9922,156.2204)(575.9573,116.2514) --
        (575.9573,156.2204)(580.9224,116.2514) --
        (580.9224,156.2204)(585.8875,116.2514) --
        (585.8875,156.2204)(546.1667,116.2514) --
        (586.1357,116.2514)(546.1667,121.2165) --
        (586.1357,121.2165)(546.1667,126.1816) --
        (586.1357,126.1816)(546.1667,131.1467) --
        (586.1357,131.1467)(546.1667,136.1118) --
        (586.1357,136.1118)(546.1667,141.0769) --
        (586.1357,141.0769)(546.1667,146.0420) --
        (586.1357,146.0420)(546.1667,151.0071) --
        (586.1357,151.0071)(546.1667,155.9721) -- (586.1357,155.9721);
  \end{scope}
  \path[draw=c00ff00,-Latex,line join=miter,line cap=butt,miter limit=4.00,even odd
    rule,line width=0.182pt] (281.6560,172.8911) -- (287.7274,172.8911) --
    (287.7274,147.3552) -- (299.3291,147.3552);
  \path[draw=c00ff00,-Latex,line join=miter,line cap=butt,miter limit=4.00,even odd
    rule,line width=0.182pt] (368.3142,147.3552) -- (396.7197,147.3552);
  \path[draw=c00ff00,-Latex,line join=miter,line cap=butt,miter limit=4.00,even odd
    rule,line width=0.182pt] (448.1019,147.3552) -- (465.7747,147.3552);
  \path[draw=c00ff00,-Latex,line join=miter,line cap=butt,miter limit=4.00,even odd
    rule,line width=0.182pt] (534.7599,147.3552) -- (563.1654,147.3552);
  \path[draw=purple,-Latex,line join=miter,line cap=butt,miter limit=4.00,even odd
    rule,line width=0.182pt] (281.6562,217.4892) -- (293.2576,217.4892);
  \path[draw=purple,-Latex,line join=miter,line cap=butt,miter limit=4.00,even odd
    rule,line width=0.182pt] (357.5999,217.4892) -- (396.7197,217.4892);
  \path[draw=black,-Latex,line join=miter,line cap=butt,miter limit=4.00,even odd
    rule,line width=0.182pt] (69.3827,217.5073) -- (82.0558,217.5073);
  \path[draw=black,-Latex,line join=miter,line cap=butt,miter limit=4.00,even odd
    rule,line width=0.182pt] (138.9820,225) -- (149.5118,225);
  \path[draw=black,-Latex,line join=miter,line cap=butt,miter limit=4.00,even odd
    rule,line width=0.182pt] (215.9558,225) -- (233.6287,225);
  \path[draw=black,-Latex,line join=miter,line cap=butt,miter limit=4.00,even odd
    rule,line width=0.182pt] (69.3827,147.3552) -- (82.0558,147.3552);
  \path[draw=black,-Latex,line join=miter,line cap=butt,miter limit=4.00,even odd
    rule,line width=0.182pt] (138.9820,147.3552) -- (149.5118,147.3552);
  \path[draw=black,-Latex,line join=miter,line cap=butt,miter limit=4.00,even odd
    rule,line width=0.182pt] (215.9558,163.7837) -- (233.6287,163.7837);
  \path[draw=black,-Latex,line join=miter,line cap=butt,miter limit=4.00,even odd
    rule,line width=0.182pt] (610.3113,184.1138) -- (623.6983,184.1138);
  \path[draw=purple,-Latex,line join=miter,line cap=butt,miter limit=4.00,even odd
    rule,line width=0.182pt] (72.1718,279.4294) -- (554.1665,279.4294) --
    (554.1665,258.6146) -- (564.8992,258.6146);
  \begin{scope}[cm={{0.91065,0.0,0.0,0.91065,(219.32692,-659.36598)}},miter limit=4.00,line 
width=0.200pt]
    \path[draw=purple,fill=cffffff,miter limit=4.00,line width=0.200pt]
      (266.1881,973.0662) -- (266.1881,973.0662) -- (266.1881,933.3147) --
      (305.9396,933.3147) -- (305.9396,933.3147);
    \path[draw=purple,fill=cffffff,miter limit=4.00,line width=0.200pt]
      (273.7191,978.0662) -- (271.1881,978.0662) -- (271.1881,938.3147) --
      (310.9396,938.3147) -- (310.9396,940.7448);
    \path[draw=purple,line join=miter,line cap=butt,miter limit=4.00,even odd
      rule,line width=0.200pt] (271.1881,938.3147) -- (266.1881,933.3147);
    \path[draw=purple,line join=miter,line cap=butt,miter limit=4.00,even odd
      rule,line width=0.200pt] (310.9396,938.3147) -- (305.9396,933.3147);
    \path[draw=purple,line join=miter,line cap=butt,miter limit=4.00,even odd
      rule,line width=0.200pt] (271.1881,978.0662) -- (266.1881,973.0662);
  \end{scope}
  \begin{scope}[cm={{0.91065,0.0,0.0,0.91065,(225.90988,-652.51159)}},miter limit=4.00,line 
width=0.200pt]
    \path[draw=purple,fill=cffffff,miter limit=4.00,line width=0.200pt]
      (266.1881,973.0662) -- (266.1881,973.0662) -- (266.1881,933.3147) --
      (305.9396,933.3147) -- (305.9396,933.3147);
    \path[draw=purple,fill=cffffff,miter limit=4.00,line width=0.200pt]
      (273.2611,978.0662) -- (271.1881,978.0662) -- (271.1881,938.3147) --
      (310.9396,938.3147) -- (310.9396,940.2091);
    \path[draw=purple,line join=miter,line cap=butt,miter limit=4.00,even odd
      rule,line width=0.200pt] (271.1881,938.3147) -- (266.1881,933.3147);
    \path[draw=purple,line join=miter,line cap=butt,miter limit=4.00,even odd
      rule,line width=0.200pt] (310.9396,938.3147) -- (305.9396,933.3147);
    \path[draw=purple,line join=miter,line cap=butt,miter limit=4.00,even odd
      rule,line width=0.200pt] (271.1881,978.0662) -- (266.1881,973.0662);
  \end{scope}
  \path[draw=purple,fill=cffffff,miter limit=4.00,line width=0.182pt]
    (474.8974,239.9790) -- (474.8974,239.9790) -- (474.8974,203.7792) --
    (511.0972,203.7792) -- (511.0972,203.7792);
  \path[draw=purple,fill=cffffff,miter limit=4.00,line width=0.182pt]
    (479.4506,208.3324) -- (515.6504,208.3324) -- (515.6504,244.5323) --
    (479.4506,244.5323) -- cycle;
  \path[draw=purple,line join=miter,line cap=butt,miter limit=4.00,even odd
    rule,line width=0.182pt] (479.4506,208.3324) -- (474.8974,203.7792);
  \path[draw=purple,line join=miter,line cap=butt,miter limit=4.00,even odd
    rule,line width=0.182pt] (515.6504,208.3324) -- (511.0972,203.7792);
  \path[draw=purple,line join=miter,line cap=butt,miter limit=4.00,even odd
    rule,line width=0.182pt] (479.4506,244.5322) -- (474.8974,239.9790);
  \path[draw=purple,-Latex,line join=miter,line cap=butt,miter limit=4.00,even odd
    rule,line width=0.182pt] (447.3898,217.5453) -- (458.9912,217.5453);
  \path[draw=purple,-Latex,line join=miter,line cap=butt,miter limit=4.00,even odd
    rule,line width=0.182pt] (523.3335,217.5453) -- (562.4533,217.5453);
  \path[fill=black,line join=miter,line cap=butt,line width=0.800pt]
    (47.2931,240) node[align=center,below] (text5885-4) {Goal \\ map};
  \path[fill=black,line join=miter,line cap=butt,line width=0.800pt]
    (47.1063,170) node[align=center, below] (text5885) {Occupancy \\ map};
  \path[fill=black,line join=miter,line cap=butt,line width=0.800pt]
    (47.2996,125.0143) node[align=center] (text6765) {$32\times32$};
  \path[fill=black,line join=miter,line cap=butt,line width=0.800pt]
    (46.7754,195) node[align=center] (text6765-5) {$32\times32$};
  \path[fill=purple,line join=miter,line cap=butt,line width=0.800pt]
    (50.2569,277.3048) node[align=center, color=purple] (text5885-4-6-6-4) {Start \\orientation};
  \path[fill=black,line join=miter,line cap=butt,line width=0.800pt]
    (111,186.6600) node[align=center] (text5885-4-4) {Abstraction \\Module};
  \path[fill=black,line join=miter,line cap=butt,line width=0.800pt]
    (190,183) node[align=center] (text5885-4-3) {Abstract environment\\ maps};
  \path[fill=black,line join=miter,line cap=butt,line width=0.800pt]
    (185,250) node[align=center, below] (text5885-4-3) {Abstract goal \\maps};

  \path[fill=purple,line join=miter,line cap=butt,line width=0.800pt]
    (331.8298,189.4724) node[above right,color=purple] (text6765-3-43)
    {10\,@\,8$\times$8$\times$4};
  \path[fill=purple,line join=miter,line cap=butt,line width=0.800pt]
    (342.5999,198.5592) node[above right,color=purple] (text6765-3-43-0)
    {5\,@\,8$\times$8$\times$8};
  \path[fill=purple,line join=miter,line cap=butt,line width=0.800pt]
    (349.5600,207.6460) node[above right,color=purple] (text6765-3-43-2)
    {1\,@\,8$\times$8$\times$16};
  \path[fill=c00ff00,line join=miter,line cap=butt,line width=0.800pt]
    (340.8173,117.4951) node[above right,color=c00ff00] (text6765-3-43-8) {6\,@\,8$\times$8};
  \path[fill=c00ff00,line join=miter,line cap=butt,line width=0.800pt]
    (351.7478,128.2119) node[above right,color=c00ff00] (text6765-3-43-8-8) {2\,@\,8$\times$8};
  \path[fill=c00ff00,line join=miter,line cap=butt,line width=0.800pt]
    (362.7434,138.9287) node[above right,color=c00ff00] (text6765-3-43-8-1) {1\,@\,8$\times$8};
  \path[fill=c00ff00,line join=miter,line cap=butt,line width=0.800pt]
    (513.6976,119.0158) node[above right,color=c00ff00] (text6765-3-43-8-1-1)
    {1\,@\,8$\times$8};
  \path[fill=c00ff00,line join=miter,line cap=butt,line width=0.800pt]
    (518.2509,127.4976) node[above right,color=c00ff00] (text6765-3-43-8-1-3)
    {1\,@\,8$\times$8};
  \path[fill=c00ff00,line join=miter,line cap=butt,line width=0.800pt]
    (522.8041,135.9794) node[above right,color=c00ff00] (text6765-3-43-8-1-8)
    {1\,@\,8$\times$8};
  \path[fill=purple,line join=miter,line cap=butt,line width=0.800pt]
    (497.2719,189.5618) node[above right,color=purple] (text6765-3-43-6)
    {1\,@\,8$\times$8$\times$4};
  \path[fill=purple,line join=miter,line cap=butt,line width=0.800pt]
    (508.7105,198.6320) node[above right,color=purple] (text6765-3-43-0-0)
    {1\,@\,8$\times$8$\times$8};
  \path[fill=purple,line join=miter,line cap=butt,line width=0.800pt]
    (515.2935,207.7021) node[above right,color=purple] (text6765-3-43-2-5)
    {1\,@\,8$\times$8$\times$16};
  \path[fill=black,line join=miter,line cap=butt,line width=0.800pt]
    (589.6213,181.7294) node[align=center] (text5885-4-6-4-7) {Reactive \\ Policy};
  \path[fill=black,line join=miter,line cap=butt,line width=0.800pt]
    (638.1798,272.7885) node[align=center] (text5885-4-6-6-3) {Action
    \\probabilities};
  \path[fill=black,line join=miter,line cap=butt,line width=0.800pt]
    (331.6502,250) node[align=center, below] (text5885-4-6) {Multilevel \\reward
    maps};
  \path[fill=black,line join=miter,line cap=butt,line width=0.800pt]
    (497.4938,250) node[align=center,below] (text5885-4-6-6) {Multilevel \\state
    value maps};
\end{scope}

\end{tikzpicture}

%% file: fig/abstraction_module.pgf
\definecolor{c0000ff}{RGB}{0,0,255}
\definecolor{cffffff}{RGB}{255,255,255}
\definecolor{cff0000}{RGB}{255,0,0}
\definecolor{ced0000}{RGB}{237,0,0}

\begin{tikzpicture}[y=0.80pt, x=0.80pt, yscale=-1.000000, xscale=1.000000, inner sep=0pt, outer sep=0pt]
\begin{scope}[shift={(0,-527.95276)},font=\footnotesize]
  \begin{scope}[cm={{0.50671,0.0,0.0,-0.50671,(183.8895,1259.0086)}},draw=c0000ff]
    \path[draw=c0000ff,fill=cffffff,miter limit=4.00,line width=0.395pt,rounded
      corners=0.0000cm] (58.3878,775.5265) rectangle (136.4247,853.5634);
    \begin{scope}[cm={{0.48735,0.0,0.0,0.48735,(38.6493,496.88912)}},draw=c0000ff,miter limit=4.00,line width=0.322pt]
      \path[draw=c0000ff,fill=cffffff,miter limit=4.00,line width=0.322pt,rounded
        corners=0.0000cm] (40.5000,571.8622) rectangle (200.5000,731.8622);
    \end{scope}
  \end{scope}
  \path[fill=c0000ff,line join=miter,line cap=butt,line width=0.800pt]
    (245.3522,890) node[align=center,color=blue] (text4409-6-1) {\emph{Level-3} Map\\ $6$\,@\,$8\times8$};
  \path[draw=c0000ff,line join=miter,line cap=butt,miter limit=4.00,even odd
    rule,line width=0.200pt] (243.3175,771.9292) -- (252.8936,826.3757);
  \begin{scope}[cm={{0.50671,0.0,0.0,-0.50671,(186.42305,1261.5422)}},draw=c0000ff]
    \path[draw=c0000ff,fill=cffffff,miter limit=4.00,line width=0.395pt,rounded
      corners=0.0000cm] (58.3878,775.5265) rectangle (136.4247,853.5634);
    \begin{scope}[cm={{0.48735,0.0,0.0,0.48735,(38.6493,496.88912)}},draw=c0000ff,miter limit=4.00,line width=0.322pt]
      \path[draw=c0000ff,fill=cffffff,miter limit=4.00,line width=0.322pt,rounded
        corners=0.0000cm] (40.5000,571.8622) rectangle (200.5000,731.8622);
    \end{scope}
  \end{scope}
  \begin{scope}[cm={{0.50671,0.0,0.0,-0.50671,(188.9566,1264.0758)}},draw=c0000ff]
    \path[draw=c0000ff,fill=cffffff,miter limit=4.00,line width=0.395pt,rounded
      corners=0.0000cm] (58.3878,775.5265) rectangle (136.4247,853.5634);
    \begin{scope}[cm={{0.48735,0.0,0.0,0.48735,(38.6493,496.88912)}},draw=c0000ff,miter limit=4.00,line width=0.322pt]
      \path[draw=c0000ff,fill=cffffff,miter limit=4.00,line width=0.322pt,rounded
        corners=0.0000cm] (40.5000,571.8622) rectangle (200.5000,731.8622);
    \end{scope}
  \end{scope}
  \begin{scope}[cm={{0.50671,0.0,0.0,-0.50671,(191.49015,1266.6093)}},draw=c0000ff]
    \path[draw=c0000ff,fill=cffffff,miter limit=4.00,line width=0.395pt,rounded
      corners=0.0000cm] (58.3878,775.5265) rectangle (136.4247,853.5634);
    \begin{scope}[cm={{0.48735,0.0,0.0,0.48735,(38.6493,496.88912)}},draw=c0000ff,miter limit=4.00,line width=0.322pt]
      \path[draw=c0000ff,fill=cffffff,miter limit=4.00,line width=0.322pt,rounded
        corners=0.0000cm] (40.5000,571.8622) rectangle (200.5000,731.8622);
    \end{scope}
  \end{scope}
  \begin{scope}[cm={{0.50671,0.0,0.0,-0.50671,(194.0237,1269.1429)}},draw=c0000ff]
    \path[draw=c0000ff,fill=cffffff,miter limit=4.00,line width=0.395pt,rounded
      corners=0.0000cm] (58.3878,775.5265) rectangle (136.4247,853.5634);
    \begin{scope}[cm={{0.48735,0.0,0.0,0.48735,(38.6493,496.88912)}},draw=c0000ff,miter limit=4.00,line width=0.322pt]
      \path[draw=c0000ff,fill=cffffff,miter limit=4.00,line width=0.322pt,rounded
        corners=0.0000cm] (40.5000,571.8622) rectangle (200.5000,731.8622);
    \end{scope}
  \end{scope}
  \path[xscale=1.000,yscale=-1.000,draw=c0000ff,fill=cffffff,miter limit=4.00,line
    width=0.200pt,rounded corners=0.0000cm] (226.1429,-878.7092) rectangle
    (265.6850,-839.1671);
  \path[xscale=1.000,yscale=-1.000,draw=c0000ff,fill=cffffff,miter limit=4.00,line
    width=0.200pt,rounded corners=0.0000cm] (226.1426,-878.6783) rectangle
    (265.6540,-839.1669);

  \path[draw=c0000ff,line join=miter,line cap=butt,miter limit=4.00,fill
    opacity=0.750,even odd rule,line width=0.079pt] (226.0500,878.8018) --
    (226.0500,839.0435)(230.9889,878.8018) --
    (230.9889,839.0435)(235.9278,878.8018) --
    (235.9278,839.0435)(240.8667,878.8018) --
    (240.8667,839.0435)(245.8057,878.8018) --
    (245.8057,839.0435)(250.7446,878.8018) --
    (250.7446,839.0435)(255.6835,878.8018) --
    (255.6835,839.0435)(260.6224,878.8018) --
    (260.6224,839.0435)(265.5614,878.8018) --
    (265.5614,839.0435)(226.0500,878.8018) --
    (265.8083,878.8018)(226.0500,873.8629) --
    (265.8083,873.8629)(226.0500,868.9240) --
    (265.8083,868.9240)(226.0500,863.9850) --
    (265.8083,863.9850)(226.0500,859.0461) --
    (265.8083,859.0461)(226.0500,854.1072) --
    (265.8083,854.1072)(226.0500,849.1683) --
    (265.8083,849.1683)(226.0500,844.2293) --
    (265.8083,844.2293)(226.0500,839.2904) -- (265.8083,839.2904);
  \path[xscale=1.000,yscale=-1.000,draw=c0000ff,fill=cffffff,miter limit=4.00,line
    width=0.203pt,rounded corners=0.0000cm] (223.1750,-771.9292) rectangle
    (243.3175,-751.7737);
  \path[xscale=1.000,yscale=-1.000,draw=c0000ff,fill=cffffff,miter limit=4.00,line
    width=0.203pt,rounded corners=0.0000cm] (225.7086,-774.4628) rectangle
    (245.8510,-754.3072);
  \path[xscale=1.000,yscale=-1.000,draw=c0000ff,fill=cffffff,miter limit=4.00,line
    width=0.203pt,rounded corners=0.0000cm] (228.2421,-776.9963) rectangle
    (248.3846,-756.8408);
  \path[xscale=1.000,yscale=-1.000,draw=c0000ff,fill=cffffff,miter limit=4.00,line
    width=0.203pt,rounded corners=0.0000cm] (230.7757,-779.5299) rectangle
    (250.9181,-759.3743);
  \path[xscale=1.000,yscale=-1.000,draw=c0000ff,fill=cffffff,miter limit=4.00,line
    width=0.203pt,rounded corners=0.0000cm] (233.3092,-782.0634) rectangle
    (253.4517,-761.9079);
  \path[xscale=1.000,yscale=-1.000,draw=c0000ff,fill=cffffff,miter limit=4.00,line
    width=0.203pt,rounded corners=0.0000cm] (235.8428,-784.5474) rectangle
    (255.9852,-764.3919);
  \path[draw=c0000ff,line join=miter,line cap=butt,miter limit=4.00,fill
    opacity=0.750,even odd rule,line width=0.080pt] (235.8427,764.3919) --
    (235.8427,784.6734)(238.3763,764.3919) --
    (238.3763,784.6734)(240.9099,764.3919) --
    (240.9099,784.6734)(243.4434,764.3919) --
    (243.4434,784.6734)(245.9770,764.3919) --
    (245.9770,784.6734)(248.5105,764.3919) --
    (248.5105,784.6734)(251.0441,764.3919) --
    (251.0441,784.6734)(253.5776,764.3919) --
    (253.5776,784.6734)(235.8427,764.3919) --
    (256.1112,764.3919)(235.8427,766.9255) --
    (256.1112,766.9255)(235.8427,769.4590) --
    (256.1112,769.4590)(235.8427,771.9926) --
    (256.1112,771.9926)(235.8427,774.5261) --
    (256.1112,774.5261)(235.8427,777.0597) --
    (256.1112,777.0597)(235.8427,779.5932) --
    (256.1112,779.5932)(235.8427,782.1268) --
    (256.1112,782.1268)(235.8427,784.6603) -- (256.1112,784.6603);
  \path[draw=c0000ff,line join=miter,line cap=butt,miter limit=4.00,even odd
    rule,line width=0.200pt] (235.8427,784.5474) -- (226.0500,839.2904);
  \path[draw=c0000ff,line join=miter,line cap=butt,miter limit=4.00,even odd
    rule,line width=0.200pt] (256.1112,784.6603) -- (265.6540,839.1670);
  \path[draw=c0000ff,line join=miter,line cap=butt,miter limit=4.00,even odd
    rule,line width=0.200pt] (223.1750,771.9292) -- (213.3822,826.3757);
  \path[fill=cff0000,line join=miter,line cap=butt,line width=0.800pt]
    (158.2511,890) node[align=center,color=red] (text4409-6) {\emph{Level-2} Map\\ $2$\,@\,$8\times8$};
  \begin{scope}[cm={{0.50671,0.0,0.0,-0.50671,(103.83287,1262.7496)}},draw=cff0000]
    \path[draw=cff0000,fill=cffffff,miter limit=4.00,line width=0.395pt,rounded
      corners=0.0000cm] (58.3878,775.5265) rectangle (136.4247,853.5634);
    \begin{scope}[cm={{0.48735,0.0,0.0,0.48735,(38.6493,496.88912)}},draw=cff0000,miter limit=4.00,line width=0.322pt]
      \path[draw=cff0000,fill=cffffff,miter limit=4.00,line width=0.322pt,rounded
        corners=0.0000cm] (40.5000,571.8622) rectangle (200.5000,731.8622);
    \end{scope}
  \end{scope}
  \path[xscale=1.000,yscale=-1.000,draw=cff0000,fill=cffffff,miter limit=4.00,line
    width=0.200pt,rounded corners=0.0000cm] (138.6045,-874.9683) rectangle
    (178.1465,-835.4262);
  \begin{scope}[cm={{0.24695,0.0,0.0,-0.24695,(128.60274,1016.1567)}},draw=cff0000,miter limit=4.00,line width=0.322pt]
    \path[draw=cff0000,fill=cffffff,miter limit=4.00,line width=0.322pt,rounded
      corners=0.0000cm] (40.5000,571.8622) rectangle (200.5000,731.8622);
    \path[draw=cff0000,line join=miter,line cap=butt,miter limit=4.00,fill
      opacity=0.750,even odd rule,line width=0.321pt] (40.1250,571.3622) --
      (40.1250,732.3622)(60.1250,571.3622) -- (60.1250,732.3622)(80.1250,571.3622)
      -- (80.1250,732.3622)(100.1250,571.3622) --
      (100.1250,732.3622)(120.1250,571.3622) --
      (120.1250,732.3622)(140.1250,571.3622) --
      (140.1250,732.3622)(160.1250,571.3622) --
      (160.1250,732.3622)(180.1250,571.3622) --
      (180.1250,732.3622)(200.1250,571.3622) --
      (200.1250,732.3622)(40.1250,571.3622) -- (201.1250,571.3622)(40.1250,591.3622)
      -- (201.1250,591.3622)(40.1250,611.3622) --
      (201.1250,611.3622)(40.1250,631.3622) -- (201.1250,631.3622)(40.1250,651.3622)
      -- (201.1250,651.3622)(40.1250,671.3622) --
      (201.1250,671.3622)(40.1250,691.3622) -- (201.1250,691.3622)(40.1250,711.3622)
      -- (201.1250,711.3622)(40.1250,731.3622) -- (201.1250,731.3622);
  \end{scope}
  \begin{scope}[cm={{0.50671,0.0,0.0,0.50671,(133.01604,252.65513)}}]
    \path[xscale=1.000,yscale=-1.000,draw=cff0000,fill=cffffff,miter limit=4.00,line
      width=0.397pt,rounded corners=0.0000cm] (0.2481,-1052.1141) rectangle
      (79.6120,-972.6988);
    \path[xscale=1.000,yscale=-1.000,draw=cff0000,dash pattern=on 0.78pt off
      0.78pt,miter limit=4.00,line width=0.781pt,rounded corners=0.0000cm]
      (20.4880,-1031.8743) rectangle (59.5124,-992.8499);
  \end{scope}
  \begin{scope}[cm={{0.50671,0.0,0.0,0.50671,(-66.79571,293.1166)}}]
    \path[xscale=1.000,yscale=-1.000,draw=cff0000,miter limit=4.00,line
      width=0.397pt,rounded corners=0.0000cm] (404.5795,-982.2628) rectangle
      (483.9434,-902.8475);
    \path[xscale=1.000,yscale=-1.000,draw=cff0000,fill=cffffff,miter limit=4.00,line
      width=0.395pt,rounded corners=0.0000cm] (424.8195,-962.0229) rectangle
      (463.8439,-922.9985);
  \end{scope}
  \path[draw=cff0000,line join=miter,line cap=butt,miter limit=4.00,even odd
    rule,line width=0.200pt] (148.4646,780.5834) -- (138.5115,835.5495);
  \path[draw=cff0000,line join=miter,line cap=butt,miter limit=4.00,even odd
    rule,line width=0.200pt] (168.2387,780.5834) -- (178.0229,835.3026);
  \path[draw=cff0000,line join=miter,line cap=butt,miter limit=4.00,even odd
    rule,line width=0.200pt] (143.3975,775.5163) -- (133.3256,830.3636);
  \path[draw=cff0000,line join=miter,line cap=butt,miter limit=4.00,fill
    opacity=0.750,even odd rule,line width=0.080pt] (148.2746,760.6193) --
    (148.2746,780.9008)(150.8082,760.6193) --
    (150.8082,780.9008)(153.3417,760.6193) --
    (153.3417,780.9008)(155.8753,760.6193) --
    (155.8753,780.9008)(158.4088,760.6193) --
    (158.4088,780.9008)(160.9424,760.6193) --
    (160.9424,780.9008)(163.4759,760.6193) --
    (163.4759,780.9008)(166.0095,760.6193) --
    (166.0095,780.9008)(148.2746,760.6193) --
    (168.5430,760.6193)(148.2746,763.1529) --
    (168.5430,763.1529)(148.2746,765.6864) --
    (168.5430,765.6864)(148.2746,768.2200) --
    (168.5430,768.2200)(148.2746,770.7535) --
    (168.5430,770.7535)(148.2746,773.2871) --
    (168.5430,773.2871)(148.2746,775.8206) --
    (168.5430,775.8206)(148.2746,778.3542) --
    (168.5430,778.3542)(148.2746,780.8877) -- (168.5430,780.8877);
  \path[draw=cff0000,line join=miter,line cap=butt,miter limit=4.00,even odd
    rule,line width=0.200pt] (164.6048,780.7116) -- (173.0839,830.3636);
  \path[fill=ced0000,line join=miter,line cap=butt,line width=0.800pt]
    (133,744.9416) node[above right, color=red] (text6765-3) {2\,@\,16$\times$16};
  \path[fill=black,line join=miter,line cap=butt,line width=0.800pt]
    (56.8104,720) node[align=center] (text7573) {Occupancy Map};
  \begin{scope}[cm={{0.50671,0.0,0.0,0.50671,(7.94986,437.85594)}}]
    \path[fill=black,line join=miter,line cap=butt,line width=0.800pt]
      (96.1921,890) node[align=center] (text4409) {\emph{Level-1} Map\\ $1$\,@\,$8\times8$};
    \path[xscale=1.000,yscale=-1.000,draw=black,fill=cffffff,miter limit=4.00,line
      width=0.395pt,rounded corners=0.0000cm] (57.4191,-857.5305) rectangle
      (135.4559,-779.4936);
    \begin{scope}[cm={{0.48735,0.0,0.0,-0.48735,(37.68055,1136.1679)}},miter limit=4.00,line width=0.322pt]
      \path[draw=black,fill=cffffff,miter limit=4.00,line width=0.322pt,rounded
        corners=0.0000cm] (40.5000,571.8622) rectangle (200.5000,731.8622);
      \path[draw=black,line join=miter,line cap=butt,miter limit=4.00,fill
        opacity=0.750,even odd rule,line width=0.321pt] (40.1250,571.3622) --
        (40.1250,732.3622)(60.1250,571.3622) -- (60.1250,732.3622)(80.1250,571.3622)
        -- (80.1250,732.3622)(100.1250,571.3622) --
        (100.1250,732.3622)(120.1250,571.3622) --
        (120.1250,732.3622)(140.1250,571.3622) --
        (140.1250,732.3622)(160.1250,571.3622) --
        (160.1250,732.3622)(180.1250,571.3622) --
        (180.1250,732.3622)(200.1250,571.3622) --
        (200.1250,732.3622)(40.1250,571.3622) -- (201.1250,571.3622)(40.1250,591.3622)
        -- (201.1250,591.3622)(40.1250,611.3622) --
        (201.1250,611.3622)(40.1250,631.3622) -- (201.1250,631.3622)(40.1250,651.3622)
        -- (201.1250,651.3622)(40.1250,671.3622) --
        (201.1250,671.3622)(40.1250,691.3622) -- (201.1250,691.3622)(40.1250,711.3622)
        -- (201.1250,711.3622)(40.1250,731.3622) -- (201.1250,731.3622);
    \end{scope}
    \path[xscale=1.000,yscale=-1.000,draw=black,fill=cffffff,miter limit=4.00,line
      width=0.395pt,rounded corners=0.0000cm] (16.9971,-731.3909) rectangle
      (176.0029,-572.2820);
    \path[xscale=1.000,yscale=-1.000,draw=black,miter limit=4.00,line
      width=0.395pt,rounded corners=0.0000cm] (76.9878,-671.4001) rectangle
      (116.0122,-632.3757);

  \path[draw=black,line join=miter,line cap=butt,miter limit=4.00,fill
    opacity=0.750,even odd rule,line width=0.080pt] (76.9878,632.3757) --
    (76.9878,632.3757)( 81.86585,671.4001) --
    ( 81.86585,632.3757)( 86.7439,671.4001) --
    ( 86.7439,632.3757)( 91.62195,671.4001) --
    ( 91.62195,632.3757)(96.5,671.4001) --
    (96.5,632.3757)( 101.37805,671.4001) --
    ( 101.37805,632.3757)( 106.2561,671.4001) --
    ( 106.2561,632.3757)(111.134151,671.4001) --
    ( 111.13415,632.3757)(76.9878,632.3757) --
    (116.0122,632.3757)(76.9878, 637.25375) --
    (116.0122, 637.25375)(76.9878,642.1318) --
    (116.0122,642.1318)(76.9878,647.00985) --
    (116.0122,647.00985)(76.9878,651.8879) --
    (116.0122,651.8879)(76.9878,656.76595) --
    (116.0122,656.76595)(76.9878, 661.644) --
    (116.0122, 661.644)(76.9878,666.52205) --
    (116.0122,666.52205)(76.9878,671.4001) -- (116.0122,671.4001);
    \path[draw=black,line join=miter,line cap=butt,miter limit=4.00,even odd
      rule,line width=0.395pt] (116.0122,671.4001) -- (135.6993,779.7369);
    \path[draw=black,line join=miter,line cap=butt,miter limit=4.00,even odd
      rule,line width=0.395pt] (76.9878,671.4001) -- (57.2356,779.2496);
    \path[fill=black,line join=miter,line cap=butt,line width=0.800pt]
      (176.4708,570.6904) node[above right] (text6765) {1\,@\,32$\times$32};
  \end{scope}
  \path[draw=black,-Latex,line join=miter,line cap=butt,miter limit=4.00,even odd
    rule,line width=0.203pt] (101.3655,766.1451) -- (126.3391,766.1451);
  \path[fill=black,line join=miter,line cap=butt,line width=0.800pt]
    (113.9189,755) node[align=center] (text5374-3-5-4) {conv.};
  \path[fill=black,line join=miter,line cap=butt,line width=0.800pt]
    (113.9189,780) node[align=center] (text5374-3-5-4) {+ Max \\pool};
  \path[draw=black,-Latex,line join=miter,line cap=butt,miter limit=4.00,even odd
    rule,line width=0.203pt] (185.8575,766.1451) -- (210.8311,766.1451);
  \path[fill=black,line join=miter,line cap=butt,line width=0.800pt]
    (198.4109,755) node[align=center] (text5374-3-5-4-2) {conv.};
  \path[fill=black,line join=miter,line cap=butt,line width=0.800pt]
    (198.4109,780) node[align=center] (text5374-3-5-4) {+ Max \\pool};
  \path[fill=c0000ff,line join=miter,line cap=butt,line width=0.800pt]
    (217,751.1829) node[above right,color=blue] (text6765-3-6) {6\,@\,8$\times$8};
\end{scope}

\end{tikzpicture}

%% file: fig/reward_module_final.pgf
\begin{tikzpicture}[y=0.80pt, x=0.80pt, inner sep=0pt, outer sep=0pt, font=\scriptsize]
\def\squaresize{4.0};
\def\gridlinewidth{0.1};
\def\borderlinewidth{0.3};
\def\arrowlength{7*\squaresize};
\def\endpoint{120*\squaresize};
\def\leveldistance{20*\squaresize};

\begin{scope}[]
    \path[fill=black,line join=miter,line cap=butt,line width=0.800pt] (4*\squaresize,0) node[] {1\,@\,8$\times$8};
	\draw[black, line width = \borderlinewidth, fill=white] (0,-\squaresize) rectangle ++(8*\squaresize,-8*\squaresize);
	\draw[step=\squaresize, black, line width = \gridlinewidth] (0,-\squaresize) grid ++(8*\squaresize,-8*\squaresize);

    \path[fill=black,line join=miter,line cap=butt,line width=0.800pt] (12.5*\squaresize,-2*\squaresize) node[] {1\,@\,8$\times$8};
	\draw[black, line width = \borderlinewidth, fill=white] (2*\squaresize,-3*\squaresize) rectangle ++(8*\squaresize,-8*\squaresize);
	\draw[step=\squaresize, black, line width = \gridlinewidth] (2*\squaresize,-3*\squaresize) grid ++(8*\squaresize,-8*\squaresize);
\end{scope}

\begin{scope}[shift={(14*\squaresize,0)}]
	\path[fill=black,line join=miter,line cap=butt,line width=0.800pt] (0.5*\arrowlength,-4.5*\squaresize) node[align=center] {conv.};
	\path[fill=black,line join=miter,line cap=butt,line width=0.800pt] (0.5*\arrowlength,-9*\squaresize) node[align=center] {Kernel:\\3$\times$3};
	\path[draw=black,-Latex,line join=miter,line cap=butt,miter limit=4.00,even odd  rule,line width=\borderlinewidth] (0,-6*\squaresize) -- ++ (\arrowlength,0);
\end{scope}

\begin{scope}[shift={(22*\squaresize,0)}]
    \path[fill=black,line join=miter,line cap=butt,line width=0.800pt] (4*\squaresize,0) node[] {150\,@\,8$\times$8};
	\draw[black, line width = \borderlinewidth, fill=white] (0,-\squaresize) rectangle ++(8*\squaresize,-8*\squaresize);
	\draw[black, line width = \borderlinewidth, fill=white] (0.5*\squaresize,-1.5*\squaresize) rectangle ++(8*\squaresize,-8*\squaresize);

	\draw[dotted, color=black, line width=\borderlinewidth] (0.5*\squaresize,-9.5*\squaresize) -- (2*\squaresize,-11.5*\squaresize);
	\draw[dotted, color=black, line width=\borderlinewidth] (0.5*\squaresize,-1.5*\squaresize) -- (2*\squaresize,-3.5*\squaresize);
	\draw[dotted, color=black, line width=\borderlinewidth] (8.5*\squaresize,-1.5*\squaresize) -- (10*\squaresize,-3.5*\squaresize);

	\draw[black, line width = \borderlinewidth, fill=white] (2*\squaresize,-3.5*\squaresize) rectangle ++(8*\squaresize,-8*\squaresize);
	\draw[black, line width = \borderlinewidth, fill=white] (2.5*\squaresize,-4*\squaresize) rectangle ++(8*\squaresize,-8*\squaresize);
\end{scope}

		\path[fill=black,line join=miter,line cap=butt,line width=0.800pt] (95*\squaresize,0.4) node[align=center] {conv., Kernel: 1$\times$1};
		\path[draw=black,line join=miter,line cap=butt,miter limit=4.00,even odd  rule,line width=\borderlinewidth] (35*\squaresize,-3*\squaresize) -- ++(\squaresize,0)--++(0,2*\squaresize)-- (\endpoint,-\squaresize)--(\endpoint,-\leveldistance)--++(0,-6*\squaresize);

\begin{scope}[shift={(0,-3*\squaresize)}]
	\begin{scope}[shift={(35*\squaresize,0)}]
		\path[fill=black,line join=miter,line cap=butt,line width=0.800pt] (0.5*\arrowlength,-3.5*\squaresize) node[align=left] {conv. +\\Max pool};
		\path[fill=black,line join=miter,line cap=butt,line width=0.800pt] (0.5*\arrowlength,-9*\squaresize) node[align=center] {Kernels:\\3$\times$3, 2$\times$2};
		\path[draw=black,-Latex,line join=miter,line cap=butt,miter limit=4.00,even odd  rule,line width=\borderlinewidth] (0,-6*\squaresize) -- ++ (\arrowlength,0);
	\end{scope}

	\begin{scope}[shift={(45.5*\squaresize,-2*\squaresize)}]
	    \path[fill=black,line join=miter,line cap=butt,line width=0.800pt] (2*\squaresize,0) node[] {150\,@\,4$\times$4};
		\draw[black, line width = \borderlinewidth, fill=white] (0,-\squaresize) rectangle ++(4*\squaresize,-4*\squaresize);
		\draw[black, line width = \borderlinewidth, fill=white] (0.5*\squaresize,-1.5*\squaresize) rectangle ++(4*\squaresize,-4*\squaresize);
	
		\draw[dotted, color=black, line width=\borderlinewidth] (0.5*\squaresize,-5.5*\squaresize) -- (2*\squaresize,-7.5*\squaresize);
		\draw[dotted, color=black, line width=\borderlinewidth] (0.5*\squaresize,-1.5*\squaresize) -- (2*\squaresize,-3.5*\squaresize);
		\draw[dotted, color=black, line width=\borderlinewidth] (4.5*\squaresize,-1.5*\squaresize) -- (6*\squaresize,-3.5*\squaresize);
	
		\draw[black, line width = \borderlinewidth, fill=white] (2*\squaresize,-3.5*\squaresize) rectangle ++(4*\squaresize,-4*\squaresize);
		\draw[black, line width = \borderlinewidth, fill=white] (2.5*\squaresize,-4*\squaresize) rectangle ++(4*\squaresize,-4*\squaresize);
	\end{scope}
	
	\begin{scope}[shift={(54*\squaresize,0)}]
		\path[fill=black,line join=miter,line cap=butt,line width=0.800pt] (0.5*\arrowlength,-4.5*\squaresize) node[align=center] {pad};
		\path[fill=black,line join=miter,line cap=butt,line width=0.800pt] (0.5*\arrowlength,-8*\squaresize) node[align=center] {(zeros)};
		\path[draw=black,-Latex,line join=miter,line cap=butt,miter limit=4.00,even odd  rule,line width=\borderlinewidth] (0,-6*\squaresize) -- ++ (\arrowlength,0);
	\end{scope}

	\begin{scope}[shift={(63*\squaresize,0)}]
    \path[fill=black,line join=miter,line cap=butt,line width=0.800pt] (4*\squaresize,0) node[] {150\,@\,8$\times$8};
	\draw[black, line width = \borderlinewidth, fill=white] (0,-\squaresize) rectangle ++(8*\squaresize,-8*\squaresize);
	\draw[black, line width = \borderlinewidth, fill=white] (0.5*\squaresize,-1.5*\squaresize) rectangle ++(8*\squaresize,-8*\squaresize);

	\draw[dotted, color=black, line width=\borderlinewidth] (0.5*\squaresize,-9.5*\squaresize) -- (2*\squaresize,-11.5*\squaresize);
	\draw[dotted, color=black, line width=\borderlinewidth] (0.5*\squaresize,-1.5*\squaresize) -- (2*\squaresize,-3.5*\squaresize);
	\draw[dotted, color=black, line width=\borderlinewidth] (8.5*\squaresize,-1.5*\squaresize) -- (10*\squaresize,-3.5*\squaresize);

	\draw[black, line width = \borderlinewidth, fill=white] (2*\squaresize,-3.5*\squaresize) rectangle ++(8*\squaresize,-8*\squaresize);
	\draw[black, line width = \borderlinewidth, fill=white] (2.5*\squaresize,-4*\squaresize) rectangle ++(8*\squaresize,-8*\squaresize);
	\draw[black, line width = \borderlinewidth, fill=white] (4.5*\squaresize,-6*\squaresize) rectangle ++(4*\squaresize,-4*\squaresize);
	\end{scope}

	  \path[draw=black,-Latex,line join=miter,line cap=butt,miter limit=4.00,even odd
    rule,line width=0.324pt] (69*\squaresize,-13*\squaresize) --++ (0,-0.1*\leveldistance) --++(-35*\squaresize,0) --++ (0,-0.25*\leveldistance) --++(\arrowlength,0);

\end{scope}

\begin{scope}[shift={(0,-\leveldistance)}]
		\begin{scope}[]
		    \path[fill=black,line join=miter,line cap=butt,line width=0.800pt] (4*\squaresize,0) node[] {1\,@\,8$\times$8};
			\draw[red, line width = \borderlinewidth, fill=white] (0,-\squaresize) rectangle ++(8*\squaresize,-8*\squaresize);
			\draw[step=\squaresize, red, line width = \gridlinewidth] (0,-\squaresize) grid ++(8*\squaresize,-8*\squaresize);
		
		    \path[fill=black,line join=miter,line cap=butt,line width=0.800pt] (12.5*\squaresize,-2*\squaresize) node[] {2\,@\,8$\times$8};
			\draw[red, line width = \borderlinewidth, fill=white] (2*\squaresize,-3*\squaresize) rectangle ++(8*\squaresize,-8*\squaresize);
			\draw[red, line width = \borderlinewidth, fill=white] (2.5*\squaresize,-3.5*\squaresize) rectangle ++(8*\squaresize,-8*\squaresize);
			\draw[step=\squaresize, red, line width = \gridlinewidth, shift={(0.5*\squaresize,-0.5*\squaresize)}] (2.0*\squaresize,-3.0*\squaresize) grid ++(8*\squaresize,-8*\squaresize);
		\end{scope}
		
		\begin{scope}[shift={(14*\squaresize,0)}]
			\path[fill=black,line join=miter,line cap=butt,line width=0.800pt] (0.5*\arrowlength,-4.5*\squaresize) node[align=center] {conv.};
	\path[fill=black,line join=miter,line cap=butt,line width=0.800pt] (0.5*\arrowlength,-9*\squaresize) node[align=center] {Kernel:\\3$\times$3};
			\path[draw=black,-Latex,line join=miter,line cap=butt,miter limit=4.00,even odd  rule,line width=\borderlinewidth] (0,-6*\squaresize) -- ++ (\arrowlength,0);
		\end{scope}
		
		\begin{scope}[shift={(22*\squaresize,0)}]
		    \path[fill=black,line join=miter,line cap=butt,line width=0.800pt] (4*\squaresize,0) node[] {150\,@\,8$\times$8};
			\draw[red, line width = \borderlinewidth, fill=white] (0,-\squaresize) rectangle ++(8*\squaresize,-8*\squaresize);
			\draw[red, line width = \borderlinewidth, fill=white] (0.5*\squaresize,-1.5*\squaresize) rectangle ++(8*\squaresize,-8*\squaresize);
		
			\draw[dotted, color=red, line width=\borderlinewidth] (0.5*\squaresize,-9.5*\squaresize) -- (2*\squaresize,-11.5*\squaresize);
			\draw[dotted, color=red, line width=\borderlinewidth] (0.5*\squaresize,-1.5*\squaresize) -- (2*\squaresize,-3.5*\squaresize);
			\draw[dotted, color=red, line width=\borderlinewidth] (8.5*\squaresize,-1.5*\squaresize) -- (10*\squaresize,-3.5*\squaresize);
		
			\draw[red, line width = \borderlinewidth, fill=white] (2*\squaresize,-3.5*\squaresize) rectangle ++(8*\squaresize,-8*\squaresize);
			\draw[red, line width = \borderlinewidth, fill=white] (2.5*\squaresize,-4*\squaresize) rectangle ++(8*\squaresize,-8*\squaresize);
		\end{scope}
	
		\begin{scope}[shift={(34*\squaresize,0)}]
			\path[fill=black,line join=miter,line cap=butt,line width=0.800pt] (0.5*\arrowlength,-4.5*\squaresize) node[align=center] {stack.};
			\path[draw=black,-Latex,line join=miter,line cap=butt,miter limit=4.00,even odd  rule,line width=\borderlinewidth] (0,-6*\squaresize) -- ++ (\arrowlength,0);
		\end{scope}

		\begin{scope}[shift={(42*\squaresize,0)}]
		    \path[fill=black,line join=miter,line cap=butt,line width=0.800pt] (4*\squaresize,0) node[] {300\,@\,8$\times$8};
			\draw[black, line width = \borderlinewidth, fill=white] (0,-\squaresize) rectangle ++(8*\squaresize,-8*\squaresize);
			
			\draw[densely dotted, color=black, line width=\borderlinewidth] (0,-9*\squaresize) --++ (1.5*\squaresize,-1.5*\squaresize);
			\draw[densely dotted, color=black, line width=\borderlinewidth] (0,-\squaresize) -- ++(1.5*\squaresize,-1.5*\squaresize);
			\draw[densely dotted, color=black, line width=\borderlinewidth] (8*\squaresize,-\squaresize) -- ++(1.5*\squaresize,-1.5*\squaresize);			

			\draw[black, line width = \borderlinewidth, fill=white] (1.5*\squaresize,-2.5*\squaresize) rectangle ++(8*\squaresize,-8*\squaresize);
			\draw[red, line width = \borderlinewidth, fill=white] (2*\squaresize,-3*\squaresize) rectangle ++(8*\squaresize,-8*\squaresize);	
			
			\draw[densely dotted, color=red, line width=\borderlinewidth] (2*\squaresize,-11*\squaresize) -- ++(1.5*\squaresize,-1.5*\squaresize);
			\draw[densely dotted, color=red, line width=\borderlinewidth] (2*\squaresize,-3*\squaresize) -- ++(1.5*\squaresize,-1.5*\squaresize);
			\draw[densely dotted, color=red, line width=\borderlinewidth] (10*\squaresize,-3*\squaresize) -- ++(1.5*\squaresize,-1.5*\squaresize);

			\draw[red, line width = \borderlinewidth, fill=white] (3.5*\squaresize,-4.5*\squaresize) rectangle ++(8*\squaresize,-8*\squaresize);
		\end{scope}
	
		\begin{scope}[shift={(55*\squaresize,0)}]
			\path[fill=black,line join=miter,line cap=butt,line width=0.800pt] (0.5*\arrowlength,-4.5*\squaresize) node[align=center] {conv.};
			\path[fill=black,line join=miter,line cap=butt,line width=0.800pt] (0.5*\arrowlength,-9*\squaresize) node[align=center] {Kernel:\\1$\times$1};
			\path[draw=black,-Latex,line join=miter,line cap=butt,miter limit=4.00,even odd  rule,line width=\borderlinewidth] (0,-6*\squaresize) -- ++ (\arrowlength,0);
		\end{scope}	

		\begin{scope}[shift={(63*\squaresize,0)}]
		    \path[fill=black,line join=miter,line cap=butt,line width=0.800pt] (4*\squaresize,0) node[] {150\,@\,8$\times$8};
			\draw[red, line width = \borderlinewidth, fill=white] (0,-\squaresize) rectangle ++(8*\squaresize,-8*\squaresize);
			\draw[red, line width = \borderlinewidth, fill=white] (0.5*\squaresize,-1.5*\squaresize) rectangle ++(8*\squaresize,-8*\squaresize);
		
			\draw[dotted, color=red, line width=\borderlinewidth] (0.5*\squaresize,-9.5*\squaresize) -- (2*\squaresize,-11.5*\squaresize);
			\draw[dotted, color=red, line width=\borderlinewidth] (0.5*\squaresize,-1.5*\squaresize) -- (2*\squaresize,-3.5*\squaresize);
			\draw[dotted, color=red, line width=\borderlinewidth] (8.5*\squaresize,-1.5*\squaresize) -- (10*\squaresize,-3.5*\squaresize);
		
			\draw[red, line width = \borderlinewidth, fill=white] (2*\squaresize,-3.5*\squaresize) rectangle ++(8*\squaresize,-8*\squaresize);
			\draw[red, line width = \borderlinewidth, fill=white] (2.5*\squaresize,-4*\squaresize) rectangle ++(8*\squaresize,-8*\squaresize);
		\end{scope}

		\path[fill=black,line join=miter,line cap=butt,line width=0.800pt] (95*\squaresize,0.4) node[align=center] {conv., Kernel: 1$\times$1};
		\path[draw=black, line join=miter,line cap=butt,miter limit=4.00,even odd  rule,line width=\borderlinewidth] (76*\squaresize,-3*\squaresize) -- ++(\squaresize,0)--++(0,2*\squaresize)-- (\endpoint-2*\squaresize,-\squaresize)--++(0,-5*\squaresize)--++(2*\squaresize,0);
		
		\begin{scope}[shift={(42*\squaresize,-3*\squaresize)}]
			\begin{scope}[shift={(34*\squaresize,0)}]
				\path[fill=black,line join=miter,line cap=butt,line width=0.800pt] (0.5*\arrowlength,-3.5*\squaresize) node[align=left] {conv. +\\Max pool};
				\path[fill=black,line join=miter,line cap=butt,line width=0.800pt] (0.5*\arrowlength,-9*\squaresize) node[align=center] {Kernels:\\3$\times$3, 2$\times$2};
				\path[draw=black,-Latex,line join=miter,line cap=butt,miter limit=4.00,even odd  rule,line width=\borderlinewidth] (0,-6*\squaresize) -- ++ (\arrowlength,0);
			\end{scope}
		
			\begin{scope}[shift={(45*\squaresize,-2*\squaresize)}]
			    \path[fill=black,line join=miter,line cap=butt,line width=0.800pt] (2*\squaresize,0) node[] {150\,@\,4$\times$4};
				\draw[red, line width = \borderlinewidth, fill=white] (0,-\squaresize) rectangle ++(4*\squaresize,-4*\squaresize);
				\draw[red, line width = \borderlinewidth, fill=white] (0.5*\squaresize,-1.5*\squaresize) rectangle ++(4*\squaresize,-4*\squaresize);
			
				\draw[dotted, color=red, line width=\borderlinewidth] (0.5*\squaresize,-5.5*\squaresize) -- (2*\squaresize,-7.5*\squaresize);
				\draw[dotted, color=red, line width=\borderlinewidth] (0.5*\squaresize,-1.5*\squaresize) -- (2*\squaresize,-3.5*\squaresize);
				\draw[dotted, color=red, line width=\borderlinewidth] (4.5*\squaresize,-1.5*\squaresize) -- (6*\squaresize,-3.5*\squaresize);
			
				\draw[red, line width = \borderlinewidth, fill=white] (2*\squaresize,-3.5*\squaresize) rectangle ++(4*\squaresize,-4*\squaresize);
				\draw[red, line width = \borderlinewidth, fill=white] (2.5*\squaresize,-4*\squaresize) rectangle ++(4*\squaresize,-4*\squaresize);
			\end{scope}
			
			\begin{scope}[shift={(53*\squaresize,0)}]
				\path[fill=black,line join=miter,line cap=butt,line width=0.800pt] (0.5*\arrowlength,-4.5*\squaresize) node[align=center] {pad};
				\path[fill=black,line join=miter,line cap=butt,line width=0.800pt] (0.5*\arrowlength,-8*\squaresize) node[align=center] {(zeros)};
				\path[draw=black,-Latex,line join=miter,line cap=butt,miter limit=4.00,even odd  rule,line width=\borderlinewidth] (0,-6*\squaresize) -- ++ (\arrowlength,0);
			\end{scope}
		
			\begin{scope}[shift={(63*\squaresize,0)}]
		    \path[fill=black,line join=miter,line cap=butt,line width=0.800pt] (4*\squaresize,0) node[] {150\,@\,8$\times$8};
			\draw[red, line width = \borderlinewidth, fill=white] (0,-\squaresize) rectangle ++(8*\squaresize,-8*\squaresize);
			\draw[red, line width = \borderlinewidth, fill=white] (0.5*\squaresize,-1.5*\squaresize) rectangle ++(8*\squaresize,-8*\squaresize);
		
			\draw[dotted, color=red, line width=\borderlinewidth] (0.5*\squaresize,-9.5*\squaresize) -- (2*\squaresize,-11.5*\squaresize);
			\draw[dotted, color=red, line width=\borderlinewidth] (0.5*\squaresize,-1.5*\squaresize) -- (2*\squaresize,-3.5*\squaresize);
			\draw[dotted, color=red, line width=\borderlinewidth] (8.5*\squaresize,-1.5*\squaresize) -- (10*\squaresize,-3.5*\squaresize);
		
			\draw[red, line width = \borderlinewidth, fill=white] (2*\squaresize,-3.5*\squaresize) rectangle ++(8*\squaresize,-8*\squaresize);
			\draw[red, line width = \borderlinewidth, fill=white] (2.5*\squaresize,-4*\squaresize) rectangle ++(8*\squaresize,-8*\squaresize);
			\draw[red, line width = \borderlinewidth, fill=white] (4.5*\squaresize,-6*\squaresize) rectangle ++(4*\squaresize,-4*\squaresize);
			\end{scope}
		\end{scope}

	  \path[draw=black,-Latex,line join=miter,line cap=butt,miter limit=4.00,even odd
    rule,line width=0.324pt] (111*\squaresize,-16*\squaresize) --++ (0,-0.1*\leveldistance) --++(-77*\squaresize,0) --++ (0,-0.25*\leveldistance) --++(\arrowlength,0);

\end{scope}

\begin{scope}[shift={(0,-2*\leveldistance)}]
		\begin{scope}[]
		    \path[fill=black,line join=miter,line cap=butt,line width=0.800pt] (4*\squaresize,0) node[] {1\,@\,8$\times$8};
			\draw[blue, line width = \borderlinewidth, fill=white] (0,-\squaresize) rectangle ++(8*\squaresize,-8*\squaresize);
			\draw[step=\squaresize, blue, line width = \gridlinewidth] (0,-\squaresize) grid ++(8*\squaresize,-8*\squaresize);
		
		    \path[fill=black,line join=miter,line cap=butt,line width=0.800pt] (12.5*\squaresize,-2*\squaresize) node[] {6\,@\,8$\times$8};
			\draw[blue, line width = \borderlinewidth, fill=white] (2*\squaresize,-3*\squaresize) rectangle ++(8*\squaresize,-8*\squaresize);
			\draw[blue, line width = \borderlinewidth, fill=white] (2.5*\squaresize,-3.5*\squaresize) rectangle ++(8*\squaresize,-8*\squaresize);
			\draw[blue, line width = \borderlinewidth, fill=white] (3*\squaresize,-4*\squaresize) rectangle ++(8*\squaresize,-8*\squaresize);
			\draw[blue, line width = \borderlinewidth, fill=white] (3.5*\squaresize,-4.5*\squaresize) rectangle ++(8*\squaresize,-8*\squaresize);
			\draw[blue, line width = \borderlinewidth, fill=white] (4*\squaresize,-5*\squaresize) rectangle ++(8*\squaresize,-8*\squaresize);
			\draw[blue, line width = \borderlinewidth, fill=white] (4.5*\squaresize,-5.5*\squaresize) rectangle ++(8*\squaresize,-8*\squaresize);
			\draw[step=\squaresize, blue, line width = \gridlinewidth, shift={(0.5*\squaresize,-0.5*\squaresize)}] (4*\squaresize,-5*\squaresize) grid ++(8*\squaresize,-8*\squaresize);
		\end{scope}
		
		\begin{scope}[shift={(14*\squaresize,0)}]
			\path[fill=black,line join=miter,line cap=butt,line width=0.800pt] (0.5*\arrowlength,-4.5*\squaresize) node[align=center] {conv.};
	\path[fill=black,line join=miter,line cap=butt,line width=0.800pt] (0.5*\arrowlength,-9*\squaresize) node[align=center] {Kernel:\\3$\times$3};
			\path[draw=black,-Latex,line join=miter,line cap=butt,miter limit=4.00,even odd  rule,line width=\borderlinewidth] (0,-6*\squaresize) -- ++ (\arrowlength,0);
		\end{scope}
		
		\begin{scope}[shift={(22*\squaresize,0)}]
		    \path[fill=black,line join=miter,line cap=butt,line width=0.800pt] (4*\squaresize,0) node[] {150\,@\,8$\times$8};
			\draw[blue, line width = \borderlinewidth, fill=white] (0,-\squaresize) rectangle ++(8*\squaresize,-8*\squaresize);
			\draw[blue, line width = \borderlinewidth, fill=white] (0.5*\squaresize,-1.5*\squaresize) rectangle ++(8*\squaresize,-8*\squaresize);
		
			\draw[dotted, color=blue, line width=\borderlinewidth] (0.5*\squaresize,-9.5*\squaresize) -- (2*\squaresize,-11.5*\squaresize);
			\draw[dotted, color=blue, line width=\borderlinewidth] (0.5*\squaresize,-1.5*\squaresize) -- (2*\squaresize,-3.5*\squaresize);
			\draw[dotted, color=blue, line width=\borderlinewidth] (8.5*\squaresize,-1.5*\squaresize) -- (10*\squaresize,-3.5*\squaresize);
		
			\draw[blue, line width = \borderlinewidth, fill=white] (2*\squaresize,-3.5*\squaresize) rectangle ++(8*\squaresize,-8*\squaresize);
			\draw[blue, line width = \borderlinewidth, fill=white] (2.5*\squaresize,-4*\squaresize) rectangle ++(8*\squaresize,-8*\squaresize);
		\end{scope}
	
		\begin{scope}[shift={(34*\squaresize,0)}]
			\path[fill=black,line join=miter,line cap=butt,line width=0.800pt] (0.5*\arrowlength,-4.5*\squaresize) node[align=center] {stack.};
			\path[draw=black,-Latex,line join=miter,line cap=butt,miter limit=4.00,even odd  rule,line width=\borderlinewidth] (0,-6*\squaresize) -- ++ (\arrowlength,0);
		\end{scope}

		\begin{scope}[shift={(42*\squaresize,0)}]
		    \path[fill=black,line join=miter,line cap=butt,line width=0.800pt] (4*\squaresize,0) node[] {300\,@\,8$\times$8};
			\draw[red, line width = \borderlinewidth, fill=white] (0,-\squaresize) rectangle ++(8*\squaresize,-8*\squaresize);
			
			\draw[densely dotted, color=red, line width=\borderlinewidth] (0,-9*\squaresize) --++ (1.5*\squaresize,-1.5*\squaresize);
			\draw[densely dotted, color=red, line width=\borderlinewidth] (0,-\squaresize) -- ++(1.5*\squaresize,-1.5*\squaresize);
			\draw[densely dotted, color=red, line width=\borderlinewidth] (8*\squaresize,-\squaresize) -- ++(1.5*\squaresize,-1.5*\squaresize);			

			\draw[red, line width = \borderlinewidth, fill=white] (1.5*\squaresize,-2.5*\squaresize) rectangle ++(8*\squaresize,-8*\squaresize);
			\draw[blue, line width = \borderlinewidth, fill=white] (2*\squaresize,-3*\squaresize) rectangle ++(8*\squaresize,-8*\squaresize);	
			
			\draw[densely dotted, color=blue, line width=\borderlinewidth] (2*\squaresize,-11*\squaresize) -- ++(1.5*\squaresize,-1.5*\squaresize);
			\draw[densely dotted, color=blue, line width=\borderlinewidth] (2*\squaresize,-3*\squaresize) -- ++(1.5*\squaresize,-1.5*\squaresize);
			\draw[densely dotted, color=blue, line width=\borderlinewidth] (10*\squaresize,-3*\squaresize) -- ++(1.5*\squaresize,-1.5*\squaresize);

			\draw[blue, line width = \borderlinewidth, fill=white] (3.5*\squaresize,-4.5*\squaresize) rectangle ++(8*\squaresize,-8*\squaresize);
		\end{scope}
	
		\begin{scope}[shift={(55*\squaresize,0)}]
			\path[fill=black,line join=miter,line cap=butt,line width=0.800pt] (0.5*\arrowlength,-4.5*\squaresize) node[align=center] {conv.};
			\path[fill=black,line join=miter,line cap=butt,line width=0.800pt] (0.5*\arrowlength,-9*\squaresize) node[align=center] {Kernel:\\1$\times$1};
			\path[draw=black,-Latex,line join=miter,line cap=butt,miter limit=4.00,even odd  rule,line width=\borderlinewidth] (0,-6*\squaresize) -- ++ (\arrowlength,0);
		\end{scope}	

		\begin{scope}[shift={(63*\squaresize,0)}]
		    \path[fill=black,line join=miter,line cap=butt,line width=0.800pt] (4*\squaresize,0) node[] {150\,@\,8$\times$8};
			\draw[blue, line width = \borderlinewidth, fill=white] (0,-\squaresize) rectangle ++(8*\squaresize,-8*\squaresize);
			\draw[blue, line width = \borderlinewidth, fill=white] (0.5*\squaresize,-1.5*\squaresize) rectangle ++(8*\squaresize,-8*\squaresize);
		
			\draw[dotted, color=blue, line width=\borderlinewidth] (0.5*\squaresize,-9.5*\squaresize) -- (2*\squaresize,-11.5*\squaresize);
			\draw[dotted, color=blue, line width=\borderlinewidth] (0.5*\squaresize,-1.5*\squaresize) -- (2*\squaresize,-3.5*\squaresize);
			\draw[dotted, color=blue, line width=\borderlinewidth] (8.5*\squaresize,-1.5*\squaresize) -- (10*\squaresize,-3.5*\squaresize);
		
			\draw[blue, line width = \borderlinewidth, fill=white] (2*\squaresize,-3.5*\squaresize) rectangle ++(8*\squaresize,-8*\squaresize);
			\draw[blue, line width = \borderlinewidth, fill=white] (2.5*\squaresize,-4*\squaresize) rectangle ++(8*\squaresize,-8*\squaresize);
		\end{scope}

				\path[draw=black,-Latex,line join=miter,line cap=butt,miter limit=4.00,even odd  rule,line width=\borderlinewidth] (76*\squaresize,-6*\squaresize) -- (\endpoint,-6*\squaresize)--++(0,\leveldistance)--++(\arrowlength,0);
			\path[fill=black,line join=miter,line cap=butt,line width=0.800pt] (95*\squaresize,-4.5*\squaresize) node[align=center] {conv., Kernel: 1$\times$1};
\end{scope}

    \path[fill=black,line join=miter,line cap=butt,line width=0.800pt]  (7*\squaresize,-2*\leveldistance-15*\squaresize) node[align=center, below] {\footnotesize Environment \&\\\footnotesize goal maps};

\begin{scope}[shift={(\endpoint+\arrowlength+5*\squaresize,-\leveldistance-5*\squaresize)}]
		\begin{scope}[shift={(-2.5*\squaresize,2.5*\squaresize)}]
		    \path[fill=black,line join=miter,line cap=butt,line width=0.800pt] (12*\squaresize,\squaresize) node[] {6\,@\,8$\times$8};
			\draw[white, line width = \borderlinewidth, fill=white] (-1*\squaresize,0) rectangle ++(8*\squaresize,-8*\squaresize);
			\draw[blue, line width = \borderlinewidth] (\squaresize,-10*\squaresize)--++(-2*\squaresize,2*\squaresize)  --++(0,8*\squaresize);	\draw[blue, line width = \borderlinewidth] (\squaresize,-2*\squaresize)--++(-2*\squaresize,2*\squaresize) --++(8*\squaresize,0)--++(2*\squaresize,-2*\squaresize);
			\draw[blue, line width = \borderlinewidth, fill=white] (\squaresize,-2*\squaresize) rectangle ++(8*\squaresize,-8*\squaresize);
		\end{scope}
		
		\begin{scope}[shift={(-1*\squaresize,1*\squaresize)}]
		    \path[fill=black,line join=miter,line cap=butt,line width=0.800pt] (12*\squaresize,0) node[] {2\,@\,8$\times$8};
			\draw[white, line width = \borderlinewidth, fill=white] (0,-\squaresize) rectangle ++(8*\squaresize,-8*\squaresize);
			\draw[red, line width = \borderlinewidth] (\squaresize,-10*\squaresize)--++(-\squaresize,\squaresize)  --++(0,8*\squaresize);	\draw[red, line width = \borderlinewidth] (\squaresize,-2*\squaresize)--++(-\squaresize,\squaresize) --++(8*\squaresize,0)--++(\squaresize,-\squaresize);
			\draw[red, line width = \borderlinewidth, fill=white] (\squaresize,-2*\squaresize) rectangle ++(8*\squaresize,-8*\squaresize);
		\end{scope}
		
		\begin{scope}[shift={(0,0)}]
		    \path[fill=black,line join=miter,line cap=butt,line width=0.800pt] (13*\squaresize,-\squaresize) node[] {1\,@\,8$\times$8};
			\draw[white, line width = \borderlinewidth, fill=white] (0.5*\squaresize,-1.5*\squaresize) rectangle ++(8*\squaresize,-8*\squaresize);
			\draw[black, line width = \borderlinewidth] (\squaresize,-10*\squaresize)--++(-0.5*\squaresize,0.5*\squaresize)  --++(0,8*\squaresize);	\draw[black, line width = \borderlinewidth] (\squaresize,-2*\squaresize)--++(-0.5*\squaresize,0.5*\squaresize) --++(8*\squaresize,0)--++(0.5*\squaresize,-0.5*\squaresize);
			\draw[black, line width = \borderlinewidth, fill=white] (\squaresize,-2*\squaresize) rectangle ++(8*\squaresize,-8*\squaresize);
			\draw[step=\squaresize, black, line width = \gridlinewidth] (\squaresize,-2*\squaresize) grid ++(8*\squaresize,-8*\squaresize);
		\end{scope}
    \path[fill=black,line join=miter,line cap=butt,line width=0.800pt]  (3*\squaresize,-12*\squaresize) node[align=center, below] {\footnotesize Multilevel\\\footnotesize reward map};
\end{scope}
\end{tikzpicture}

%% file: fig/vi_module_new.pgf
\definecolor{cffffff}{RGB}{255,255,255}
\definecolor{c0a0a0a}{RGB}{10,10,10}
\definecolor{c00ff00}{RGB}{0,120,0}

\begin{tikzpicture}[y=0.80pt, x=0.80pt, yscale=-1.050000, xscale=1.050000, inner sep=0pt, outer sep=0pt,font=\footnotesize]
\begin{scope}[shift={(0,-560)}]
  \path[draw=black,fill=cffffff,dash pattern=on 1.51pt off 0.19pt,miter
    limit=4.00,line width=0.189pt,rounded corners=0.0000cm] (103.9878,598.8210)
    rectangle (141.7951,636.6283);
  \path[draw=black,fill=cffffff,dash pattern=on 1.34pt off 0.17pt,miter
    limit=4.00,line width=0.167pt,rounded corners=0.0000cm] (106.1824,601.0156)
    rectangle (139.6005,634.4337);
  \begin{scope}[cm={{0.5736,0.0,0.0,0.5736,(90.42632,287.26833)}}]
    \path[xscale=1.000,yscale=-1.000,draw=black,fill=cffffff,miter limit=4.00,line
      width=0.255pt,rounded corners=0.0000cm] (162.1253,-605.0493) rectangle
      (213.0128,-554.1288);
    \path[xscale=1.000,yscale=-1.000,draw=black,fill=cffffff,miter limit=4.00,line
      width=0.255pt,rounded corners=0.0000cm] (164.9595,-608.1165) rectangle
      (215.8470,-557.1960);
    \path[xscale=1.000,yscale=-1.000,draw=black,fill=cffffff,miter limit=4.00,line
      width=0.255pt,rounded corners=0.0000cm] (167.7936,-611.1838) rectangle
      (218.6811,-560.2633);
    \path[xscale=1.000,yscale=-1.000,draw=black,fill=cffffff,miter limit=4.00,line
      width=0.255pt,rounded corners=0.0000cm] (170.6277,-614.2510) rectangle
      (221.5152,-563.3305);
  \end{scope}
  \path[xscale=1.000,yscale=-1.000,draw=black,fill=cffffff,miter limit=4.00,line
    width=0.146pt,rounded corners=0.0000cm] (253.1836,-636.9641) rectangle
    (282.3727,-607.7561);
  \begin{scope}[cm={{0.5736,0.0,0.0,0.5736,(89.77685,287.52136)}}]
    \begin{scope}[shift={(-51.07143,-48.57143)}]
      \path[draw=black,fill=cffffff,dash pattern=on 0.58pt off 0.29pt,miter
        limit=4.00,nonzero rule,line width=0.291pt,rounded corners=0.0000cm]
        (227.0714,643.0051) rectangle (235.0714,651.0051);
      \path[draw=black,fill=cffffff,dash pattern=on 0.58pt off 0.29pt,miter
        limit=4.00,nonzero rule,line width=0.291pt,rounded corners=0.0000cm]
        (229.4762,644.6956) rectangle (237.4762,652.6956);
      \path[draw=black,fill=cffffff,dash pattern=on 0.58pt off 0.29pt,miter
        limit=4.00,nonzero rule,line width=0.291pt,rounded corners=0.0000cm]
        (231.8810,646.3860) rectangle (239.8810,654.3860);
      \path[draw=black,fill=cffffff,miter limit=4.00,nonzero rule,line
        width=0.291pt,rounded corners=0.0000cm] (234.2857,648.0765) rectangle
        (242.2857,656.0765);
    \end{scope}
    \path[draw=black,fill=cffffff,miter limit=4.00,nonzero rule,line
      width=0.139pt,rounded corners=0.0000cm] (291.2979,600.5242) rectangle
      (295.1238,604.3501);
    \path[draw=black,dash pattern=on 0.25pt off 0.25pt,line join=miter,line
      cap=butt,miter limit=4.00,even odd rule,line width=0.254pt]
      (191.2143,607.5051) -- (291.2979,604.3501);
    \path[draw=black,dash pattern=on 0.25pt off 0.25pt,line join=miter,line
      cap=butt,miter limit=4.00,even odd rule,line width=0.254pt]
      (184.0000,594.4336) -- (291.2979,600.5242);
  \end{scope}
  \path[fill=black,line join=miter,line cap=butt,line width=0.800pt]
    (267.7803,653) node[align=center] (text5324-7-3) {Level $l$\\State-Value \\Map};
  \path[draw=black,-Latex,line join=miter,line cap=butt,miter limit=4.00,even odd
    rule,line width=0.218pt] (291.1517,625.2522) -- (297.7664,625.2522) --
    (297.7664,670) -- (19.4696,670) -- (19.4696,622.3842) --
    (27.9099,622.3842);
  \path[fill=black,line join=miter,line cap=butt,line width=0.800pt]
    (144.6317,675) node[align=center] (text5324-7-5) {K recurrence};
  \path[draw=black,-Latex,line join=miter,line cap=butt,even odd rule,line width=0.186pt]
    (151.3440,622.3601) -- (181.3243,622.3601);
  \path[fill=black,line join=miter,line cap=butt,line width=0.800pt]
    (165.7907,615) node[align=center] (text5374-3-5-4) {conv.};
  \path[fill=black,line join=miter,line cap=butt,line width=0.800pt]
    (168,625) node[align=center,below] (text5374-3-5-2-1) {Kernel:
    \\\color{c00ff00}3$\times$3\\\color{purple}3$\times$3$\times$3};
  \path[fill=black,line join=miter,line cap=butt,line width=0.800pt]
    (235.1272,610) node[align=center] (text5374-3-5-9-1) {Max\\pool};
  \path[draw=black,-Latex,line join=miter,line cap=butt,even odd rule,line width=0.186pt]
    (218.6427,624.9032) -- (251.7952,624.9032);
  \path[draw=black,fill=cffffff,dash pattern=on 1.34pt off 0.17pt,miter
    limit=4.00,line width=0.167pt,rounded corners=0.0000cm] (29.3483,598.6101)
    rectangle (62.7664,632.0281);
  \path[draw=black,fill=cffffff,miter limit=4.00,line width=0.167pt]
    (35.3585,639.1973) -- (35.3585,639.1973) -- (35.3585,605.7792) --
    (68.7766,605.7792) -- (68.7766,605.7792);
  \path[draw=black,line join=miter,line cap=butt,miter limit=4.00,even odd
    rule,line width=0.161pt] (37.4602,641.2989) -- (35.3585,639.1973);
  \path[draw=black,line join=miter,line cap=butt,miter limit=4.00,even odd
    rule,line width=0.161pt] (37.4602,607.8808) -- (35.3585,605.7792);
  \path[draw=black,line join=miter,line cap=butt,miter limit=4.00,even odd
    rule,line width=0.161pt] (70.8783,607.8808) -- (68.7766,605.7792);
  \begin{scope}[cm={{0.2087,0.0,0.0,0.2087,(29.00769,488.55861)}},draw=black]
    \path[draw=black,fill=cffffff,miter limit=4.00,line width=0.800pt]
      (40.5000,571.8622) -- (200.5000,571.8622) -- (200.5000,731.8622) --
      (40.5000,731.8622) -- cycle;
    \path[draw=black,line join=miter,line cap=butt,fill opacity=0.750,even odd
      rule,line width=0.200pt] (40.1250,571.3622) --
      (40.1250,732.3622)(60.1250,571.3622) -- (60.1250,732.3622)(80.1250,571.3622)
      -- (80.1250,732.3622)(100.1250,571.3622) --
      (100.1250,732.3622)(120.1250,571.3622) --
      (120.1250,732.3622)(140.1250,571.3622) --
      (140.1250,732.3622)(160.1250,571.3622) --
      (160.1250,732.3622)(180.1250,571.3622) --
      (180.1250,732.3622)(200.1250,571.3622) --
      (200.1250,732.3622)(40.1250,571.3622) -- (201.1250,571.3622)(40.1250,591.3622)
      -- (201.1250,591.3622)(40.1250,611.3622) --
      (201.1250,611.3622)(40.1250,631.3622) -- (201.1250,631.3622)(40.1250,651.3622)
      -- (201.1250,651.3622)(40.1250,671.3622) --
      (201.1250,671.3622)(40.1250,691.3622) -- (201.1250,691.3622)(40.1250,711.3622)
      -- (201.1250,711.3622)(40.1250,731.3622) -- (201.1250,731.3622);
  \end{scope}
  \path[fill=black,line join=miter,line cap=butt,line width=0.800pt]
    (56.2578,655) node[align=center] (text5324-7-39) {
    Level $l$\\Reward Map};
  \path[fill=c0a0a0a,line join=miter,line cap=butt,line width=0.800pt]
    (45.9391,588) node[align=center] (text5324-7-3-8) {old  State-Value\\ Map (Level $l$)};
  \path[fill=black,line join=miter,line cap=butt,line width=0.800pt]
    (87.0462,615) node[align=center] (text5374-3-5-9-1-6) {pad};
  \path[draw=black, -Latex,line join=miter,line cap=butt,even odd rule,line width=0.186pt]
    (73.3004,622.4977) -- (100.3690,622.4977);
  \begin{scope}[cm={{0.5736,0.0,0.0,0.5736,(21.57036,198.41855)}}]
    \path[draw=black,fill=cffffff,miter limit=4.00,line width=0.329pt]
      (151.9509,773.6567) -- (151.9508,773.6567) -- (151.9508,707.9142) --
      (217.6933,707.9142) -- (217.6933,707.9142);
    \path[draw=black,line join=miter,line cap=butt,miter limit=4.00,even odd
      rule,line width=0.317pt] (156.0855,777.7913) -- (151.9509,773.6567);
    \path[draw=black,line join=miter,line cap=butt,miter limit=4.00,even odd
      rule,line width=0.317pt] (156.0855,712.0488) -- (151.9509,707.9142);
    \path[draw=black,line join=miter,line cap=butt,miter limit=4.00,even odd
      rule,line width=0.317pt] (221.8280,712.0488) -- (217.6934,707.9142);
    \begin{scope}[shift={(-248.51361,8.66134)}]
      \path[draw=black,fill=cffffff,dash pattern=on 2.64pt off 0.33pt,miter
        limit=4.00,line width=0.329pt,rounded corners=0.0000cm] (404.9450,703.7334)
        rectangle (470.8572,769.6456);
      \begin{scope}[cm={{0.36384,0.0,0.0,0.36384,(394.03522,499.53621)}},draw=black]
        \path[draw=black,fill=cffffff,miter limit=4.00,line width=0.800pt]
          (40.5000,571.8622) -- (200.5000,571.8622) -- (200.5000,731.8622) --
          (40.5000,731.8622) -- cycle;
        \path[draw=black,line join=miter,line cap=butt,fill opacity=0.750,even odd
          rule,line width=0.200pt] (40.1250,571.3622) --
          (40.1250,732.3622)(60.1250,571.3622) -- (60.1250,732.3622)(80.1250,571.3622)
          -- (80.1250,732.3622)(100.1250,571.3622) --
          (100.1250,732.3622)(120.1250,571.3622) --
          (120.1250,732.3622)(140.1250,571.3622) --
          (140.1250,732.3622)(160.1250,571.3622) --
          (160.1250,732.3622)(180.1250,571.3622) --
          (180.1250,732.3622)(200.1250,571.3622) --
          (200.1250,732.3622)(40.1250,571.3622) -- (201.1250,571.3622)(40.1250,591.3622)
          -- (201.1250,591.3622)(40.1250,611.3622) --
          (201.1250,611.3622)(40.1250,631.3622) -- (201.1250,631.3622)(40.1250,651.3622)
          -- (201.1250,651.3622)(40.1250,671.3622) --
          (201.1250,671.3622)(40.1250,691.3622) -- (201.1250,691.3622)(40.1250,711.3622)
          -- (201.1250,711.3622)(40.1250,731.3622) -- (201.1250,731.3622);
      \end{scope}
    \end{scope}
  \end{scope}
\end{scope}

\end{tikzpicture}

%% file: fig/padding.pgf
\definecolor{cff0000}{RGB}{255,0,0}
\definecolor{cffffff}{RGB}{255,255,255}

\begin{tikzpicture}[y=0.80pt, x=0.80pt, yscale=-1.000000, xscale=1.000000, inner sep=0pt, outer sep=0pt,font=\footnotesize]
\begin{scope}
  \path[fill=black,line join=miter,line cap=butt,line width=0.800pt]
    (350.7249,675.4743) node[align=center] (text4393) {};
    \begin{scope}[shift={(-107.83571,1.87114)}]

\begin{scope}[cm={{0.48459,0.0,0.0,0.48459,(121.27163,524.71333)}}]
          \path[draw=cff0000,line join=miter,line cap=butt,fill opacity=0.750,even odd
            rule,line width=0.200pt] 
(88.3154,584) -- (88.3154,718)
(108.3154,584)-- (108.3154,718)
(128.3154,584) -- (128.3154,718)
(148.3154,584) --(148.3154,718)
(168.3154,584) --(168.3154,718)
(188.3154,584) --(188.3154,718)
(208.3154,584) --(208.3154,718)
(81,591.1546)-- (215,591.1546)
(81,611.1546) -- (215,611.1546)
(81,631.1546) -- (215,631.1546)
(81,651.1546) -- (215,651.1546)
(81,671.1546) --(215,671.1546)
(81,691.1546) -- (215,691.1546)
(81,711.1546) -- (215,711.1546);
        \end{scope}
        \begin{scope}[cm={{0.48459,0.0,0.0,0.48459,(111.77409,533.81729)}}]
          \path[draw=black,miter limit=4.00,line width=0.800pt,rounded corners=0.0000cm]
            (128.2894,592.7428) rectangle (208.2894,672.7428);
          \path[draw=black,line join=miter,line cap=butt,fill opacity=0.750,even odd
            rule,line width=0.200pt] (127.9144,592.3678) --
            (127.9144,673.3678)(137.9144,592.3678) --
            (137.9144,673.3678)(147.9144,592.3678) --
            (147.9144,673.3678)(157.9144,592.3678) --
            (157.9144,673.3678)(167.9144,592.3678) --
            (167.9144,673.3678)(177.9144,592.3678) --
            (177.9144,673.3678)(187.9144,592.3678) --
            (187.9144,673.3678)(197.9144,592.3678) --
            (197.9144,673.3678)(207.9144,592.3678) --
            (207.9144,673.3678)(127.9144,592.3678) --
            (208.9144,592.3678)(127.9144,602.3678) --
            (208.9144,602.3678)(127.9144,612.3678) --
            (208.9144,612.3678)(127.9144,622.3678) --
            (208.9144,622.3678)(127.9144,632.3678) --
            (208.9144,632.3678)(127.9144,642.3678) --
            (208.9144,642.3678)(127.9144,652.3678) --
            (208.9144,652.3678)(127.9144,662.3678) --
            (208.9144,662.3678)(127.9144,672.3678) -- (208.9144,672.3678);
        \end{scope}
	\begin{scope}[shift={(0,10)}]
          \path[fill=cff0000,line join=miter,line cap=butt,line width=0.800pt]
            (169,806.5) node[align=center,color=red] (text4199) {$1$};
          \path[fill=cff0000,line join=miter,line cap=butt,line width=0.800pt]
            (169,816) node[align=center,color=red] (text4199-5) {$2$};
          \path[fill=cff0000,line join=miter,line cap=butt,line width=0.800pt]
            (169,826) node[align=center,color=red] (text4199-56) {$3$};
          \path[fill=cff0000,line join=miter,line cap=butt,line width=0.800pt]
            (169,835.5) node[align=center,color=red] (text4199-9) {$4$};
          \path[fill=cff0000,line join=miter,line cap=butt,line width=0.800pt]
            (169,845) node[align=center,color=red] (text4199-2) {$5$};
          \path[fill=cff0000,line join=miter,line cap=butt,line width=0.800pt]
            (169,855) node[align=center,color=red] (text4199-0) {$6$};
	\end{scope}
        \begin{scope}[cm={{0.48459,0.0,0.0,0.48459,(-92.49252,512.1368)}}]
          \path[draw=black,dash pattern=on 1.5pt off 0.50pt,line join=miter,line
            cap=butt,miter limit=4.00,fill opacity=0.750,even odd rule,line width=0.396pt]
            (490.2472,762.6094) -- (490.2472,962.3622)(510.0247,762.6094) --
            (510.0247,962.3622)(529.8022,762.6094) --
            (529.8022,962.3622)(549.5797,762.6094) --
            (549.5797,962.3622)(569.3572,762.6094) --
            (569.3572,962.3622)(589.1347,762.6094) --
            (589.1347,962.3622)(608.9122,762.6094) --
            (608.9122,962.3622)(628.6897,762.6094) --
            (628.6897,962.3622)(648.4672,762.6094) --
            (648.4672,962.3622)(668.2448,762.6094) --
            (668.2448,962.3622)(688.0223,762.6094) --
            (688.0223,962.3622)(490.2472,762.6094) --
            (690.0000,762.6094)(490.2472,782.3869) --
            (690.0000,782.3869)(490.2472,802.1644) --
            (690.0000,802.1644)(490.2472,821.9419) --
            (690.0000,821.9419)(490.2472,841.7194) --
            (690.0000,841.7194)(490.2472,861.4969) --
            (690.0000,861.4969)(490.2472,881.2744) --
            (690.0000,881.2744)(490.2472,901.0519) --
            (690.0000,901.0519)(490.2472,920.8294) --
            (690.0000,920.8294)(490.2472,940.6069) --
            (690.0000,940.6069)(490.2472,960.3844) -- (690.0000,960.3844);
          \path[draw=black,fill=cffffff,miter limit=4.00,line width=1.582pt,rounded
            corners=0.0000cm] (510.7664,783.1285) rectangle (668.9864,941.3486);
          \path[draw=black,line join=miter,line cap=butt,fill opacity=0.750,even odd
            rule,line width=0.396pt] (509.7775,782.1397) --
            (509.7775,942.3375)(529.5550,782.1397) --
            (529.5550,942.3375)(549.3325,782.1397) --
            (549.3325,942.3375)(569.1100,782.1397) --
            (569.1100,942.3375)(588.8875,782.1397) --
            (588.8875,942.3375)(608.6650,782.1397) --
            (608.6650,942.3375)(628.4425,782.1397) --
            (628.4425,942.3375)(648.2200,782.1397) --
            (648.2200,942.3375)(667.9975,782.1397) --
            (667.9975,942.3375)(509.7775,782.1397) --
            (669.9753,782.1397)(509.7775,801.9172) --
            (669.9753,801.9172)(509.7775,821.6947) --
            (669.9753,821.6947)(509.7775,841.4722) --
            (669.9753,841.4722)(509.7775,861.2497) --
            (669.9753,861.2497)(509.7775,881.0272) --
            (669.9753,881.0272)(509.7775,900.8047) --
            (669.9753,900.8047)(509.7775,920.5822) --
            (669.9753,920.5822)(509.7775,940.3597) -- (669.9753,940.3597);
          \path[fill=cff0000,line join=miter,line cap=butt,line width=0.800pt]
            (501,773) node[align=center,color=red] (text4199-3) {$1$};
          \path[fill=cff0000,line join=miter,line cap=butt,line width=0.800pt]
            (501,793) node[align=center,color=red] (text4199-5-0) {$2$};
          \path[fill=cff0000,line join=miter,line cap=butt,line width=0.800pt]
            (501,833) node[align=center,color=red] (text4199-56-2) {$3$};
          \path[fill=cff0000,line join=miter,line cap=butt,line width=0.800pt]
            (501,892) node[align=center,color=red] (text4199-9-1) {$4$};
          \path[fill=cff0000,line join=miter,line cap=butt,line width=0.800pt]
            (501,931) node[align=center,color=red] (text4199-2-7) {$5$};
          \path[fill=cff0000,line join=miter,line cap=butt,line width=0.800pt]
            (501,951) node[align=center,color=red] (text4199-0-2) {$6$};
          \path[fill=cff0000,line join=miter,line cap=butt,line width=0.800pt]
            (501,813) node[align=center,color=red] (text4199-5-0-2) {$2$};
          \path[fill=cff0000,line join=miter,line cap=butt,line width=0.800pt]
            (501,852) node[align=center,color=red] (text4199-56-2-7) {$3$};
          \path[fill=cff0000,line join=miter,line cap=butt,line width=0.800pt]
            (501,872) node[align=center,color=red] (text4199-9-1-9) {$4$};
          \path[fill=cff0000,line join=miter,line cap=butt,line width=0.800pt]
            (501,911) node[align=center,color=red] (text4199-2-7-2) {$5$};
        \end{scope}
    \end{scope}
    \begin{scope}[cm={{1.4773,0.0,0.0,1.4773,(157.42609,-542.34123)}}]
        \begin{scope}[shift={(9.5,-61.625)}]
          \begin{scope}[shift={(-351.2212,159.59522)}]
            \path[draw=black,fill=cffffff,miter limit=4.00,line width=0.388pt,rounded
              corners=0.0000cm] (351.4635,853.6967) rectangle (390.2310,892.4641);
            \path[draw=black,line join=miter,line cap=butt,fill opacity=0.750,even odd
              rule,line width=0.097pt] (351.2818,853.5149) --
              (351.2818,892.7670)(356.1277,853.5149) --
              (356.1277,892.7670)(360.9737,853.5149) --
              (360.9737,892.7670)(365.8196,853.5149) --
              (365.8196,892.7670)(370.6655,853.5149) --
              (370.6655,892.7670)(375.5115,853.5149) --
              (375.5115,892.7670)(380.3574,853.5149) --
              (380.3574,892.7670)(385.2033,853.5149) --
              (385.2033,892.7670)(390.0493,853.5149) --
              (390.0493,892.7670)(351.2818,853.5149) --
              (390.5339,853.5149)(351.2818,858.3609) --
              (390.5339,858.3609)(351.2818,863.2068) --
              (390.5339,863.2068)(351.2818,868.0527) --
              (390.5339,868.0527)(351.2818,872.8987) --
              (390.5339,872.8987)(351.2818,877.7446) --
              (390.5339,877.7446)(351.2818,882.5905) --
              (390.5339,882.5905)(351.2818,887.4365) --
              (390.5339,887.4365)(351.2818,892.2824) -- (390.5339,892.2824);
          \end{scope}
          \begin{scope}[shift={(-346.2212,164.59522)}]
            \path[draw=black,fill=cffffff,miter limit=4.00,line width=0.388pt,rounded
              corners=0.0000cm] (351.4635,853.6967) rectangle (390.2310,892.4641);
            \path[draw=black,line join=miter,line cap=butt,fill opacity=0.750,even odd
              rule,line width=0.097pt] (351.2818,853.5149) --
              (351.2818,892.7670)(356.1277,853.5149) --
              (356.1277,892.7670)(360.9737,853.5149) --
              (360.9737,892.7670)(365.8196,853.5149) --
              (365.8196,892.7670)(370.6655,853.5149) --
              (370.6655,892.7670)(375.5115,853.5149) --
              (375.5115,892.7670)(380.3574,853.5149) --
              (380.3574,892.7670)(385.2033,853.5149) --
              (385.2033,892.7670)(390.0493,853.5149) --
              (390.0493,892.7670)(351.2818,853.5149) --
              (390.5339,853.5149)(351.2818,858.3609) --
              (390.5339,858.3609)(351.2818,863.2068) --
              (390.5339,863.2068)(351.2818,868.0527) --
              (390.5339,868.0527)(351.2818,872.8987) --
              (390.5339,872.8987)(351.2818,877.7446) --
              (390.5339,877.7446)(351.2818,882.5905) --
              (390.5339,882.5905)(351.2818,887.4365) --
              (390.5339,887.4365)(351.2818,892.2824) -- (390.5339,892.2824);
          \end{scope}
          \begin{scope}[shift={(-341.2212,169.59522)}]
            \path[draw=black,fill=cffffff,miter limit=4.00,line width=0.388pt,rounded
              corners=0.0000cm] (351.4635,853.6967) rectangle (390.2310,892.4641);
            \path[draw=black,line join=miter,line cap=butt,fill opacity=0.750,even odd
              rule,line width=0.097pt] (351.2818,853.5149) --
              (351.2818,892.7670)(356.1277,853.5149) --
              (356.1277,892.7670)(360.9737,853.5149) --
              (360.9737,892.7670)(365.8196,853.5149) --
              (365.8196,892.7670)(370.6655,853.5149) --
              (370.6655,892.7670)(375.5115,853.5149) --
              (375.5115,892.7670)(380.3574,853.5149) --
              (380.3574,892.7670)(385.2033,853.5149) --
              (385.2033,892.7670)(390.0493,853.5149) --
              (390.0493,892.7670)(351.2818,853.5149) --
              (390.5339,853.5149)(351.2818,858.3609) --
              (390.5339,858.3609)(351.2818,863.2068) --
              (390.5339,863.2068)(351.2818,868.0527) --
              (390.5339,868.0527)(351.2818,872.8987) --
              (390.5339,872.8987)(351.2818,877.7446) --
              (390.5339,877.7446)(351.2818,882.5905) --
              (390.5339,882.5905)(351.2818,887.4365) --
              (390.5339,887.4365)(351.2818,892.2824) -- (390.5339,892.2824);
          \end{scope}
          \begin{scope}[shift={(-321.2212,189.59522)}]
            \path[draw=black,fill=cffffff,miter limit=4.00,line width=0.388pt,rounded
              corners=0.0000cm] (351.4635,853.6967) rectangle (390.2310,892.4641);
            \path[draw=black,line join=miter,line cap=butt,fill opacity=0.750,even odd
              rule,line width=0.097pt] (351.2818,853.5149) --
              (351.2818,892.7670)(356.1277,853.5149) --
              (356.1277,892.7670)(360.9737,853.5149) --
              (360.9737,892.7670)(365.8196,853.5149) --
              (365.8196,892.7670)(370.6655,853.5149) --
              (370.6655,892.7670)(375.5115,853.5149) --
              (375.5115,892.7670)(380.3574,853.5149) --
              (380.3574,892.7670)(385.2033,853.5149) --
              (385.2033,892.7670)(390.0493,853.5149) --
              (390.0493,892.7670)(351.2818,853.5149) --
              (390.5339,853.5149)(351.2818,858.3609) --
              (390.5339,858.3609)(351.2818,863.2068) --
              (390.5339,863.2068)(351.2818,868.0527) --
              (390.5339,868.0527)(351.2818,872.8987) --
              (390.5339,872.8987)(351.2818,877.7446) --
              (390.5339,877.7446)(351.2818,882.5905) --
              (390.5339,882.5905)(351.2818,887.4365) --
              (390.5339,887.4365)(351.2818,892.2824) -- (390.5339,892.2824);
          \end{scope}
          \begin{scope}[shift={(-316.2212,194.59522)}]
            \path[draw=black,fill=cffffff,miter limit=4.00,line width=0.388pt,rounded
              corners=0.0000cm] (351.4635,853.6967) rectangle (390.2310,892.4641);
            \path[draw=black,line join=miter,line cap=butt,fill opacity=0.750,even odd
              rule,line width=0.097pt] (351.2818,853.5149) --
              (351.2818,892.7670)(356.1277,853.5149) --
              (356.1277,892.7670)(360.9737,853.5149) --
              (360.9737,892.7670)(365.8196,853.5149) --
              (365.8196,892.7670)(370.6655,853.5149) --
              (370.6655,892.7670)(375.5115,853.5149) --
              (375.5115,892.7670)(380.3574,853.5149) --
              (380.3574,892.7670)(385.2033,853.5149) --
              (385.2033,892.7670)(390.0493,853.5149) --
              (390.0493,892.7670)(351.2818,853.5149) --
              (390.5339,853.5149)(351.2818,858.3609) --
              (390.5339,858.3609)(351.2818,863.2068) --
              (390.5339,863.2068)(351.2818,868.0527) --
              (390.5339,868.0527)(351.2818,872.8987) --
              (390.5339,872.8987)(351.2818,877.7446) --
              (390.5339,877.7446)(351.2818,882.5905) --
              (390.5339,882.5905)(351.2818,887.4365) --
              (390.5339,887.4365)(351.2818,892.2824) -- (390.5339,892.2824);
          \end{scope}
          \begin{scope}[shift={(-311.2212,199.59522)}]
            \path[draw=black,fill=cffffff,miter limit=4.00,line width=0.388pt,rounded
              corners=0.0000cm] (351.4635,853.6967) rectangle (390.2310,892.4641);
            \path[draw=black,line join=miter,line cap=butt,fill opacity=0.750,even odd
              rule,line width=0.097pt] (351.2818,853.5149) --
              (351.2818,892.7670)(356.1277,853.5149) --
              (356.1277,892.7670)(360.9737,853.5149) --
              (360.9737,892.7670)(365.8196,853.5149) --
              (365.8196,892.7670)(370.6655,853.5149) --
              (370.6655,892.7670)(375.5115,853.5149) --
              (375.5115,892.7670)(380.3574,853.5149) --
              (380.3574,892.7670)(385.2033,853.5149) --
              (385.2033,892.7670)(390.0493,853.5149) --
              (390.0493,892.7670)(351.2818,853.5149) --
              (390.5339,853.5149)(351.2818,858.3609) --
              (390.5339,858.3609)(351.2818,863.2068) --
              (390.5339,863.2068)(351.2818,868.0527) --
              (390.5339,868.0527)(351.2818,872.8987) --
              (390.5339,872.8987)(351.2818,877.7446) --
              (390.5339,877.7446)(351.2818,882.5905) --
              (390.5339,882.5905)(351.2818,887.4365) --
              (390.5339,887.4365)(351.2818,892.2824) -- (390.5339,892.2824);
          \end{scope}
        \end{scope}
        \path[fill=black,line join=miter,line cap=butt,line width=0.800pt]
          (48.4892,950.7244) node[above right] (text5012) {$\theta= 0$};
        \path[fill=black,line join=miter,line cap=butt,line width=0.800pt]
          (53.5498,955.7244) node[above right] (text5012-1) {$\theta= 15$};
        \path[fill=black,line join=miter,line cap=butt,line width=0.800pt]
          (58.4892,960.7244) node[above right] (text5012-9) {$\theta= 14$};
        \path[fill=black,line join=miter,line cap=butt,line width=0.800pt]
          (78.4892,980.7244) node[above right] (text5012-19) {$\theta=1$};
        \path[fill=black,line join=miter,line cap=butt,line width=0.800pt]
          (83.5498,985.7244) node[above right] (text5012-7) {$\theta= 0$};
        \path[fill=black,line join=miter,line cap=butt,line width=0.800pt]
          (88.5498,990.7244) node[above right] (text5012-4) {$\theta=15$};
        \path[draw=black,dash pattern=on 0.30pt off 3.55pt,line join=miter,line
          cap=butt,miter limit=4.00,even odd rule,dash phase=1.200pt,line width=0.296pt]
          (19.3291,1000.6181) -- (39.5459,1020.0849);
        \path[draw=black,dash pattern=on 0.30pt off 3.55pt,line join=miter,line
          cap=butt,miter limit=4.00,even odd rule,dash phase=1.200pt,line width=0.296pt]
          (58.5166,961.3788) -- (78.2334,981.4706);
        \path[draw=black,dash pattern=on 0.30pt off 3.55pt,line join=miter,line
          cap=butt,miter limit=4.00,even odd rule,dash phase=1.200pt,line width=0.296pt]
          (19.6416,961.5038) -- (39.6084,981.3456);
        \path[fill=black,line join=miter,line cap=butt,line width=0.800pt]
          (77,967) node[align=center] (text5012-9-6) {$\ddots$};
    \end{scope}
\end{scope}

\end{tikzpicture}

%% file: fig/move_2d.pgf
\definecolor{cffffff}{RGB}{255,255,255}
\definecolor{c3232ff}{RGB}{50,50,255}
\definecolor{c0000ff}{RGB}{0,0,255}

\begin{tikzpicture}[y=0.80pt, x=0.80pt, yscale=-1.000000, xscale=1.000000, inner sep=0pt, outer sep=0pt]

    \begin{scope}[shift={(7.76109,875.56976)}]
      \path[draw=black,fill=cffffff,opacity=0.641,miter limit=4.00,line width=0.800pt]
        (203.2108,88.0016) -- (271.5035,88.0016) -- (271.5035,156.2943) --
        (203.2108,156.2943) -- cycle;
      \path[draw=black,line join=miter,line cap=butt,fill opacity=0.750,even odd
        rule,line width=0.500pt] (202.7108,87.5016) --
        (202.7108,156.7943)(212.7108,87.5016) -- (212.7108,156.7943)(222.7108,87.5016)
        -- (222.7108,156.7943)(232.7108,87.5016) --
        (232.7108,156.7943)(242.7108,87.5016) -- (242.7108,156.7943)(252.7108,87.5016)
        -- (252.7108,156.7943)(262.7108,87.5016) --
        (262.7108,156.7943)(202.7108,87.5016) -- (272.0035,87.5016)(202.7108,97.5016)
        -- (272.0035,97.5016)(202.7108,107.5016) --
        (272.0035,107.5016)(202.7108,117.5016) --
        (272.0035,117.5016)(202.7108,127.5016) --
        (272.0035,127.5016)(202.7108,137.5016) --
        (272.0035,137.5016)(202.7108,147.5016) -- (272.0035,147.5016);
    \end{scope}
    \path[fill=c0000ff,opacity=0.673,miter limit=4.00,line width=0.081pt]
      (245.1182,997.7177) ellipse (0.1129cm and 0.1129cm);
    \begin{scope}[shift={(87.10999,-6.96252)},draw=black,-latex]
      \begin{scope}[cm={{1.00001,0.0,0.0,1.0,(-247.93124,773.39916)}},draw=black,-latex]
        \path[draw=black,-latex,line join=miter,line cap=butt,even odd rule,line
          width=0.800pt] (405.9345,231.2812) -- (416.1318,231.0838);
        \path[draw=black,-latex,line join=miter,line cap=butt,even odd rule,line
          width=0.800pt] (405.9345,231.2812) -- (406.1319,221.0838);
        \path[draw=black,-latex,line join=miter,line cap=butt,even odd rule,line
          width=0.800pt] (405.9345,231.2812) -- (396.1320,231.0838);
        \path[draw=black,-latex,line join=miter,line cap=butt,even odd rule,line
          width=0.800pt] (405.9345,231.2812) -- (406.1319,241.0838);
      \end{scope}
      \begin{scope}[cm={{0.70712,-0.70712,0.70711,0.70711,(-292.57468,1128.1824)}},draw=black,-latex]
        \path[draw=black,-latex,line join=miter,line cap=butt,even odd rule,line
          width=0.800pt] (405.9345,231.2812) -- (420.3556,231.2811);
        \path[draw=black,-latex,line join=miter,line cap=butt,even odd rule,line
          width=0.800pt] (406.2137,231.2811) -- (406.2137,217.1390);
        \path[draw=black,-latex,line join=miter,line cap=butt,even odd rule,line
          width=0.800pt] (405.9345,231.2812) -- (392.0717,231.2811);
        \path[draw=black,-latex,line join=miter,line cap=butt,even odd rule,line
          width=0.800pt] (405.9345,231.2812) -- (406.2137,245.4233);
      \end{scope}
    \end{scope}

\end{tikzpicture}

%% file: fig/move_3d.pgf
\definecolor{cffffff}{RGB}{255,255,255}
\definecolor{c3232ff}{RGB}{50,50,255}
\definecolor{c0000ff}{RGB}{0,0,255}

\begin{tikzpicture}[y=0.80pt, x=0.80pt, yscale=-1.000000, xscale=1.000000, inner sep=0pt, outer sep=0pt]
    \begin{scope}[shift={(38.13326,-23.99112)}]
      \begin{scope}[shift={(-202.3983,895.56793)}]
        \path[draw=black,fill=cffffff,opacity=0.641,miter limit=4.00,line width=0.800pt]
          (203.2108,88.0016) -- (271.5035,88.0016) -- (271.5035,156.2943) --
          (203.2108,156.2943) -- cycle;
        \path[draw=black,line join=miter,line cap=butt,fill opacity=0.750,even odd
          rule,line width=0.500pt] (202.7108,87.5016) --
          (202.7108,156.7943)(212.7108,87.5016) -- (212.7108,156.7943)(222.7108,87.5016)
          -- (222.7108,156.7943)(232.7108,87.5016) --
          (232.7108,156.7943)(242.7108,87.5016) -- (242.7108,156.7943)(252.7108,87.5016)
          -- (252.7108,156.7943)(262.7108,87.5016) --
          (262.7108,156.7943)(202.7108,87.5016) -- (272.0035,87.5016)(202.7108,97.5016)
          -- (272.0035,97.5016)(202.7108,107.5016) --
          (272.0035,107.5016)(202.7108,117.5016) --
          (272.0035,117.5016)(202.7108,127.5016) --
          (272.0035,127.5016)(202.7108,137.5016) --
          (272.0035,137.5016)(202.7108,147.5016) -- (272.0035,147.5016);
      \end{scope}
      \begin{scope}[shift={(-420.61258,841.63936)}]
        \path[fill=c3232ff,opacity=0.641,miter limit=4.00,line width=0.798pt]
          (436.4271,161.2180) -- (475.4299,161.2180) -- (475.4299,190.2208) --
          (436.4271,190.2208) -- cycle;
        \begin{scope}[shift={(76.42855,5.0)},fill=black]
          \path[draw=black,fill=black,opacity=0.641,miter limit=4.00,line width=0.511pt]
            (364.8195,186.0388) -- (374.1805,186.0388) -- (374.1805,190.3999) --
            (364.8195,190.3999) -- cycle;
          \path[draw=black,fill=black,opacity=0.641,miter limit=4.00,line width=0.511pt]
            (384.8195,186.0388) -- (394.1805,186.0388) -- (394.1805,190.3999) --
            (384.8195,190.3999) -- cycle;
        \end{scope}
        \begin{scope}[shift={(56.28566,-92.5)},fill=black]
          \path[draw=black,fill=black,opacity=0.641,miter limit=4.00,line width=0.511pt]
            (384.9623,249.4674) -- (394.3234,249.4674) -- (394.3234,253.8285) --
            (384.9623,253.8285) -- cycle;
          \path[draw=black,fill=black,opacity=0.641,miter limit=4.00,line width=0.511pt]
            (404.9623,249.4674) -- (414.3234,249.4674) -- (414.3234,253.8285) --
            (404.9623,253.8285) -- cycle;
        \end{scope}
      \end{scope}
    \end{scope}
    \begin{scope}[shift={(-85.07239,-11.11167)},draw=black,-latex]
      \begin{scope}[cm={{1.00001,0.0,0.0,1.0,(-247.93124,773.39916)}},draw=black,-latex]
        \path[draw=black,-latex,line join=miter,line cap=butt,even odd rule,line
          width=0.800pt] (405.9345,231.2812) -- (416.1318,231.0838);
        \path[draw=black,-latex,line join=miter,line cap=butt,even odd rule,line
          width=0.800pt] (405.9345,231.2812) -- (406.1319,221.0838);
        \path[draw=black,-latex,line join=miter,line cap=butt,even odd rule,line
          width=0.800pt] (405.9345,231.2812) -- (396.1320,231.0838);
        \path[draw=black,-latex,line join=miter,line cap=butt,even odd rule,line
          width=0.800pt] (405.9345,231.2812) -- (406.1319,241.0838);
      \end{scope}
      \begin{scope}[cm={{0.70712,-0.70712,0.70711,0.70711,(-292.57468,1128.1824)}},draw=black,-latex]
        \path[draw=black,-latex,line join=miter,line cap=butt,even odd rule,line
          width=0.800pt] (405.9345,231.2812) -- (420.3556,231.2811);
        \path[draw=black,-latex,line join=miter,line cap=butt,even odd rule,line
          width=0.800pt] (406.2137,231.2811) -- (406.2137,217.1390);
        \path[draw=black,-latex,line join=miter,line cap=butt,even odd rule,line
          width=0.800pt] (405.9345,231.2812) -- (392.0717,231.2811);
        \path[draw=black,-latex,line join=miter,line cap=butt,even odd rule,line
          width=0.800pt] (405.9345,231.2812) -- (406.2137,245.4233);
      \end{scope}
    \end{scope}

\end{tikzpicture}

%% file: fig/rotate.pgf
\definecolor{cffffff}{RGB}{255,255,255}
\definecolor{c3232ff}{RGB}{50,50,255}
\definecolor{c0000ff}{RGB}{0,0,255}

\begin{tikzpicture}[y=0.80pt, x=0.80pt, yscale=-1.000000, xscale=1.000000, inner sep=0pt, outer sep=0pt]
\tikzset{>={Latex[width=1.5mm,length=1.75mm]}}

    \path[fill=cffffff,miter limit=4.00,line width=0.800pt] (124.3592,959.5784) --
      (192.6519,959.5784) -- (192.6519,1027.8711) -- (124.3592,1027.8711) -- cycle;
    \begin{scope}[shift={(-41.92133,8.33376)}]
      \begin{scope}[shift={(-263.99127,809.01428)}]
        \path[fill=c3232ff,opacity=0.641,miter limit=4.00,line width=0.798pt]
          (436.4271,161.2180) -- (475.4299,161.2180) -- (475.4299,190.2208) --
          (436.4271,190.2208) -- cycle;
        \begin{scope}[shift={(76.42855,5.0)},fill=black]
          \path[draw=black,fill=black,opacity=0.641,miter limit=4.00,line width=0.511pt]
            (364.8195,186.0388) -- (374.1805,186.0388) -- (374.1805,190.3999) --
            (364.8195,190.3999) -- cycle;
          \path[draw=black,fill=black,opacity=0.641,miter limit=4.00,line width=0.511pt]
            (384.8195,186.0388) -- (394.1805,186.0388) -- (394.1805,190.3999) --
            (384.8195,190.3999) -- cycle;
        \end{scope}
        \begin{scope}[shift={(56.28566,-92.5)},fill=black]
          \path[draw=black,fill=black,opacity=0.641,miter limit=4.00,line width=0.511pt]
            (384.9623,249.4674) -- (394.3234,249.4674) -- (394.3234,253.8285) --
            (384.9623,253.8285) -- cycle;
          \path[draw=black,fill=black,opacity=0.641,miter limit=4.00,line width=0.511pt]
            (404.9623,249.4674) -- (414.3234,249.4674) -- (414.3234,253.8285) --
            (404.9623,253.8285) -- cycle;
        \end{scope}
      \end{scope}
    \end{scope}
    \path[draw=black,-latex,style=<->,line join=miter,line cap=butt,even odd rule,line
      width=0.800pt] (179.0495,983.0056) .. controls (184.3066,989.9925) and
      (183.1926,996.9794) .. (179.0495,1003.9662);
    \path[draw=black,dash pattern=on 0.40pt off 0.40pt,line join=miter,line
      cap=butt,miter limit=4.00,even odd rule,line width=0.400pt]
      (150.0159,993.0674) -- (191.6764,993.5317);
   
\end{tikzpicture}

%% file: fig/plot/new_training_performance.pgf
\pgfplotsset{footnotesize,samples=10}

\sbox0{\begin{tikzpicture}[]
\begin{axis}[width=0.5\linewidth,ylabel=Success,y label style={at={(0.13,0.45)}},ymin=0, ymax=1.05,,tick label style={font=\scriptsize},legend style={font = \scriptsize,at={(0.5,1.5)},anchor=south}, title=$32\times32$ grid worlds,xtick={0,250,500,750,1000}]
 \addplot[color=red] 
	table[x=Epoch,y=Success] {log/validation_VIN_32.txt};
 \addplot[color=blue] 
	table[x=Epoch,y=Success] {log/validation_AVIN_32_new.txt};
 \addplot[color=orange] 
	table[x=Epoch,y=Success] {log/validation_HVIN_32.txt};
\end{axis}
\end{tikzpicture} }

\sbox1{\begin{tikzpicture}[]
\begin{axis}[width=0.5\linewidth,ylabel=Success,y label style={at={(0.13,0.45)}},ymin=0, ymax=1.05,,tick label style={font=\scriptsize},legend style={font = \scriptsize,at={(0.5,1.5)},anchor=south}, title=$64\times64$ grid worlds,xtick={0,250,500,750,1000}]
 \addplot[color=red] 
	table[x=Epoch,y=Success] {log/validation_VIN_64.txt};
 \addplot[color=blue] 
	table[x=Epoch,y=Success] {log/validation_AVIN_64_new.txt};
 \addplot[color=orange] 
	table[x=Epoch,y=Success] {log/validation_HVIN_64.txt};
\end{axis}
\end{tikzpicture} }

\sbox2{\begin{tikzpicture}[]
\begin{axis}[width=0.5\linewidth,xlabel=Epoch,x label style={at={(0.5,0.1)}},ylabel=Success,y label style={at={(0.13,0.45)}},ymin=0, ymax=1.05,,tick label style={font=\scriptsize},legend style={font = \scriptsize,at={(0.5,1.5)},anchor=south}, title=$128\times128$ grid worlds,xtick={0,250,500,750,1000}]
 \addplot[color=red] 
	table[x=Epoch,y=Success] {log/validation_VIN_128.txt};
 \addplot[color=blue] 
	table[x=Epoch,y=Success] {log/validation_AVIN_128_new.txt};
 \addplot[color=orange] 
	table[x=Epoch,y=Success] {log/validation_HVIN_128.txt};
 \addplot[color=blue,densely dotted] 
	table[x=Epoch,y=Success] {log/validation_AVIN_4_128_new.txt};
 \addplot[color=orange, densely dotted] 
	table[x=Epoch,y=Success] {log/validation_HVIN_4_128.txt};
\end{axis}
\end{tikzpicture} }

\sbox3{\begin{tikzpicture}[]
\begin{axis}[width=0.5\linewidth, ymin=0, ymax=1.05,tick label style={font=\scriptsize},legend style={font = \scriptsize,at={(0.5,1.5)},anchor=south}, title=$32\times32$ grid worlds,xtick={0,250,500,750,1000}]
 \addplot[color=red] 
	table[x=Epoch,y=Success] {log/validation_VIN_32_cyclic.txt};
 \addplot[color=blue] 
	table[x=Epoch,y=Success] {log/validation_AVIN_32_cyclic_new.txt};
 \addplot[color=orange] 
	table[x=Epoch,y=Success] {log/validation_HVIN_32_cyclic.txt};
\end{axis}
\end{tikzpicture}}

\sbox4{\begin{tikzpicture}[]
\begin{axis}[width=0.5\linewidth,ymin=0, ymax=1.05,tick label style={font=\scriptsize},legend style={font = \scriptsize,at={(0.5,1.5)},anchor=south}, title=$64\times64$ grid worlds,xtick={0,250,500,750,1000}]
 \addplot[color=red] 
	table[x=Epoch,y=Success] {log/validation_VIN_64_cyclic.txt};
 \addplot[color=blue] 
	table[x=Epoch,y=Success] {log/validation_AVIN_64_cyclic_new.txt};
 \addplot[color=orange] 
	table[x=Epoch,y=Success] {log/validation_HVIN_64_cyclic.txt};
\end{axis}
\end{tikzpicture} }

\sbox5{\begin{tikzpicture}[]
\begin{axis}[width=0.5\linewidth,xlabel=Epoch,x label style={at={(0.5,0.1)}},ymin=0, ymax=1.05,,tick label style={font=\scriptsize},legend style={font = \scriptsize,at={(0.5,1.5)},anchor=south}, title=$128\times128$ grid worlds,xtick={0,250,500,750,1000}]
 \addplot[color=red] 
	table[x=Epoch,y=Success] {log/validation_VIN_128_cyclic.txt};
 \addplot[color=blue] 
	table[x=Epoch,y=Success] {log/validation_AVIN_128_cyclic_new.txt};
 \addplot[color=orange] 
	table[x=Epoch,y=Success] {log/validation_HVIN_128_cyclic.txt};
 \addplot[color=blue,densely dotted] 
	table[x=Epoch,y=Success] {log/validation_AVIN_4_128_cyclic_new.txt};
 \addplot[color=orange, densely dotted] 
	table[x=Epoch,y=Success] {log/validation_HVIN_4_128_cyclic.txt};
\end{axis}
\end{tikzpicture} }

\sbox6{\begin{tikzpicture}[]
\begin{axis}[width=0.5\linewidth,ylabel=Learning Rate, y label style={at={(0.13,0.45)}}, ymin=0, ymax=0.00119,tick label style={font=\scriptsize},legend style={font = \scriptsize,at={(0.5,1.5)},anchor=south}, title=Fixed learning rate,xtick={0,250,500,750,1000}, title style={at={(0.5,1.12)}}]

 \addplot[color=black] 
	table[x=Epoch,y=lr] {log/learning_rate_const.txt};
\end{axis}
\end{tikzpicture} }

\sbox7{\begin{tikzpicture}[]
\begin{axis}[width=0.5\linewidth, ymin=0, ymax=0.00119,tick label style={font=\scriptsize},legend style={font = \scriptsize,at={(0.,1.5)},anchor=south}, title=Cyclic learning rate,xtick={0,250,500,750,1000}, title style={at={(0.5,1.12)}}]

 \addplot[color=black] 
	table[x=Epoch,y=lr] {log/learning_rate.txt};
\end{axis}
\end{tikzpicture} }

\begin{minipage}{\dimexpr \wd0+\wd1}
\hspace{2.4cm}
\begin{tikzpicture}
\begin{customlegend}[legend columns=2,legend style={align=center,draw=none,column sep=0.5em, font=\footnotesize},
        legend entries={ HVIN, HVIN (4 Levels), AVIN,  AVIN (4 Levels), VIN, Learning Rate}]
        \addlegendimage{solid,line legend,color=orange}
        \addlegendimage{densely dotted,line legend,color=orange}
        \addlegendimage{solid,line legend,color=blue}
        \addlegendimage{densely dotted,line legend,color=blue}
        \addlegendimage{solid,line legend,color=red}
        \addlegendimage{solid,line legend,color=black}
        \end{customlegend}
\end{tikzpicture}

\usebox6\usebox7

\usebox0\usebox3

\usebox1\usebox4

\usebox2\usebox5
\end{minipage}

%% file: fig/3D/occ_map_final.pgf
\begin{tikzpicture}[font=\scriptsize]
\def\step{1.103};
\tikzset{>={Latex[width=1mm,length=1mm]}}
	\draw (0, 0) node[inner sep=0] {\includegraphics[width=0.4\linewidth]{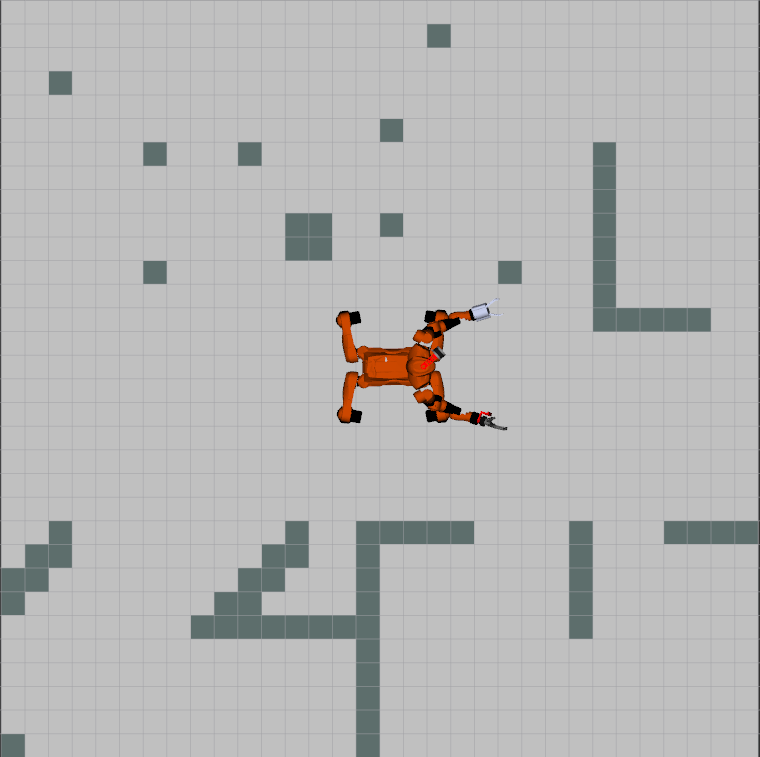}};

\begin{scope}[scale=0.1]
	\begin{scope}[shift={(-32,-32)}]
\begin{scope}[]
 \path[draw=blue] (32.5,32.5)--++(1*\step,-1*\step)
			--++(1*\step,-1*\step)
			--++(1*\step,-1*\step)
			--++(0*\step,-1*\step)
			--++(1*\step,0*\step)
			--++(1*\step,0*\step)
			--++(1*\step,0*\step)
			--++(1*\step,-1*\step)
			--++(0*\step,-1*\step)
			--++(0*\step,-1*\step)
			--++(0*\step,-1*\step)
			--++(0*\step,-1*\step)
			--++(1*\step,-1*\step)
			--++(1*\step,-1*\step)
			--++(0*\step,-1*\step)
			--++(0*\step,-1*\step);

	\path[draw=blue,-Latex,style=->] (32.5+3*\step,32.5-3*\step) --++(1.5,0);
	\path[draw=blue,-Latex,style=->] (32.5+6*\step,32.5-4*\step) --++(1.5,0);
	\path[draw=blue,-Latex,style=->] (32.5+7*\step,32.5-5*\step) --++(1.5,0);
	\path[draw=blue,-Latex,style=->] (32.5+7*\step,32.5-9*\step) --++(1.5,0);
	\path[draw=blue,-Latex,style=->] (32.5+9*\step,32.5-11*\step) --++(1.5,0);
\end{scope}

\begin{scope}[shift={(0.5,0.5)}]
	\draw[-Latex, red](18,19.5) -- ++(0, -2.5);
	  \path[fill=red,line join=miter,line cap=butt]
   (16,20.5)node[align=center,below,red] {I)};

	\draw[-Latex, red](32.7,42.1) -- ++(-1.3,2);
	  \path[fill=red,line join=miter,line cap=butt]
   (32,47.5)node[align=center,below,red] {VII)};

	\draw[-Latex, red](40.5,18) -- ++(2.5,0);
	  \path[fill=red,line join=miter,line cap=butt]
   (38.5,20)node[align=center,below,red] {III)};

\draw[-Latex, red](39.5,45) -- ++(2.2,-0.9);
	  \path[fill=red,line join=miter,line cap=butt]
   (42.5,49)node[align=center,below,red] {VI)};

\draw[-Latex, red](21,35.3) -- ++(-2.5,0);
	  \path[fill=red,line join=miter,line cap=butt]
   (16.5,36)node[align=center,below,red] {IX)};
\draw[-Latex, red](45.7,30) -- ++(-1.3,-2);
	  \path[fill=red,line join=miter,line cap=butt]
   (46.5,29)node[align=center,below,red] {IV)};

\draw[-Latex, red](45,42) -- ++(0,-2.5);
	  \path[fill=red,line join=miter,line cap=butt]
   (46.5,44.5)node[align=center,below,red] {V)};

\draw[-Latex, red](24.3,17) -- ++(2.5,0);
	  \path[fill=red,line join=miter,line cap=butt]
   (23,18)node[align=center,below,red] {II)};

\draw[-Latex, red](21,42.4) -- ++(-2.2,0.9);
	  \path[fill=red,line join=miter,line cap=butt]
   (20,47.)node[align=center,below,red] {VIII)};
\end{scope}

\end{scope}
\end{scope}
	
\end{tikzpicture} 